\documentclass{article} % For LaTeX2e
\usepackage{iclr2021_conference,times}
\usepackage{bm}
\usepackage[utf8]{inputenc} % allow utf-8 input
\usepackage[T1]{fontenc}    % use 8-bit T1 fonts
\usepackage{hyperref}       % hyperlinks
\usepackage{url}            % simple URL typesetting
\usepackage{booktabs}       % professional-quality tables
\usepackage{amsfonts}       % blackboard math symbols
\usepackage{nicefrac}       % compact symbols for 1/2, etc.
\usepackage{microtype}      % microtypography
\usepackage{makecell}

\usepackage{algorithmic}

\usepackage{amsmath,amssymb,amsfonts}
\usepackage{graphicx}
\usepackage{graphics}
\usepackage{multirow} %added new
\usepackage{booktabs} %added new
\usepackage{amssymb} %added new
\usepackage{wrapfig}
\usepackage{hhline}
\usepackage{appendix}
\usepackage{caption}
\usepackage{subcaption}
\usepackage{csquotes}
\usepackage{mathrsfs} 
\usepackage[outline]{contour}
\usepackage{amssymb,amsmath}
\usepackage[ruled,vlined]{algorithm2e}
\usepackage{enumitem}
\usepackage{fancyhdr}
\usepackage{xcolor}
\setlist[enumerate]{label*=\arabic*.}
\usepackage{mathtools}
\DeclarePairedDelimiterX{\infdivx}[2]{(}{)}{%
  #1\;\delimsize\|\;#2%
}
\usepackage{soul}
\usepackage{xcolor}
\definecolor{Red}{rgb}{0, 0, 0}
\newcommand{\red}[1]{{\color{Red} #1}}
\colorlet{usercolorname}{white!70}
\sethlcolor{usercolorname}

\usepackage{hyperref}
\usepackage{url}
\usepackage{adjustbox}

\newsavebox\CBox
\def\textBF#1{\sbox\CBox{#1}\resizebox{\wd\CBox}{\ht\CBox}{\textbf{#1}}}

\usepackage{natbib}
\setcitestyle{citesep={;}, aysep={,}}

\title{Generating Synthetic Mixed-type Longitudinal Electronic Health Records for Artificial Intelligent Applications}

\author{
	Jin Li$^\dag$$^\ddag$, Benjamin J. Cairns$^\ddag$, Jingsong Li$^\dag$$^*$, Tingting Zhu$^\ddag$\thanks{Corresponding authors.}\\
	$^\dag$Zhejiang University, $^\ddag$University of Oxford\\
	\{jin.li, tingting.zhu\}@eng.ox.ac.uk, ben.cairns@ndph.ox.ac.uk, ljs@zju.edu.cn
}

\iclrfinalcopy
\begin{document}

\maketitle

\begin{abstract}
The recent availability of electronic health records (EHRs) have provided enormous opportunities to develop artificial intelligence (AI) algorithms. However, patient privacy has become a major concern that limits data sharing across hospital settings and subsequently hinders the advances in AI. \textit{Synthetic data}, which benefits from the development and proliferation of generative models, has served as a promising substitute for real patient EHR data. 
However, the current generative models are limited as they only generate \textit{single} \textit{type} of clinical data for a synthetic patient, i.e., either continuous-valued or discrete-valued. 
To mimic the nature of clinical decision-making which encompasses various data types/sources, in this study, we propose a generative adversarial network (GAN) entitled EHR-M-GAN which simultaneously synthesizes \textit{mixed-type} timeseries EHR data.
EHR-M-GAN is capable of capturing the multidimensional, heterogeneous, and correlated temporal dynamics in patient trajectories. 
We have validated EHR-M-GAN on three publicly-available intensive care unit databases with records from a total of 141,488 unique patients, and performed privacy risk evaluation of the proposed model.
EHR-M-GAN has demonstrated its superiority over state-of-the-art benchmarks for synthesizing clinical timeseries with high fidelity, while addressing the limitations regarding data types and dimensionality in the current generative models.
Notably, prediction models for outcomes of intensive care performed significantly better when training data was augmented with the addition of EHR-M-GAN-generated timeseries. EHR-M-GAN may have use in developing AI algorithms in resource-limited settings, lowering the barrier for data acquisition while preserving patient privacy. 
\end{abstract}

\section{Introduction}
The past decade has witnessed ground-breaking advancements been made in computational health, owing to the explosion of medical data, such as electronic health records (EHRs) \cite{artzi2020prediction, raket2020dynamic, menger2019machine}. The secondary uses of EHRs give rise to research in a wide range of varieties, especially machine learning (ML)-based digital health solutions for improving the delivery of care \cite{wilkinson2020time, watson2019clinical, futoma2020myth, esteva2021deep, rajkomar2019machine}. However, in practice, the benefits of data-driven research are limited to healthcare organizations (HCOs) who possess the data \cite{wirth2021privacy, dinov2016methodological}. Due to concerns about patient privacy, HCO stakeholders are reluctant to share patient data \cite{miotto2018deep, kim2021privacy, simon2019assessing}. Access to clinical data is often restricted, or can be prohibitively expensive to obtain, meaning that ML in biomedical research lags behind other areas in AI.

To accelerate the progress of developing AI methods in medicine, one promising alternative is for the data holder to create \textit{synthetic} yet realistic data \cite{jordon2018pate, frid2018gan}. By avoiding ``one-to-one'' mapping to the genuine data compared with data anonymization, synthetic data offers a solution to circumvent the issue of privacy, while the correlations in the original data distributions are preserved for downstream AI applications. There have been successes in the literature using synthetic data to improve AI models where otherwise not possible due to limited availability of resources \cite{jordon2020synthetic, chen2021synthetic, tucker2020generating, el2021evaluating}. For example, large-scale data sharing programs have been demanded for advancing studies related to COVID-19, such as in National COVID Cohort Collaborative (N3C) \cite{n3clink}, and Clinical Practice Research Datalink (CPRD) database in the UK \cite{cprdlink}. 

% The availability of data synthesis facilitates the development of data-driven clinical models in research community for healthcare. However, some of the methods for generating synthetic patient EHRs rely heavily upon hand-crafted rules or clinical expertise \cite{mclachlan2016using, lombardo2008ta}. For example, McLachlan et al. \cite{mclachlan2016using} proposed to generate synthetic EHRs by formalizing the clinical practice guidelines into a state transition machine (STM), with the whole process being domain knowledge-intensive. In addition, since these methods can only handle data with low dimensionality (and are typically disease-specific), the general utility of these methods is severely limited. Recent advances in generative adversarial networks (GANs) \cite{goodfellow2014generative} and their variants offer more efficacious means to generate multidimensional, high-fidelity data with complex correlations for many different applications \cite{kearney2020dosegan, yang2018low, marouf2020realistic}. Furthermore, owing to their flexible tuning techniques, GAN models can better generate samples of anomalous or sparse events, mitigating the issue of the rarity which potentially leads to bias during training downstream ML models \cite{zhou2020sparse, schlegl2019f}.

Recent advances in generative adversarial networks (GANs) \cite{goodfellow2014generative} and their variants offer efficacious means to generate EHRs for a wide range of clinical applications \cite{kearney2020dosegan, yang2018low, marouf2020realistic}. \red{Over the past years, EHR synthesizers have evolved from generating static patient information to producing longitudinal EHR timeseries \cite{esteban2017real, lee2020generating, zhang2021synteg}. 
As longitudinal EHRs contain patient trajectories for describing the underlying health condition, synthesizing such EHR timeseries, therefore, enables new clinical applications related to the status of disease progression \cite{zhang2022keeping}, such as dynamic forecasting of risks, predicting the onset of diseases, and survival analysis based on the time-to-event data.} \red{However, existing studies focus on synthesizing the longitudinal EHRs of a single data type \cite{esteban2017real, yoon2019time, lee2020generating}, whereas the clinical decision-making in real practice includes a variety of information sources in the form of \textit{mixed-type} timeseries. For example, patient physiological signals and laboratory test results are collected in the EHR as \textit{continuous-valued} timeseries, while the medication and diagnostic information are recorded as \textit{discretized-valued} data as binary indicators or categorical ICD codes. Information provided in these mixed-type longitudinal EHRs offer opportunities for more precise and complex clinical analysis.
Furthermore, the predictive power and robustness of the ML models can be boosted by utilizing longitudinal EHR timeseries with various types/sources.}

\red{Existing GANs are limited in simulating mixed-type EHRs due to two reasons. Firstly, it is intrinsically difficult to model the underlying joint distribution of mixed data type timeseries using a single unified framework. Since GANs require the network architectures of the generator and discriminator to be fully differentiable \cite{hjelm2017boundary}, its success is typically limited to generating real-valued, continuous data while facing obstacles for directly generating sequences of discrete tokens, such as ICD codes, that also commonly appear in EHRs. Previous methods \cite{yu2017seqgan, choi2017generating} circumvent this problem by learning representations from the original data which further enables backpropagation in discrete settings,  but there is still a lack of a generative approach for joint modelling of the mixed-type timeseries with heterogeneous nature.}
%While mixed-type EHR timeseries can provide a rich environment for developing ML algorithms for assisting the decision-making of clinicians, they also impose complexity in creating generative models. \red{When extending vanilla-GANs to mixed-type EHR timeseries synthesis, major challenges are found in two aspects. First, as GANs have serious limitations in the data types they can simulate \cite{hjelm2017boundary}, it is hard to accurately model the underlying distribution of timeseries varying in data types in a unified framework. More specifically, since GANs require the network architectures of the generator and discriminator to be fully differentiable \cite{hjelm2017boundary}, its success is typically limited to generating real-valued, continuous data while facing obstacles for directly generating sequences of discrete tokens, such as ICD codes, that also commonly appear in EHRs. Previous methods circumvent this problem by learning representations from the original data which further enables backpropagation in discrete settings \cite{yu2017seqgan, choi2017generating}. But there is still a lack of a generative approach for jointly modelling the mixed-type timeseries with heterogeneous nature.}
\red{Second, although mixed-type clinical timeseries differ in syntax and distributions, they are highly correlated and inform one another of the underlying health of an individual \cite{yu2019inverse, ghassemi2017predicting, wang2019continuous}. It is therefore important to capture the temporal correlations between them when generating the synthetic EHR data.
	For example, the medications (documented in the form of discrete data) prescribed to patients are based on measurements of patients' physiological status (presented as continuous-valued signals). Concurrently, the efficacy of the medical treatments, affect the patient's physiological condition directly. 
	It is therefore critical to accurately capture the temporal correlation between the mixed-type patient trajectories simultaneously to improve clinical decision support.}
%%%such as treatment recommendations based on patients' physiological signals, and early prediction of adverse events based on prior information of various data types.

\red{To address the aforementioned limitations, for the first time, we propose a GAN framework for simultaneously synthesizing mixed-type longitudinal EHR data (denoted as EHR-M-GAN thereafter). 
	Specifically, we focus on generating timeseries in the critical care setting, where the intensive care units (ICU) patients are continuously and closely monitored  (see Fig. \ref{fig_key}a).}
Patient trajectories with high-dimensionality and heterogeneous data types (both continuous-valued and discrete-valued timeseries) are generated while the underlying temporal dependencies are captured. The main contributions of our work are as follows:

\begin{figure*}[ht!]
	\centering
	\includegraphics[width=1.\textwidth]{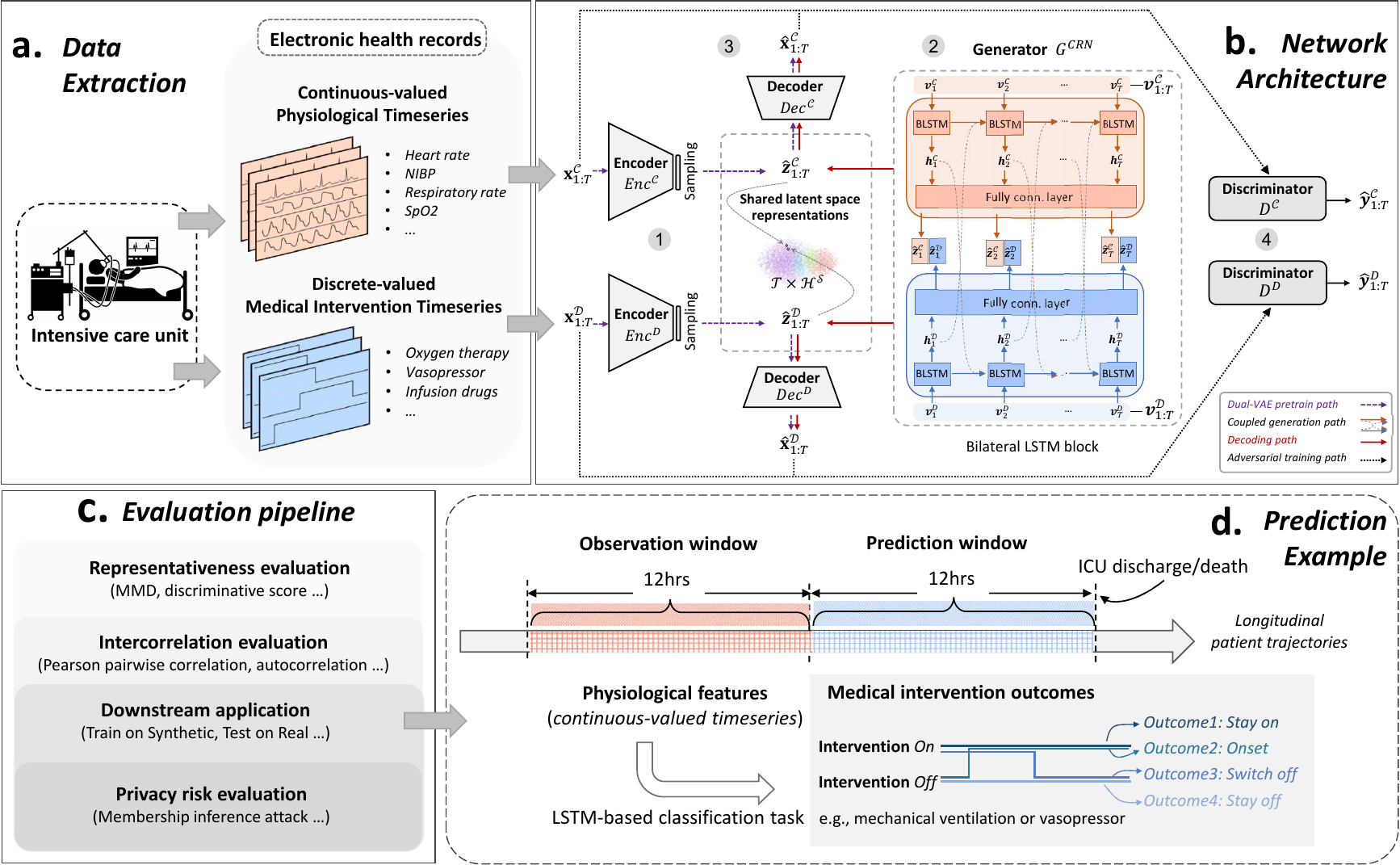}
	\caption{\textbf{Overall schematics.} 
		\textbf{a. Data extraction.} Electronic health records (EHRs) data of mixed-type are routinely collected for patients in intensive care units (ICUs). 
		\textbf{b. Network architecture.} EHR-M-GAN contains two key components --- \textit{Dual-VAE} and \textit{Coupled Recurrent Network} (CRN). 
		\textbf{\textit{Step 1}}: \red{\textit{Dual-VAE} is first pretrained for mapping heterogeneous data ($\mathbf{x}_{t}^{c}, \mathbf{x}_{t}^{d}$) into shared latent representations (${\mathbf{z}}_{t}^{c}, {\mathbf{z}}_{t}^{d}$)}. Multiple objective loss constraints are used to bridge the domain/distribution gap. The training process for Step 1 is indicated in the \textit{Dual-VAE pretrain path} (dashed purple line). \textbf{\textit{Step 2}}: Then, a CRN is established as the generator based on the parallel bilateral LSTM block, \red{which takes the random noise vectors ($\bm{\upsilon}_{t}^{c}, \bm{\upsilon}_{t}^{d}$) as inputs} (see the \textit{Coupled generation path}). \textbf{\textit{Step 3}}: \red{The synthetic latent representations ($\hat{\mathbf{z}}_{t}^{c}, \hat{\mathbf{z}}_{t}^{d}$) provided by CRN are then decoded into synthetic samples ($\hat{\mathbf{x}}_{t}^{c}, \hat{\mathbf{x}}_{t}^{d}$) using the pretrained decoder in Dual-VAE}, which is indicated in the \textit{Decoding path} (solid red line).
		\textbf{\textit{Step 4}}: Finally, the adversarial loss is derived from the discriminators and backpropagated to update the network, which is indicated in the \textit{Adversarial training path} (dotted black line).
		\textbf{c. Evaluation pipeline.} The pipeline includes metrics for evaluating the synthetic data.
		\textbf{d. Prediction example.} 
		Data within 24-hours prior to the patient's endpoints in the ICU (discharge or mortality) is extracted. Both the observation window and prediction window are fixed as 12 hours. The classification task is to use patients' continuous-valued physiological measurements within the observation window as input, to predict the forthcoming discrete-valued medical intervention status in the prediction window. The four outcomes of the intervention status can be categorized as follows: 
		\textit{\textbf{Stay On}}: The intervention begins with \textit{on} and \textit{stays on} within the prediction window; 
		\textit{\textbf{Onset}}: The intervention begins with \textit{off} and is \textit{turned on} within the prediction window; 
		\textit{\textbf{Switch off}}: The intervention begins with \textit{on} and is \textit{stopped} within the prediction window; 
		\textit{\textbf{Stay Off}}: The intervention begins with \textit{off} and \textit{stays off} within the prediction window. } \label{fig_key}
\end{figure*}

\begin{enumerate}[leftmargin=0.5cm]

	\item A novel GAN model entitled EHR-M-GAN is proposed for simultaneously generating mixed-type multivariate EHR timeseries with high fidelity, and overcoming the challenges when extending GANs into the mixed-type data settings (see Fig. \ref{fig_key}b). First, to jointly model the underlying distributions of the heterogeneous features, EHR-M-GAN first maps data from different observational spaces into a reversible, lower-dimensional, shared latent space through a \textit{dual variational autoencoder} (dual-VAE). Then, to capture the correlated temporal dynamics of the mixed-type data, a sequentially coupled generator that is built upon a \textit{coupled recurrent network} (CRN) is employed. In addition, a conditional version of our model --- EHR-M-GAN$_\texttt{cond}$ --- is also implemented, which is capable of synthesizing condition-specific EHR patient data, such as those result in \textit{ICU mortality} or \textit{hospital readmission}. The code of our proposed work is publicly available on GitHub\footnote{https://github.com/jli0117/ehrMGAN}.

\item Evaluations are performed based on three publicly available ICU datasets: MIMIC-III \cite{johnson2016mimic}, eICU \cite{pollard2018eicu} and HiRID \cite{yeche2021hirid} from a total of 141,488 patients. Standardized preprocessing pipelines are applied for the three ICU datasets to provide generalizable machine learning benchmarks. The code for the end-to-end preprocessing pipelines is also available on GitHub\footnote{https://github.com/jli0117/preprocessing\_physionet}.

\item Our EHR-M-GAN outperforms the state-of-the-art benchmarks on a diverse spectrum of evaluation metrics. When compared to real EHR data, both qualitative and quantitative metrics are used to assess the representativeness of the mixed-type data and their inter-dependencies. We further demonstrate the advantages offered by EHR-M-GAN in augmenting clinical timeseries for downstream tasks under various clinical scenarios.

\item 
In the evaluation of privacy risks, we perform an empirical analysis on EHR-M-GAN based on membership inference attack \cite{shokri2017membership}. We then further evaluate the performance of EHR-M-GAN under the framework of differential privacy for its application in downstream task \cite{dwork2006differential}.

\end{enumerate}

\section{Methods}
In this section, we first formulate the problem based on the \textit{mixed-type} temporal EHR data and its corresponding mathematical notation. \red{Then, we discuss the challenges of synthesizing mixed-type EHR timeseries and the intuition behind the proposed model. Finally, we introduce the proposed EHR-M-GAN model in detail.}

\subsection{Problem formulation}
The longitudinal patient EHR dataset is denoted as $\smash{\mathscr{D} = \{(\mathbf{x}_{i, 1:T_i})\}_{i=1}^{N}}$, with each record (e.g., individual patient) being indexed by $\smash{i\in\{1, 2, ... N\}}$. Here we consider the \textit{i}-th instance tuple $\smash{\mathbf{x}_{i, 1:T_i} = \{\mathbf{x}_{i, 1:T_i}^{\mathcal{C}}, \mathbf{x}_{i, 1:T_i}^{\mathcal{D}} \}}$ consists of two components (i.e., two types of data). Let
$\smash{\mathbf{x}_{i, 1:T_i}^{\mathcal{C}}} \in \mathbb{R}^{|J|} $ denote the $|J|$-dimensional continuous-valued timeseries, such as physiological signals from real-time bedside monitors. And $\smash{\mathbf{x}_{i, 1:T_i}^{\mathcal{D}}} \in \mathbb{Z}^{|K|} $ denotes the $|K|$-dimensional discrete-valued timeseries, such as life-support interventions with the categorical value indicate its status (presence or absence).

\subsection{Challenges in mixed-type timeseries generation}
\red{There are two main challenges when synthesizing mixed-type EHR timeseries.
	First, GANs have serious limitations on the type of data they can model \cite{hjelm2017boundary}. Specifically, as GANs require generators and discriminators to be both fully differentiable, generating discrete-valued timeseries using traditional GANs architectures would raise problems during backpropagation as no direct gradient can be provided \cite{choi2017generating, yu2017seqgan}. Therefore, it is intrinsically difficult to model the underlying joint distribution of mixed data type timeseries using a single unified framework.
	Second, as the mixed-type timeseries are correlated (such as correlations between ICU patients’ physiological signals and treatment status in the critical care setting), it is therefore important to model the interdependencies among heterogeneous types of timeseries.}

\subsection{Intuition behind EHR-M-GAN} 
\red{First, to jointly model the distribution of continuous-valued and discrete-valued timeseries using GANs, we build the generative model based on the latent space encoded by VAE networks. Instead of directly synthesizing discrete-valued timeseries that deactivate the backpropagation in GANs, the generator first synthesizes latent representations that allow the direct gradient in the network, therefore satisfying the prerequisite for GANs architecture to be fully differentiable. The synthetic latent representations for both types of data can be further decoded into raw timeseries using the decoders in VAEs.
	
	Even though the aforementioned network architectures enable the joint modelling of mixed-type data distribution, it still lacks the capability of capturing the inter-dependencies in heterogeneous data. In order to address the second issue, we devised \textit{dual-VAE} module for pretraining step and \textit{sequentially coupled generator} module for generation step. The \textit{dual-VAE} incorporates multiple loss constraints, which were previously adopted in domains such as self-supervised learning (SSL), timeseries representation learning, and domain adaptation (DA), to extract useful hierarchical representations from heterogeneous but correlated data types. The \textit{sequentially coupled generator} module replaces the traditional LSTM cell with the novel bilateral LSTM (BLSTM) cell we propose, where the “communication” of the two types of information are introduced into the networks. Therefore, the temporal dynamics between the mixed-type data can be preserved during the iteration.}

\subsection{Proposed model}
As illustrated above, EHR-M-GAN can be factorized into two key components (see Fig. \ref{fig_key}b): (1) a \textit{dual-VAE} framework for learning the shared latent space representations; (2) an RNN-based \textit{sequentially coupled generator} and its corresponding sequence discriminators. 

As shown in Fig. \ref{fig_key}b, during the \textit{pretrain} stage, both continuous-valued and discrete-valued temporal trajectories are first jointly mapped into a shared latent space using the \textit{dual-VAE} component (Step 1). Then, the \textit{sequentially coupled generator} in EHR-M-GAN produces the synthetic latent representations (Step 2), which further can be recovered into features in the observational space by the pretrained decoders in the \textit{dual-VAE} (Step 3). Finally, the adversarial loss is provided based on discriminative results and backpropagated to update the network (Step 4). The following sections discuss them in turn.

\subsubsection{Dual-VAE pretraining for shared latent space representations}
\red{One premise of successfully training EHR-M-GAN to generate reversible latent codes is to meet the assumption that for the \textit{same} patient indexed with $i$, both $\smash{\mathbf{x}_{i, 1:T_i}^{\mathcal{C}}}$ and $\smash{\mathbf{x}_{i, 1:T_i}^{\mathcal{D}}}$ can be encoded into the \textit{same} latent space $\smash{\mathcal{H}^{\mathcal{S}} \subset
		\mathbb{R}^{|S|}}$, where $|S|$ denotes its spatial dimension.} For the sake of simplicity, the subscripts $i$ are omitted throughout most of the paper. 
To achieve this, we propose to use a \textit{dual-VAE} framework, which exploits two VAE networks to encode both continuous and discrete multivariate timeseries into dense representations within $\smash{\mathcal{H}^{\mathcal{S}}}$ based on multiple constraints. 

Fig. S2 (see Section S.1.C in Supplementary materials) diagrams the details of the proposed \textit{dual-VAE} framework for learning the shared latent representations. We start with training two encoders, i.e.,
$\smash{\mathit{Enc}^{\mathcal{C}}}$: $\smash{ \phi_{\mathcal{T} \times \mathcal{X}^{\mathcal{C}}} \to 
	\phi_{\mathcal{T} \times \mathcal{H}^{\mathcal{S}}}}$ and $\smash{\mathit{Enc}^{\mathcal{D}}}$: $\smash{\phi_{\mathcal{T} \times \mathcal{X}^{\mathcal{D}}} \to \phi_{\mathcal{T} \times \mathcal{H}^{\mathcal{S}}}}$, with the embedding functions:
\begin{equation}\label{eq1}
	\mathbf{z}_{1:T}^{\mathcal{C}}= 
	\mathit{Enc}^{\mathcal{C}}(\mathbf{x}_{1:T}^{\mathcal{C}})
	\quad \quad
	\mathbf{z}_{1:T}^{\mathcal{D}}=
	\mathit{Enc}^{\mathcal{D}}(\mathbf{x}_{1:T}^{\mathcal{D}})
\end{equation}

After passing data from $\smash{\mathcal{X}^{\mathcal{C}}}$ and $\smash{\mathcal{X}^{\mathcal{D}}}$ through two encoders, a pair of embedding vectors $\smash{(\mathbf{z}^{\mathcal{C}}_{1:T}, \mathbf{z}^{\mathcal{D}}_{1:T})}$ in the shared latent space $\smash{\mathcal{H}^{\mathcal{S}}}$ can be obtained.
Then the decoders for both domains $\smash{\mathit{Dec}^{\mathcal{C}}: \psi_{\mathcal{T} \times \mathcal{H}^{\mathcal{S}}}
	\to 
	\psi_{\mathcal{T} \times \mathcal{X}^{\mathcal{C}}}}$ and $\smash{\mathit{Dec}^{\mathcal{D}}: 
	\psi_{\mathcal{T} \times \mathcal{H}^{\mathcal{S}}}
	\to 
	\psi_{\mathcal{T} \times \mathcal{X}^{\mathcal{D}}}}$ further reconstruct features based on the latent embeddings using mapping functions that operate in the opposite direction: 
\begin{equation}
	\tilde{\mathbf{x}}_{1:T}^{\mathcal{C}}= 
	\mathit{Dec}^{\mathcal{C}}(\mathbf{z}_{1:T}^{\mathcal{C}})
	\quad \quad
	\tilde{\mathbf{x}}_{1:T}^{\mathcal{D}} = 
	\mathit{Dec}^{\mathcal{D}}(\mathbf{z}_{1:T}^{\mathcal{D}})
\end{equation}

\red{Also, to incentivize dual-VAE to better bridge the gap between domains of mixed-type timeseries, we enforce a weight-sharing constraint \cite{liu2017unsupervised, liu2016coupled} within specific layers of both the encoders pairs and the decoders pairs (See Section S.1.B for details).}

In the following subsections, we define multiple loss constraints for the optimization of \textit{dual-VAE}, including \textit{ELBO loss}, \textit{matching loss}, \textit{contrastive loss}, as well as \textit{semantic loss} for EHR-M-GAN$_{\texttt{cond}}$. 
\red{Among these losses, \textit{ELBO loss} ensures that the mixed-type timeseries can be successfully reconstructed after being encoded into latent representations. The \textit{matching loss} ensures that heterogeneous types of features from a single patient share contexts during representation learning (instance-wise). The goal of \textit{contrastive loss} is to ensure that patients with similar trajectories stay close to each other in the latent space (population-wise). And \textit{semantic loss} used in EHR-M-GAN$_{\texttt{cond}}$ encourages patients with the same conditional labels (e.g., outcomes) to share similar latent representations.}
Intuitions and descriptions behind the objectives are discussed in turn.

% loss1 - VAE loss
{\setlength{\parindent}{0pt}\vspace{6pt}
	\textBF{Evidence Lower Bound (ELBO).} We first incorporate the standard VAE loss, with the optimization objective as the evidence lower bound (ELBO). VAE holds the assumption of spherical Gaussian prior for the distribution of latent embeddings, where features can then be reconstructed by sampling from that space. The re-parameterization tricks enable differentiable stochastic sampling and network optimization. For encoder and decoder in the dual-VAE for domain $d \in \{\mathcal{C}, \mathcal{D}\}$, the objective function is defined as:
	\begin{equation}\label{eq:elbo}
		\begin{split}
			\mathcal{L}^{\mathrm{ELBO}}_{d}=
			-\mathbb{E}_{q_{\phi}(\mathbf{z}|\mathbf{x})}[\log p_{\psi}&(\mathbf{x} | \mathbf{z})] + \\
			\beta_{\mathrm{KL}}  D_{\mathrm{KL}}(q_{\phi}&(\mathbf{z} | \mathbf{x}) \| p_{\psi}(\mathbf{z}))
		\end{split}
	\end{equation}
	where $ \mathbf{z} \sim \mathit{Enc}(\mathbf{x}) \triangleq q_{\phi}(\mathbf{z}|\mathbf{x}),
	\tilde{\mathbf{x}} \sim \mathit{Dec}(\mathbf{z}) \triangleq p_{\psi}(\mathbf{x}|\mathbf{z}) $, and $D_{\mathrm{KL}}$ is the Kullback-Leibler divergence. The first term in Eq. (\ref{eq:elbo}) is the expected log-likelihood term that penalizes the deviations in reconstructing the inputs, while the second term of KL-divergence is the regularization imposed over the latent distribution from its Gaussian prior (normally chosen to be $\mathcal{N}(\mathbf{0}, \boldsymbol{I})$). $\beta_{\mathrm{KL}}$ is the hyperparameter for balancing the weights between two terms.} 

% loss2 - matching loss
{\setlength{\parindent}{0pt}\vspace{6pt}
	\textBF{Matching loss.}
	Typically, representations derived from the \textit{same} patient are assumed to capture the shared context. Therefore, embedding vectors $\smash{(\mathbf{z}^{\mathcal{C}}_{i, 1:T_i}, \mathbf{z}^{\mathcal{D}}_{i, 1:T_i})}$ projected from the \textit{same} patient $i$, are supposed to be positioned closely in the shared latent space (See Fig. S2 in Supplementary materials). 
	\red{Therefore, in this study, we borrow the concept of matching loss from domain alignment in DA, which enables efficient representation learning crossing domains/modalities \cite{wan2020old}.}
	In this study, the matching loss ensures that low-dimensional latent space can be shared between heterogeneous features. Hence, the pairwise matching loss is incorporated to motivate the encoders to minimize the distance within the corresponding representation pairs. In the low-dimensional Euclidean space, we optimize the network by using the following objective: 
	\begin{equation}
		\mathcal{L}^{\mathrm{Match}}=
		\mathbb{E}_
		{\mathbf{z} \sim p_{\mathbf{z}}		
		}[
		\sum_{t \in \mathcal{T}}
		||
		\mathbf{z}^{\mathcal{C}}_{t}
		-
		\mathbf{z}^{\mathcal{D}}_{t} ||^{2}]
	\end{equation}
	The pairwise matching loss achieve its optimal when the distance proxy $\mathcal{L}^{\mathrm{Match}}$ becomes zero. 
	
	% loss3 - contrastive loss
	{\setlength{\parindent}{0pt}\vspace{6pt}\textBF{Contrastive loss. }On the flip side, pairwise reconstruction error (i.e., intra-correlations within one instance) measured by \textit{matching loss} neglects the commonalities present across patients (inter-correlations of data) \cite{kiyasseh2021clocs}. In order to guarantee sufficient bound for representation learning, we incorporate \textit{contrastive loss} as another distance metric to capture the inter-correlations among the population.}
	
	\setlength\parindent{10pt} Contrastive learning is a concept that has recently been popularized in self-supervised learning (SSL) \cite{liu2021self}, which aims to capture intrinsic patterns from input data without human annotations. In this study, we instantiate the contrastive loss via \textit{NT-Xent}, which is proposed by Chen et al. in their work SimCLR \cite{chen2020simple}. 
	\red{The core of contrastive learning is to encourage networks to attract positive pairs closer and repulse negative pairs apart in the latent space. 
		In this study, we adapt the corresponding auxiliary tasks for calculating contrastive loss to the scenario of learning representations from mixed-type timeseries. The objective of the task is to determine whether a set of representations transformed from the observational space belong to the \textit{same} patient.} And this leads to the corresponding positive pairs (true) and negative pairs (false).
	
	For patient data of $N$ records,  we can obtain $N$ pairs of latent representations from the encoders in \textit{dual-VAE}. For patient indexed with $i$, $\mathbf{h}_i^{\mathcal{C}}$ and $\mathbf{h}_i^{\mathcal{D}}$ denotes the 
	embeddings derived from the continuous-valued and discrete-valued observational space, respectively. Due to the symmetric architecture of \textit{dual-VAE}, here we use $d$ and $d^{\prime}$ to represent one of each different domain, i.e., $d, d^{\prime} \in \{\mathcal{C}, \mathcal{D} \} $ and $d \neq d^{\prime}$. Therefore, the positive pairs for patient $i$ can be referred as $(i^{d}, i^{d^{\prime}})$, while the other $2(N-1)$ samples are regarded as negative pairs. Then the contrastive loss for a positive pair $(i^{d}, i^{d^{\prime}})$ is defined as:
	\begin{equation}
		\resizebox{0.6\hsize}{!}{$
			\mathcal{L}_{i^{d}, i^{d^{\prime}}}^{\mathrm{Contra}}=-\log \frac{\exp \left(\operatorname{sim}\left(\mathbf{h}_{i^{d}}, \mathbf{h}_{i^{d^{\prime}}}\right) / \tau\right)}{\sum_{i^{dd^{\prime}}=1}^{2 N} \mathbb{1}_{[{i^{dd^{\prime}}} \neq i^{d}]} \exp \left(\operatorname{sim}\left(\mathbf{h}_{i^{d}}, \mathbf{h}_{{i^{dd^{\prime}}}}\right) / \tau\right)}
			$}
	\end{equation}
	where $\operatorname{sim}(u, v)=u^{T} v /\|u\|\|v\|$ denotes the cosine similarity between two vectors.
	$\tau>0$ denotes a temperature hyperparameter. $\mathbb{1}_{[n \neq m]} \in\{0,1\}$ is an indicator evaluating to 1 iff $n \neq m$. And $i^{dd^{\prime}} \in \{1, 2, ..., 2N\}$ represents the index of latent embeddings from \textit{both} data types.
	The final contrastive loss is computed across the total number of $|i^{d}-i^{d^{\prime}}|=N$ positive pairs for both $(i^{d}, i^{d^{\prime}})$ and $(i^{d^{\prime}}, i^{d})$, and is defined as: 
	\begin{equation}
		\mathcal{L}^{\mathrm{Contra}}=\frac{1}{2 N} \sum^{N}_{i^{d}=1}
		\sum^{N}_{i^{d^{\prime}}=1}
		[\mathcal{L}_{i^{d}, i^{d^{\prime}}}^{\mathrm{Contra}}+\mathcal{L}_{i^{d^{\prime}}, i^{d}}^{\mathrm{Contra}}]
	\end{equation}
	
	% loss4 - semantic loss - optional for conditional version
	{\setlength{\parindent}{0pt}\vspace{2pt}
		\textBF{Semantic loss.} In EHR-M-GAN$_{\mathtt{cond}}$, semantic loss is imposed to better align patients with same labels (conditions) into the same latent space clusters. For example, if the label of \textit{severe clinical deterioration} in the ICU is given for conditional data generation, the corresponding synthetic continuous-valued timeseries (e.g., severely deranged vital signs) is supposed to be strongly associated with the discrete-valued timeseries (e.g., intensive medical interventions) under the same label. For both domains, additional linear classifiers are trained to classify the latent embeddings based on their corresponding conditions in the observational space. We implement logistic regression as the linear classifiers and calculate the cross entropy as the semantic losses for both domains. For $d \in \{\mathcal{C}, \mathcal{D} \}$, given the latent embedding vector $\mathbf{z}^{d}$ and the conditional information vector $\mathbf{y}$:
		\begin{equation}
			\mathcal{L}_{d}^{\mathrm{Class}}=\mathbb{E}_{\mathbf{z}^d \in 
				\mathcal{H}^{\mathcal{S}}
			} \mathrm{CE}\left(f^d_{\mathtt{linear}}(\mathbf{z}^{d}), 
			\mathbf{y}
			\right)
		\end{equation}
		where $f^d_{\mathtt{linear}}$ denotes the linear classifier for the corresponding domain. And $\mathrm{CE} = - \sum_{j} y_{j} \log \left(\widehat{y}_{j}\right),  \; (j = 1, 2, ..., |L|)$ denotes the cross entropy loss, where $\hat{y}_j$ is the output of the linear classifier, and $y_j$ is the ground truth label for class $j$ in condition vector $\mathbf{y} $.}
	
	% sum up
	In summary, to train the \textit{dual-VAE} for learning the shared latent space representation, the total objective function for $d \in \{\mathcal{C}, \mathcal{D} \}$ is:
	\begin{equation}\label{eq:vae1}
		\mathcal{L}_{d} = \beta_0 \mathcal{L}^{\mathrm{ELBO}}_{d} + \beta_1 \mathcal{L}^{\mathrm{Match}} + \beta_2 \mathcal{L}^{\mathrm{Contra}}
	\end{equation}
	Under the conditional learning scenario of EHR-M-GAN$_{\mathtt{cond}}$, the total loss becomes:
	\begin{equation}\label{eq:vae2}
		\mathcal{L}_{d} = \beta_0\mathcal{L}^{\mathrm{ELBO}}_{d} + \beta_1\mathcal{L}^{\mathrm{Match}} + \beta_2\mathcal{L}^{\mathrm{Contra}} + \beta_3\mathcal{L}_{d}^{\mathrm{Class}}
	\end{equation}
	where $\beta_0$, $\beta_1$, $\beta_2$, and $\beta_3$ are scalar loss weights used to balance the loss terms. 
	
	To validate the effectiveness of multiple losses and the weight-sharing constraint in the proposed \textit{dual-VAE} network, we perform the corresponding ablation study using MIMIC-III dataset as an example. The results can be found in S.3.B in the Supplementary materials. As shown in Table S7, all the components in the proposed \textit{dual-VAE} network contribute to the improvement of EHR-M-GAN's performance when generating mixed-type timeseries data.
	
	\subsubsection*{Sequentially coupled generator based on CRN}
	\red{We propose the \textit{sequentially coupled generator} for generating latent representations for mixed-type timeseries, which is built based on the network architecture of \textit{coupled recurrent network (CRN)}. 
		Specifically, a CRN exploits bilateral long short-term memory (BLSTM) cells as its recurrent layer to preserve the temporal dependencies between the continuous and discrete-valued sequences. 
		The novel network architecture of bilateral-LSTM we proposed can extract and transmit the correlations between the mixed-type timeseries, as opposed to vanilla-LSTM which has only one recursive connection.} In the following section, we first discuss the structure of BLSTM in detail as its essential recurrent layer of CRN, and then build the \textit{sequentially coupled generator} based on CRN.
	
	{\setlength{\parindent}{0pt}\vspace{6pt}
		\textBF{Bilateral long short-term memory.}
		\red{As the traditional LSTM only considers temporal dynamics from single-type timeseries, therefore is incapable to extract and transmit temporal correlation from heterogeneous features. Therefore, we propose the novel bilateral-LSTM cell with two network connections to characterize the correlations between two types of data.}}
	\red{Given $d, d^{\prime} \in \{\mathcal{C}, \mathcal{D} \}$, $\bm{\upsilon}^{d}_{t}$ and $\mathbf{h}^{d}_{t}$  denotes the input vector (i.e., the random noise during GANs' training) and hidden state vector for domain $d$ at time step $t$, respectively.}
	An additional set of weights for introducing hidden states representations $\mathbf{h}^{d^{\prime}}_{t}$ from domain $d^{\prime}$ is included when updating the input gate $\mathbf{i}^{d}_{t}$, forget gate $\mathbf{f}^{d}_{t}$, output gate $\mathbf{o}^{d}_{t}$, and cell memory $\tilde{\mathbf{c}}^{d}_{t}$. The state transition functions for BLSTM are:
	\begin{equation}\label{eqn:blstm}
		\resizebox{0.6\hsize}{!}{$
			\begin{aligned}
				\mathbf{i}^{d}_{t} &=\sigma\left(\mathbf{W}_{i d {\upsilon}}\bm{\upsilon}^{d}_{t}+\mathbf{W}_{i d h^{d^{\prime}}} \mathbf{h}^{d^{\prime}}_{t-1}+\mathbf{W}_{i d h^{d}} \mathbf{h}^{d}_{t-1}+\mathbf{b}_{i d}\right) \\
				\mathbf{f}^{d}_{t} &=\sigma\left(\mathbf{W}_{f d {\upsilon}}\bm{\upsilon}^{d}_{t}+\mathbf{W}_{f d h^{d^{\prime}}} \mathbf{h}^{d^{\prime}}_{t-1}+\mathbf{W}_{f d h^{d}} \mathbf{h}^{d}_{t-1}+\mathbf{b}_{fd}\right) \\
				\mathbf{o}^{d}_{t} &=\sigma\left(\mathbf{W}_{o d {\upsilon}}\bm{\upsilon}^{d}_{t}+\mathbf{W}_{o d h^{d^{\prime}}}  \mathbf{h}^{d^{\prime}}_{t-1}+\mathbf{W}_{o d h^{d}}  \mathbf{h}^{d}_{t-1}+\mathbf{b}_{o d}\right) \\
				\tilde{\mathbf{c}}^{d}_{t} 
				&=\tanh \left(\mathbf{W}_{c d {\upsilon}}\bm{\upsilon}^{d}_{t}+\mathbf{W}_{c d h^{d^{\prime}}} \mathbf{h}^{d^{\prime}}_{t-1}+\mathbf{W}_{c d h^{d}} \mathbf{h}^{d}_{t-1}+\mathbf{b}_{c d}\right) \\
				\mathbf{c}^{d}_{t} &=\mathbf{f}^{d}_{t} \odot \mathbf{c}^{d}_{t-1}+\mathbf{i}^{d}_{t} \odot \tilde{\mathbf{c}}^{d}_{t} \\
				\mathbf{h}^{d}_{t} &=\mathbf{o}^{d}_{t} \odot \tanh \left(\mathbf{c}^{d}_{t}\right)
			\end{aligned}
			$}
	\end{equation}
	
	\red{As indicated by Eq. \ref{eqn:blstm}, the proposed BLSTM network overcomes the limitation of vanilla-LSTM network on modelling the correlation between the mixed-type timeseries by establishing the supplemental recursive connection. The new connection facilitates the model to intrinsically decide how much information it should pass through the gates from its counterpart. A diagram of the BLSTM cell in contrast to vanilla-LSTM cell can be found in the Supplementary materials (see Fig. S3).}
	
	{\setlength{\parindent}{0pt}\vspace{6pt}
		\textBF{Coupled recurrent network.}
		The architecture of CRN consists of three layers: the \textit{input layers}, the \textit{recurrent layers}, and the \textit{fully connected layers}. First, the random noise vectors $\smash{\bm{\upsilon}_{t}^{d}}$ and $\smash{\bm{\upsilon}_{t}^{d^{\prime}}}$ for two domains, \red{which are sampled from uniform distributions (i.e., $\smash{\bm{\upsilon}_{t}^{d}}, \smash{\bm{\upsilon}_{t}^{d^{\prime}}} \in \mathcal{U}(0,1)$), are separately fed into the \textit{input layers}.} 
		% As shown in Fig. \ref{fig_key}, two streams of BLSTM entangles with each other in a CRN, and are jointly trained to sequentially exploit the correlations between two types of trajectories. 
		Then the \textit{recurrent layers} $\smash{f_{\mathtt{rec}}}$, which are built based on two streams of BLSTM, one for each data type, are used to recursively iterate hidden states from both branches. Finally, the \textit{fully connected layers} $f_{\mathtt{conn}}^{d}$ and $f_{\mathtt{conn}}^{d^{\prime}}$ produce the generated latent vectors $\smash{\hat{\mathbf{z}}_{t}^{d}}$ and $\smash{\hat{\mathbf{z}}_{t}^{d^{\prime}}}$ for the decoding stage in \textit{dual-VAE}. At time step $t$, CRN can be formulated as:}
	\begin{equation}
		\begin{aligned}
			(\mathbf{h}_{t}^{d}, \mathbf{h}_{t}^{d^{\prime}})
			&= f_{\mathtt{rec}}(
			(\bm{\upsilon}_{t}^{d}, \bm{\upsilon}_{t}^{d^{\prime}}), 
			(\mathbf{h}_{t-1}^{d}, \mathbf{h}_{t-1}^{d^{\prime}})
			)\\
			\hat{\mathbf{z}}_{t}^{d}&= f^{d}_{\mathtt{conn}}(\mathbf{h}_{t}^{d})\\
			\hat{\mathbf{z}}_{t}^{d^{\prime}} &=  
			f^{d^{\prime}}_{\mathtt{conn}}(\mathbf{h}_{t}^{d^{\prime}})
		\end{aligned}
	\end{equation}
	
	\red{In summary, heterogeneous timeseries that exhibits mutual influence on each other are integrated into CRN to model their interdependencies. By exploiting the BLSTM cell as its recurrent layer,} two streams of the inputs in the generator are encouraged to ``communicate'' with each other. CRN is therefore capable of exploiting the interplay between mixed-type data that correlates over time. 
	
	\subsubsection*{Joint training and optimization}
	The overall architecture of EHR-M-GAN is shown in Fig. \ref{fig_key}. In this section, we give a detailed description of the entire network's structure and the optimization objective of the model. The steps for the training and optimization of EHR-M-GAN are as follows:
	
	\begin{itemize}
		
		\item The \textit{pretraining of dual-VAE}: 
		\red{First, a dual-VAE network which consists of a pair of encoders ($\mathit{Enc}^{\mathcal{C}}, \mathit{Enc}^{\mathcal{D}}$) and decoders ($\mathit{Dec}^{\mathcal{C}}, \mathit{Dec}^{\mathcal{D}}$) is pretrained with both continuous and discrete data.}
		Based on multiple objective constraints in Eq. \ref{eq:vae1} (for EHR-M-GAN$_{\texttt{cond}}$ the objective function can be referred in Eq. \ref{eq:vae2}), a shared latent space is learnt using \textit{dual-VAE}, where the gap between the embedding representations $( \mathbf{z}_{1:T}^{\mathcal{C}}, \mathbf{z}_{1:T}^{\mathcal{D}} )$ from both domains is minimized. 
		
		\item The \textit{generation of latent representations based on CRN}: Then, during the joint training stage, the \textit{sequentially coupled generator} \red{which is built based on CRN, takes the random noise vector $( \hat{\mathbf{z}}_{1:T}^{\mathcal{C}}, \hat{\mathbf{z}}_{1:T}^{\mathcal{D}} )$ as inputs and  iterating across the timesteps $t \in \{1, 2, ..., T\}$ by the internal transition functions.} 
		Therefore, the synthetic latent embedding representations $( \hat{\mathbf{z}}_{1:T}^{\mathcal{C}}, \hat{\mathbf{z}}_{1:T}^{\mathcal{D}} )$ for both continuous and discrete data can be obtained.
		%Therefore, the synthetic latent embedding representations $( \hat{\mathbf{z}}_{1:T}^{\mathcal{C}}, \hat{\mathbf{z}}_{1:T}^{\mathcal{D}} )$ for both types of data can be obtained. 
		
		\item The \textit{decoding for the mixed-type timeseries}: Next, the generated latent embeddings $( \hat{\mathbf{z}}_{1:T}^{\mathcal{C}},
		\hat{\mathbf{z}}_{1:T}^{\mathcal{D}} )$ are further fed into the pretrained decoders ($\mathit{Dec}^{\mathcal{C}}, \mathit{Dec}^{\mathcal{D}}$) and decoded into the corresponding synthetic patient trajectories $( \hat{\mathbf{x}}_{1:T}^{\mathcal{C}},
		\hat{\mathbf{x}}_{1:T}^{\mathcal{D}} )$ in the observational space. 
		
		\item The \textit{adversarial loss update based on the discriminators}: Finally, the adversarial loss can be calculated from the LSTM network-based discriminators $D^{\mathcal{C}}$ and $D^{\mathcal{D}}$ by distinguishing between the real samples and synthetic timeseries for both data types.
		%...by distinguishing between the real samples and synthetic timeseries. 
		
	\end{itemize}
	
The mathematical expression for the min-max objectives in EHR-M-GAN is provided as follows:
\begin{equation}
		\resizebox{0.55\hsize}{!}{$
			\begin{split}
				\min _{G} \max _{D} &V_{\textrm{EHR-M-GAN}} = \\
				&\mathbb{E}_	{\mathbf{x} \sim p_{{\mathbf{x}}}}
				[\log D^{\mathcal{C}}(\mathbf{x}^{\mathcal{C}} ) 
				+ \log D^{\mathcal{D}}(\mathbf{x}^{\mathcal{D}})] + \\
				&\mathbb{E}_{\bm{\upsilon} \sim p_{\bm{\upsilon}}}
				[\log (1-D^{\mathcal{C}}(\hat{\mathbf{x}}^{\mathcal{C}}))
				+ \log (1-D^{\mathcal{D}}(\hat{\mathbf{x}}^{\mathcal{D}}))]
			\end{split}
			$}
\end{equation}
	
\subsection{Conditional version of EHR-M-GAN}
For the conditional extension of EHR-M-GAN$_\texttt{cond}$, the auxiliary label information is first used during the pretraining step of \textit{dual-VAE}. Both the encoders and decoders condition on the auxiliary (one-hot) labels from $\mathcal{L}$, to make the model better adapted to particular contexts. In \textit{dual-VAE}, the additional semantic loss is also incorporated during the optimization for the shared latent space (see Eq. \ref{eq:vae2}). 
Meanwhile, the same conditional labels are also applied in the sequentially coupled generator and discriminators, where the classes are fed as additional inputs through concatenation, as in the original CGAN architecture proposed by Mirza et al \mbox{\cite{mirza2014conditional}}. 
	
\red{The t-SNE visualisation of the latent embeddings induced from \textit{dual-VAE} can be found in Supplementary materials (see Section S.4.C), which indicates that the conditional information carried into EHR-M-GAN$_\texttt{cond}$ can be beneficial when synthesizing patient trajectories under specific medical conditions.}
Overall, the adversarial loss for EHR-M-GAN$_\texttt{cond}$ can be denoted as follows:
\begin{equation}
		\resizebox{0.58\hsize}{!}{$
			\begin{split}
				\min _{G} \max _{D} &V_{\textrm{EHR-M-GAN$_{\texttt{cond}}$}} = \\ 
				&\mathbb{E}_{\mathbf{y},\mathbf{x} \sim p_{\mathbf{y}, \mathbf{x} }}
				[\log D^{\mathcal{C}}(\mathbf{x}^{\mathcal{C}} | \mathbf{y}) +
				\log D^{\mathcal{D}}(\mathbf{x}^{\mathcal{D}} | \mathbf{y}) ] + \\
				&\mathbb{E}_{\mathbf{y} \sim p_{\mathbf{y}}, \bm{\upsilon} \sim p_{ { \bm{\upsilon} }}} [\log (1-D^{\mathcal{C}}(\hat{\mathbf{x}}^{{\mathcal{C}}} | \mathbf{y}))+ 
				\log (1-D^{\mathcal{D}}(\hat{\mathbf{x}}^{{\mathcal{D}}} | \mathbf{y}))]
			\end{split}
			$}
	\end{equation}
	
The pseudocodes for dual-VAE and EHR-M-GAN are provided in the Supplementary materials (see Section S.1.E).

\section{Data and Evaluation}
\subsection{Dataset Description}
The following three publicly accessible ICU datasets are used for evaluating the performance of EHR-M-GAN in generating the longitudinal EHR data:
\begin{itemize}
	\item \textbf{MIMIC-III} (Medical Information Mart for Intensive Care) \cite{johnson2016mimic} --- a freely accessible database that comprises de-identified EHRs associated with approximately 60,000 ICU admitted patients and 312 million observations to Beth Israel Deaconess Medical Center. 
	
	\item \textbf{eICU} (eICU Collaborative Research Database) \cite{pollard2018eicu} --- a multi-center critical care database containing data for over 200,000 admissions and 827 million observations to ICUs from 208 hospitals located
	throughout the United States. 
	
	\item \textbf{HiRID} (High time-resolution ICU dataset) \cite{yeche2021hirid} --- a high-resolution ICU dataset relating to more than 3 billion observations from almost 34,000 ICU patient admissions, monitored at the Department of Intensive Care Medicine, Bern University Hospital, Switzerland. 
\end{itemize}

All these critical care databases include vital sign measurements, laboratory tests, treatment information, survival records, and other routinely collected data from hospital EHR systems. \red{From these clinical observations, we featurize the patient trajectories as the combination of continuous-valued physiological timeseries (such as  heart rate, oxygen saturation, and measurements from blood gas tests) and discrete-valued medical intervention timeseries (such as the usage of therapeutic devices or intravenous medications). 
	Temporal trajectories \textbf{24 hours prior to patients' ICU endpoints} (discharge or death) are extracted for the three critical care databases.}
Data are preprocessed following an open-source framework --- MIMIC-Extract \cite{wang2020mimic}, \red{where the patients' physiological and intervention signals are hourly aggregated for denser representations.} Details on data curation, including the cohort selection criteria, full list of features, and imputation method, are explained in Supplementary Materials (see S.2 Datasets). Overall, the summarising statistics of the finalised cohorts for three databases are shown in Table \ref{tb_data_descrption}.

\begin{table*}[t!]
	
	\caption{\label{tb_data_descrption} \textbf{Summary of the cohorts after preprocessing on three critical care databases.} Number of patients and ICU admissions, \red{as well as the dimensions of continuous-valued and discrete-valued variables,} are provided for each dataset.     \red{Temporal trajectories 24 hours prior to patients' ICU endpoints are extracted for the three critical care databases.} Note that only the first ICU admission is selected for each patient. The dimension of the continuous- and discrete-valued data are provided. The conditional labels for training EHR-M-GAN$_\texttt{cond}$ and the corresponding counts for each class are also listed.}
	\begin{adjustbox}{width=1\textwidth,center}	
		\begin{tabular}{l c c c c l c} %{l c c c c l c }
			\hline			
			& \makecell[c]{Number of patients} 
			& \makecell[c]{Number of ICU admissions}
			& \makecell[c]{Dimension of continuous\\-valued variables} 
			& \makecell[c]{Dimension of discrete\\ -valued variables} 
			& Conditional labels
			& Counts \\ \hline
			\multirow{4}{*}{MIMIC-III} &
			\multirow{4}{*}{$28,344$} & \multirow{4}{*}{$28,344$} & \multirow{4}{*}{$78$} & \multirow{4}{*}{$20$} & ICU mortality & $1,870 \, (6.59\%)$ \\
			&  &  &  &  & Hospital mortality & $\;\;\; 911 \, (3.21\%)$ \\
			&  &  &  &  & 30-day readmission & $1,122 \, (3.95\%)$ \\
			&  &  &  &  & No 30-day readmission & $24,441 \, (86.22\%)$ \\ \hline
			\multirow{3}{*}{eICU} & \multirow{3}{*}{$99,015$} & \multirow{3}{*}{$99,015$} & \multirow{3}{*}{$55$} & \multirow{3}{*}{$19$} & ICU mortality & $4,500 \, (4.54\%)$ \\
			&  &  &  &  & Hospital mortality & $3,291 \, (3.32\%)$ \\
			&  &  &  &  & Hospital discharge & $91,224 \, (92.13\%)$ \\ \hline
			\multirow{2}{*}{HiRID} & \multirow{2}{*}{$14,129$} & \multirow{2}{*}{$14,129$} & \multirow{2}{*}{$50$} & \multirow{2}{*}{$39$} & ICU mortality & $1,266 \, (8.96\%)$ \\
			&  &  &  &  & ICU discharge & $12,963 \, (91.74\%)$ \\ \hline
		\end{tabular}
	\end{adjustbox}
\end{table*}

\subsection{Baseline models}
We compare the performance of EHR-M-GAN with eight state-of-the-art generative methods in literature. \red{However, as these benchmarks typically face challenges when modeling mixed-type EHR timeseries \cite{hjelm2017boundary} and can only synthesize single-type EHRs, we draw the comparison between EHR-M-GAN and the benchmark models using the corresponding partial component of our synthetic results}, i.e., either the continuous-valued part or the discrete-valued part.
For continuous-valued timeseries generation, benchmark GAN models include C-RNN-GAN \cite{mogren2016c}, R(C)GAN \cite{esteban2017real} and TimeGAN \cite{yoon2019time}.
For discrete-valued timeseries generation, classic medGAN \cite{choi2017generating}, seqGAN \cite{yu2017seqgan}, and two recently proposed work --- SynTEG \cite{zhang2021synteg} and DualAAE \cite{lee2020generating} are used for comparison. Apart from these GAN-based models, we also incorporate PrivBayes \cite{zhang2017privbayes} to synthesize discrete-valued timeseries, which falls in the class of non-GAN generative approaches using a Bayesian framework \cite{tucker2020generating}.
As the original paper of PrivBayes focuses on data anonymization using differential privacy, we therefore implemented its `Non-Private' version for a fair comparison with other baselines (see Section 4.1 Non-Private Methods in\cite{zhang2017privbayes}). 
For medGAN and PrivBayes, we feed the flattened temporal sequence as the input since the models cannot produce timeseries data.

\begin{figure*}[hb!]
	\centering
	\begin{subfigure}[b]{0.31\textwidth}
		\centering
		\includegraphics[width=1.\linewidth]{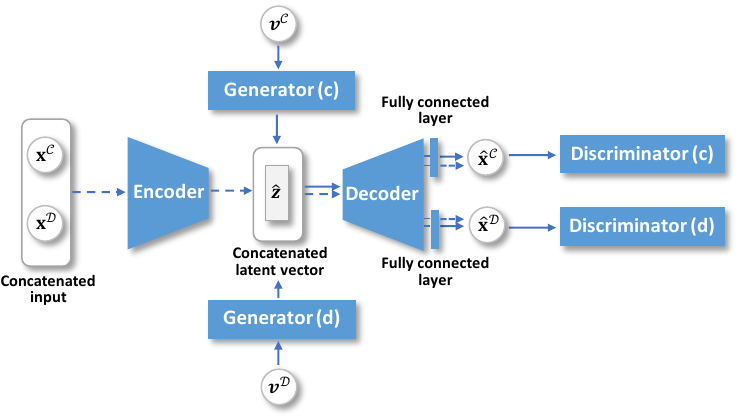} \caption{\textbf{GAN$_{\mathtt{Unified}}$} }
	\end{subfigure}
	\quad
	\begin{subfigure}[b]{0.31\textwidth}  
		\centering 
		\includegraphics[width=0.9\linewidth]{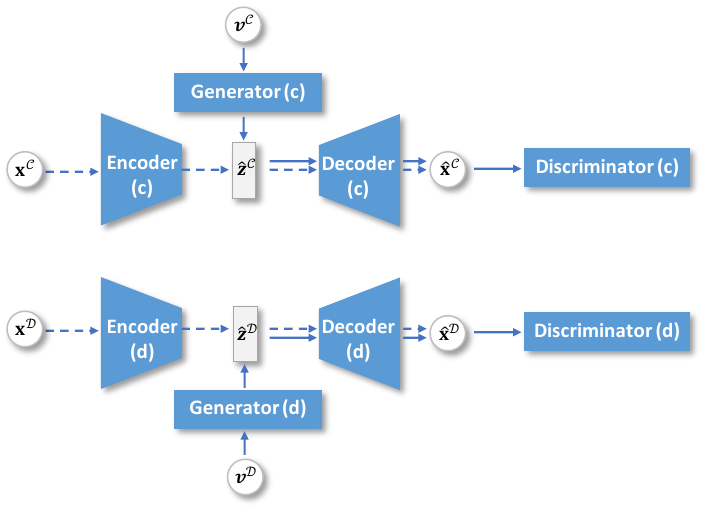} \caption{\textbf{GAN$_{\mathtt{VAE}}$} }
	\end{subfigure}
	\quad
	\begin{subfigure}[b]{0.31\textwidth}  
		\centering 
		\includegraphics[width=0.9\linewidth]{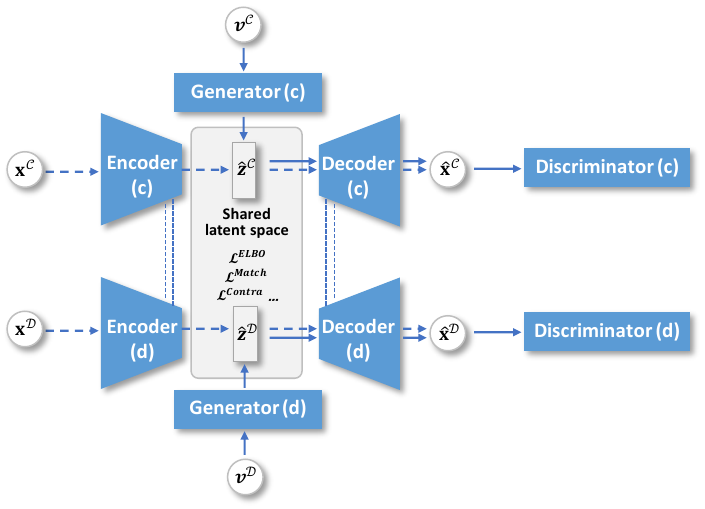} \caption{\textbf{GAN$_{\mathtt{SL}}$} }
	\end{subfigure}
	
	\caption{\textbf{The network architectures in the ablation study.} Three variants of EHR-M-GAN are implemented in the ablation study. Compared with the full model of EHR-M-GAN, \textbf{GAN$_{\mathtt{Unified}}$} learns the joint representations of heterogeneous types of data in a unified network; \textbf{GAN$_{\mathtt{VAE}}$} maintains the basic architecture of EHR-M-GAN, but ignore the dependency learning (i.e., separate networks for two streams of inputs are trained in parallel); \textbf{GAN$_{\mathtt{SL}}$} constructs the shared latent space using the \mbox{\textit{dual-VAE}} module but omit the \mbox{\textit{sequentially coupled generator}} for learning the temporal correlations in the mixed-type timeseries.}
	\label{fig_ablation}
\end{figure*}

\red{We further perform the \textbf{ablation study} to investigate whether our introduced novel components in the proposed model have advantages over its variants that also model mixed-type EHRs.
	First, as EHR-M-GAN learns the joint representations from heterogeneous types of data using separate (but inherently correlated and weights-sharing) VAE networks, we compare it with a variant that jointly models the mixed-type data using a single unified VAE network (denoted as GAN$_{\mathtt{Unified}}$).
	Then, we test the variant that encodes the mixed-type inputs separately with two independent VAE networks, and then combines the resulted synthesis of different data types as outputs (denoted as GAN$_{\mathtt{VAE}}$).
	Lastly, we assess the effectiveness of the proposed \textit{dual-VAE} component in our model alone by implementing GAN$_{\mathtt{SL}}$.}
The architectures of different variants of EHR-M-GAN in the ablation study are detailed as follows (also see Fig. \ref{fig_ablation} for illustration):

\begin{itemize}	
	\item {GAN$_{\mathtt{Unified}}$: It contains a unified VAE module and two separate GANs. \red{The continuous-valued and discrete-valued timeseries is concatenated together, via normalization and one-hot encoding, as input to the encoder in the unified VAE network. The decoder receives the concatenation of the generated latent vectors as the input, and then decodes it into synthetic timeseries with the corresponding data types using the separate fully connected layers.} Each component in the architecture of GAN$_{\mathtt{Unified}}$ (unified encoder and decoder, separate generators and discriminators) is implemented with LSTMs, which are the same as EHR-M-GAN.}
	
	\item {GAN$_{\mathtt{VAE}}$: It is composed of a pair of VAE networks and GANs (one for each type of inputs). \red{The continuous-valued timeseries and discrete-valued timeseries from the same patients are separately fed into the corresponding paths in GAN$_{\mathtt{VAE}}$, and then run in parallel. The synthetic outputs of each data type are then combined as the final results.} It maintains the basic structure of EHR-M-GAN but lacks the latent space sharing with \textit{dual-VAE} and the \textit{sequentially coupled generator} in the original EHR-M-GAN.} 
	
	\item {GAN$_{\mathtt{SL}}$: In addition to GAN$_{\mathtt{VAE}}$, it learns the shared latent space representations through \textit{dual-VAE} by adding the corresponding loss functions \hl{in EHR-M-GAN, including \mbox{\textit{ELBO loss}}, \mbox{\textit{Matching loss}} and \mbox{\textit{Contrastive loss}}.} This model lacks the \textit{sequentially coupled generator}. }
	
	\item {EHR-M-GAN: In addition to GAN$_{\mathtt{SL}}$, it incorporates the \textit{sequentially coupled generator} for the learning the correlated temporal dynamics in timeseries of different data types. This is the proposed full model.}
	
	\item {EHR-M-GAN$_{\mathtt{cond}}$: This version is implemented on the basis of conditional GAN \cite{mirza2014conditional}, where the conditional inputs are fed into EHR-M-GAN to generate patients under specific labels.}
	
\end{itemize}

For training EHR-M-GAN$_{\mathtt{cond}}$, auxiliary information from the patient status is used as conditional input. \hl{These conditional inputs are selected since synthesizing EHR information of patient subgroups with potential outcomes would be valuable for clinicians in their decision-making process. Other conditional labels (such as patient demographics in the categorized format) can also be used in the proposed conditional synthesizer for other research purposes.} For MIMIC-III dataset, the classes are (1) \textit{ICU mortality}: patient died within the ICU; (2) \textit{Hospital mortality}: patient discharged alive from the ICU, and died within the hospital; (3) \textit{30-day readmission}: patient discharged alive from the hospital, and readmitted to the hospital within 30 days; (4) \textit{No 30-day readmission}: patient discharged alive from the hospital, and had no readmission record to the hospital within 30 days. For eICU and HiRID datasets, the corresponding labels are also extracted based on the availability of the patient outcomes (see Table \ref{tb_data_descrption}).

\begin{table*}[t!]
	\caption{\red{\textbf{Summary of the evaluation protocol in this study.} A comprehensive set of evaluation metrics are used to test the \textit{Fidelity}, \textit{Correlation}, \textit{Utility} and \textit{Privacy} of the synthetic EHR data. Definitions of evaluation metrics for corresponding data types are explained. The last column illustrates when the corresponding evaluation metric indicates better performance.}}
	\centering
	\begin{adjustbox}{width=0.99\textwidth,center}
		{\color{Red}
			\begin{tabular}{lll p{.45\textwidth} p{.34\textwidth}}  %%llp{.5\textwidth}
				\hline
				& \textbf{Evaluation metric} & \textbf{Data type} & \textbf{Definition} & \textbf{Better performance} \\ \hline
				\textbf{Fidelity} & Maximum mean discrepancy & Continuous & 
				A kernel-based statistic is calculated to determine whether the real and synthetic data are from the same distribution.
				& Lower value \\
				\specialrule{0em}{1pt}{1pt}
				& Dimension-wise probability & Discrete & The Bernoulli success probability of each feature dimension (i.e., the probability of whether the treatment is active) at the given timestamp is calculated. Probabilities from real and synthetic data are represented on the x and y axis in a single plot to compare the consistency.
				& Scatters closer to the diagonal line (lower RMSE and CC) \\
				\specialrule{0em}{1pt}{1pt}
				& Discriminative score & Continuous and discrete & 
				A classifier is trained post-hoc to tell the difference between the real and synthetic data with its accuracy calculated.
				& Lower accuracy \\
				\specialrule{0em}{1pt}{1pt}
				& Patient trajectories & Continuous and discrete & 
				The mean and standard deviation per time point of real and synthetic patient trajectories are compared and visualized.
				& Similar distributions between the real and synthetic data \\
				\specialrule{0em}{1pt}{1pt}
				\textbf{Correlation} & Pearson pairwise correlations & Continuous and discrete & 
				The correlation between different features is calculated and visualized in a heatmap for both real and synthetic data.
				& Heatmaps corresponding to real and synthetic data more similar (lower CorAcc and $\mu_{abs}$) \\
				\specialrule{0em}{1pt}{1pt}
				& Autocorrelation function & Continuous and discrete & 
				The correlation between the timeseries and its lagged version is calculated and visualized as an ACF curve for both real and synthetic data.
				& ACF curves corresponding to real and synthetic data more similar \\
				\specialrule{0em}{1pt}{1pt}
				\textbf{Utility} & TSTR (downstream task) & Continuous and discrete & 
				The downstream classifier is trained which uses synthetic data as training set, and (hold-out) real data as test set. The result is compared with TRTR to see whether it can maintain the same.
				& Higher AUROCs (with TRTR as baselines) \\
				& TSRTR (downstream task) & Continuous and discrete & 
				The downstream classifier is trained which uses real data and synthetic data as training set, and (hold-out) real data as test set. The result is compared with TRTR to see whether the performance can be improved.
				& Higher AUROCs (with TRTR as baselines) \\
				\specialrule{0em}{1pt}{1pt}
				\textbf{Privacy} & Membership inference attack & Continuous and discrete & 
				A threat model is trained under the black-box setting to determine whether a record is used for training GANs. This quantifies the risk of sensitive information from real data being revealed by synthetic data.
				& Lower accuracy or recall \\
				\specialrule{0em}{1pt}{1pt}
				& Differential privacy & Continuous and discrete & The downstream classifier is trained with differential privacy guarantee. The result is compared with TRTR to see whether it can maintain the same.
				& Higher AUROCs (with TRTR as baselines) 
				\\ \hline
			\end{tabular}
		}
	\end{adjustbox}
	\label{tb:evalsummary}
\end{table*}

\subsection{Evaluation metrics}
Evaluating GAN models is a notoriously challenging task. Advantages and pitfalls of commonly used evaluation metrics for GANs are discussed in \cite{borji2021pros}.
\red{In this work,  a systematic evaluation framework is adopted to assess the quality of synthetic patient EHRs with respect to its \textit{fidelity}, \textit{correlation}, \textit{utility}, and \textit{privacy} (see Table \ref{tb:evalsummary})}. 
We first individually assess the representativeness of the synthetic continuous-valued and discrete-valued timeseries. This includes measuring the distance between underlying data distributions (such as \textit{Maximum mean discrepancy} and \textit{Dimension-wise probability}), \red{comparing the feature-level statistics between the real and synthetic data (\textit{Patient trajectories})}, and assessing the indistinguishability of the synthetic data to the true data (i.e., \textit{Discriminative score}).
Secondly, \red{we evaluate to which extent our model can reconstruct the interdependency between different features (\textit{Pearson pairwise correlations}), and the temporal dynamics in the patient trajectories (\textit{Autocorrelation function}), by using a set of qualitative and quantitative metrics.}
%Secondly, we assess our model by using a set of qualitative metrics (such as \textit{Embedding visualisation}, \textit{Patient trajectory plot} and \textit{Pearson pairwise correlations}) for evaluating to which extent our model can reconstruct the interdependency between two types of data. 
Thirdly, we introduce data augmentation by incorporating synthesized EHR timeseries under various settings, and quantitatively assess the improvement provided by EHR-M-GAN in the \textit{Downstream tasks} for medical intervention prediction in the ICU \red{(i.e., the utility of the synthetic data)}.
Lastly, we measure the attribute of patient privacy-preserving of EHR-M-GAN under \textit{Membership inference attack}. We also evaluate the performance of the same downstream tasks under \textit{Differential privacy} guarantees (See Fig. \ref{fig_key}c \red{and Table \ref{tb:evalsummary}} for the evaluation pipeline).

\section{Results} 
\subsection{Maximum mean discrepancy}
% MMD can assess whether two samples, in our case, true data $x$ and synthetic data $x'$, are from the same distributions. It can be expressed as the first-order moments, i.e., mean embeddings, of the two samples in a reproducing kernel Hilbert space (RKHS): $\mathrm{k}(\mathbf{x}, \mathbf{x'})=\sum_{i} \exp \left({\|\mathbf{x}-\mathbf{x'}\|^{2}}/{\sigma_{i}^{2}}\right)$.
To measure the similarity between the continuous-valued synthetic data and the real data, maximum mean discrepancy (MMD) is used. 
\red{MMD can assess whether two sets of samples are from the same distributions, and in our case, one from the true data $x$ and one from synthetic data $x'$ generated by GANs. To calculate the statistics, a kernel function $K: X \times X' \rightarrow \mathbb{R}$ is used to quantify the similarity between the two distributions.
	In this study, a sum of Gaussian kernel sets is adopted following the implementations in \cite{sutherland2016generative}, which can be expressed as:
	\begin{equation}\label{eq:kernel}
		K(\mathbf{x}, \mathbf{x}')=\sum_{i} \exp \left(- \frac{\|\mathbf{x}-\mathbf{x}'\|^{2}_{F}}{\sigma_{i}^{2}}\right)
	\end{equation}
	where $\sigma_{i}$ is the value of the $i$-th selected bandwidth for calculating $\mathrm{MMD}$.
	As in our study, the real and synthetic samples are multivariate timeseries aligned along the fixed time axis (i.e., 24 data points per patient), we therefore handle these multivariate timeseries as matrices and use the kernel function to calculate the Frobenius norm ($\left\| \cdot \right\|_F$) between them \cite{esteban2017real}.
	
	Finally, given samples $\left\{\mathbf{x}_i\right\}_{i=1}^N$ from real distributions, and samples 
	$\left\{\smash{\mathbf{x}'_j}\right\}_{j=1}^M$ from the synthetic distributions (with $N$ and $M$ denoting the corresponding sample sizes), the estimation of $\mathrm{MMD}$ can be defined as:
	\begin{equation}\label{eq:mmd}
		\begin{aligned}
			\widehat{\mathrm{MMD}^{2}} &= \frac{1}{n(n-1)} \sum_{i=1}^n \sum_{j \neq i}^n K\left(\mathbf{x}_i, \mathbf{x}_j\right)-\frac{2}{m n} \sum_{i=1}^n \sum_{j=1}^m K\left(\mathbf{x}_i, \mathbf{x}'_j\right) \\
			&+ \frac{1}{m(m-1)} \sum_{i=1}^m \sum_{j \neq i}^m K\left(\mathbf{x}'_i, \mathbf{x}'_j\right)
		\end{aligned}
	\end{equation}
	It can be inferred from the equation [\ref{eq:mmd}] that higher similarity between the two distributions leads to the lower $\mathrm{MMD}$ value, with the lower bound value zero indicating that the two distributions are identical.}

As indicated in Table \ref{tb_mmd}, EHR-M-GAN outperforms the state-of-the-art benchmarks among all three datasets in synthesizing continuous-valued timeseries. The conditional version --- EHR-M-GAN$_\texttt{cond}$ further boosts the performance of the model by leveraging the information of the condition-specific inputs. Furthermore, as shown in the ablation study, EHR-M-GAN and EHR-M-GAN$_\texttt{cond}$ produce  smaller MMD values when compared to their variants. Using MIMIC-III as an example, compared with the basic model GAN$_\texttt{VAE}$, by integrating the shared latent space learning using \textit{dual-VAE} under multiple loss constraints, the performance of GAN$_\texttt{SL}$ significantly improves (GAN$_\texttt{SL}$ vs. GAN$_\texttt{VAE}$, 0.745 to 0.926, \textit{p}-value < 0.05 from \textit{t}-test\footnote{Unpaired (two-sample) \textit{t}-test with a significance level of 0.05 is used throughout the paper unless specified otherwise.}). By further building the sequentially coupled generator based on BLSTMs and exploiting the information within mixed-type data, the MMD of EHR-M-GAN shows a nearly 24\% improvement over GAN$_\texttt{VAE}$. 
When synthesizing mixed-type timeseries using the unified network, the performance of GAN$_\texttt{Unified}$ for generating continuous-valued timeseries lags behind the proposed EHR-M-GAN. It therefore can be inferred that, compared with EHR-M-GAN which extracts useful hierarchical representations for each data type using tailored encoding layers, it is quite challenging for GAN$_\texttt{Unified}$ to learn marginal distributions from raw mixed-type timeseries with a unified architecture.

\begin{table*}[t!]
	\caption{\label{tb_mmd}\textbf{Maximum mean discrepancy (MMD) of continuous-valued synthetic data.} Lower values of MMD indicate models which can better learn the distribution of the real data.}
	\begin{adjustbox}{width=0.90\textwidth,center}
		\begin{tabular}{l cccccccc} 
			\hline
			
			& 
			C-RNN-GAN & 
			R(C)GAN & 
			TimeGAN &
			GAN$_{\mathtt{Unified}}$ &
			GAN$_{\mathtt{VAE}}$&
			GAN$_{\mathtt{SL}}$& 
			EHR-M-GAN & 
			EHR-M-GAN$_{\mathtt{cond}}$\\ 
			\hline
			
			\textBF{MIMIC-III} & 
			$ 1.038\pm 0.013 $ &   %C-RNN-GAN
			$ 0.971\pm 0.029 $ &   %R(C)GAN
			$ 0.694\pm 0.025 $ &   %TimeGAN
			$ 0.893 \pm 0.027 $ & %GAN_unified
			$ 0.926\pm 0.038 $ &   %GAN_VAE
			$ 0.745\pm 0.040 $ &   %GAN_SL
			$ \textBF{0.692}\pm \textBF{0.034} $ &   %EHR-M-GAN
			$ \textBF{0.604}\pm\textBF{0.027} $ \\  %EHR-M-GAN_c
			
			\textBF{eICU} & 
			$ 1.139\pm 0.023 $ &   %C-RNN-GAN
			$ 1.106\pm 0.042 $ &   %R(C)GAN
			$ 0.672\pm 0.038 $ &   %TimeGAN
			$ 0.850 \pm 0.032 $ & %GAN_unified
			$ 0.842\pm 0.029 $ &   %GAN
			$ 0.670\pm 0.034 $ &   %GAN_VAE
			$ \textBF{0.651}\pm\textBF{0.026}$ &   %EHR-M-GAN
			$ \textBF{0.540}\pm\textBF{0.018}$ \\  %EHR-M-GAN_c
			
			\textBF{HiRID} & 
			$ 0.982\pm 0.017 $ &   %C-RNN-GAN
			$ 0.865\pm 0.020 $ &   %R(C)GAN
			$ 0.470\pm 0.024 $ &   %TimeGAN
			$ 0.518 \pm 0.030 $ & %GAN_unified
			$ 0.532\pm 0.035 $ &   %GAN
			$ 0.508\pm 0.028 $ &   %GAN_VAE
			$ \textBF{0.428}\pm\textBF{0.015}$ &   %EHR-M-GAN
			$ \textBF{0.389}\pm\textBF{0.024}$ \\  %EHR-M-GAN_c
			
			\hline
		\end{tabular}
	\end{adjustbox}
\end{table*}

\subsection{Dimension-wise probability} 
To evaluate the representativeness of the synthetic discrete-valued timeseries, the dimension-wise probability test is employed. \red{To test the probability distributions between the real and synthetic binary features, the Bernoulli success probability $p \in [0, 1]$ is calculated for the discrete-valued timeseries, and is visualized through scatterplot.
	As a sanity check, it investigates if the probability of the medical intervention being active at the given timestamps is matched between the real data ($x$-axis) and synthetic data ($y$-axis).} 
The correlation coefficients (CCs) and root-mean-square errors (RMSEs) are also adopted \mbox{\cite{baowaly2019synthesizing}} based on the Bernoulli success probabilities to quantitatively measure the distribution divergence between real and synthetic data.

As shown in Fig. \ref{fig_pairwise_prob} (see Fig. S4 and S5 for more results on eICU and HiRID datasets), the optimal results are provided by EHR-M-GAN and EHR-M-GAN$_{\mathtt{cond}}$. 
The close-to-real probability distributions that appear along the diagonal line indicate the remarkable similarity between the real data and the synthetic data provided by our models. 
The quantified CC and RMSE also correspond with the visulisation results, which are close to the highest mark (EHR-M-GAN: RMSE = 0.0095, CC = 0.9973).
Similar to the results in MMD, the dimensional-wise distributions are better captured when modules such as \mbox{\textit{dual-VAE}} and \mbox{\textit{sequentially coupled generator}} are introduced in EHR-M-GAN. 
GAN$_\texttt{Unified}$ suffers from mode collapse (the generator fails to produce outputs with sufficient diversity), and therefore shows poor performance compared with other variants when synthesizing discrete-valued timeseries. As the mixed-type features are treated as unimodal input without differentiating their heterogeneous nature, no marginal representations are explicitly learned.

Among all state-of-the-art benchmark models, DualAAE shows the best result but is slightly sub-optimal when compared to EHR-M-GAN. In contrast, both skewed distribution and low performance scores are observed in medGAN, as it lacks the ability to capture the temporal correlations within timeseries.
SynTEG shows improved performance over medGAN, as it is capable of synthesizing discrete-valued features in EHRs with timestamps. The non-GAN generative method PrivBayes also shows good performance among all the benchmark synthesizers when modeling the underlying probability distribution of the discrete-valued EHR timeseries. On the other hand, despite the well-known performance of SeqGAN in natural language generation, it is not quite applicable in synthesizing sequential clinical EHRs.

Generating discrete-valued features are known to be problematic for traditional GANs. Due to their limitation in passing the gradients from the critic models, vanilla GANs cannot update their generators efficiently based on the adversarial loss \cite{yu2017seqgan, choi2017generating}. However, the result of EHR-M-GAN shows its superiority in explicitly capturing each dimension of the discrete-valued sequences. EHR-M-GAN mitigates this problem by learning the shared latent representations using \textit{dual-VAE}. Discrete-valued timeseries are encoded into a gradient-differentiable space for further optimizing the generators and thus solving the problem.

\begin{figure*}[t!]
	\centering
	\begin{subfigure}[b]{0.16\textwidth}
		\centering
		\includegraphics[width=1.05\linewidth]{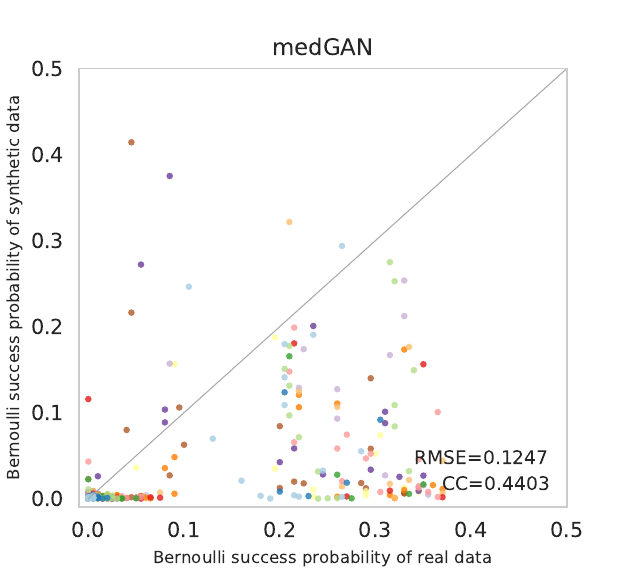} 
	\end{subfigure}
	\quad
	\begin{subfigure}[b]{0.16\textwidth}  
		\centering 
		\includegraphics[width=1.05\linewidth]{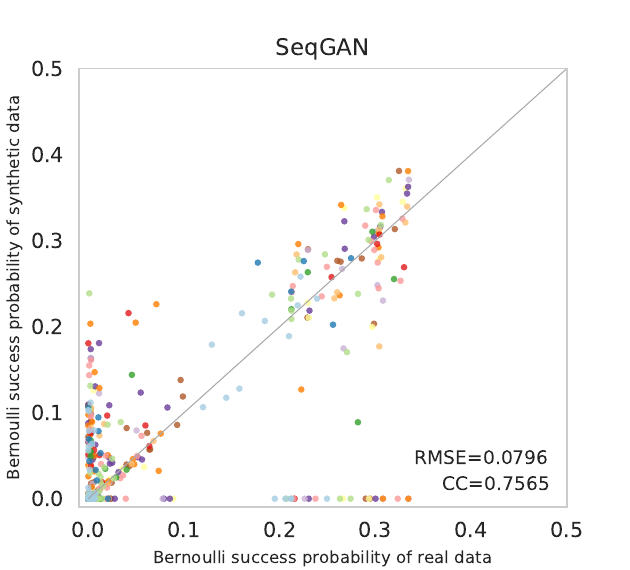} 
	\end{subfigure}
	\quad
	\begin{subfigure}[b]{0.16\textwidth}  
		\centering 
		\includegraphics[width=1.05\linewidth]{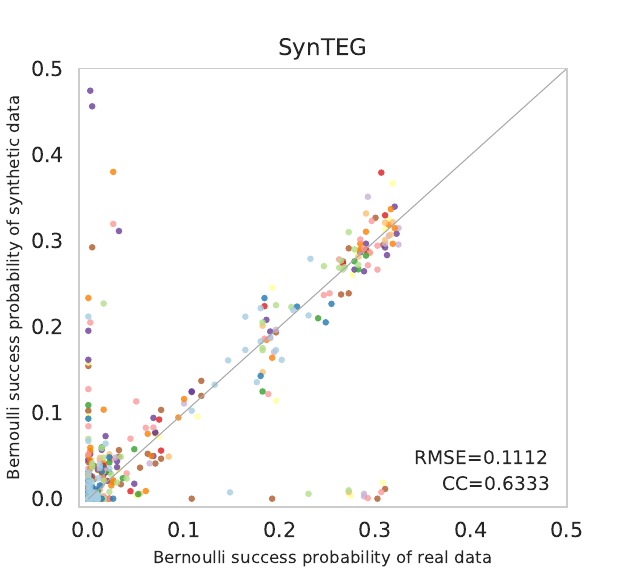} 
	\end{subfigure}
	\quad
	\begin{subfigure}[b]{0.16\textwidth}  
		\centering 
		\includegraphics[width=1.05\linewidth]{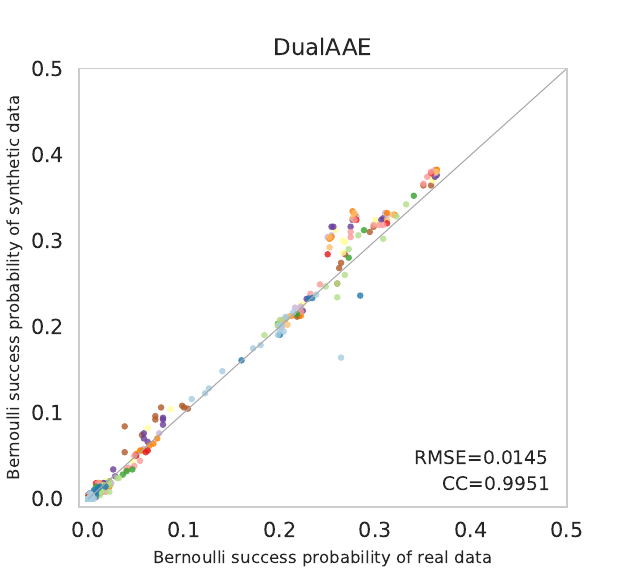}  
	\end{subfigure}
	\quad
	\begin{subfigure}[b]{0.16\textwidth}  
		\centering 
		\includegraphics[width=1.05\linewidth]{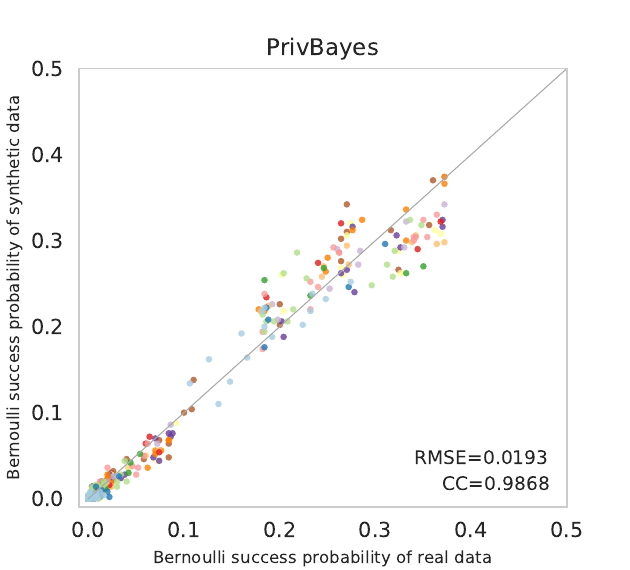}
	\end{subfigure}
	%%%%%%%%%%%%%%%%%
	\medskip
	\begin{subfigure}[b]{0.16\textwidth}
		\centering
		\includegraphics[width=1.05\linewidth]{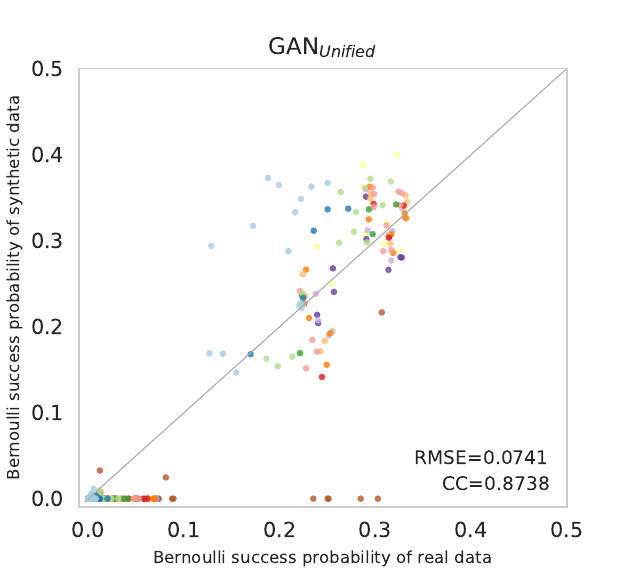}
	\end{subfigure}
	\quad
	\begin{subfigure}[b]{0.16\textwidth}
		\centering
		\includegraphics[width=1.05\linewidth]{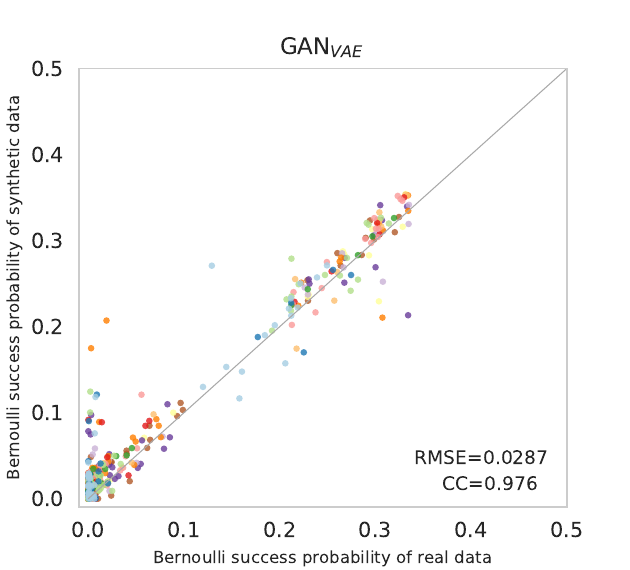}
	\end{subfigure}
	\quad
	\begin{subfigure}[b]{0.16\textwidth}  
		\centering 
		\includegraphics[width=1.05\linewidth]{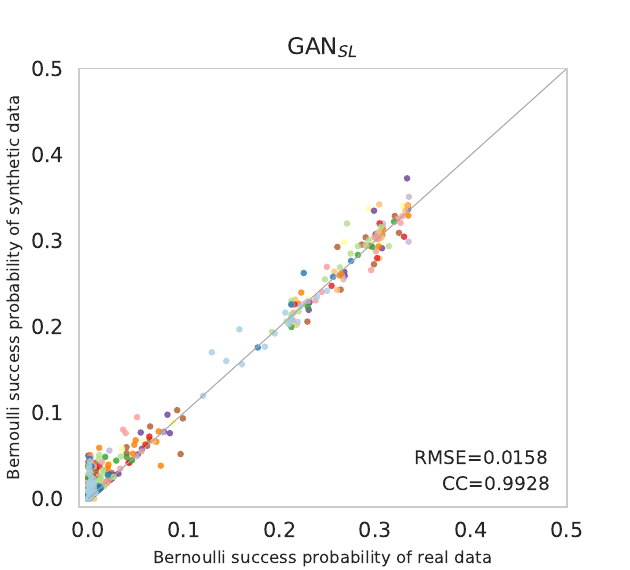}  
	\end{subfigure}
	\quad
	\begin{subfigure}[b]{0.16\textwidth}  
		\centering 
		\includegraphics[width=1.05\linewidth]{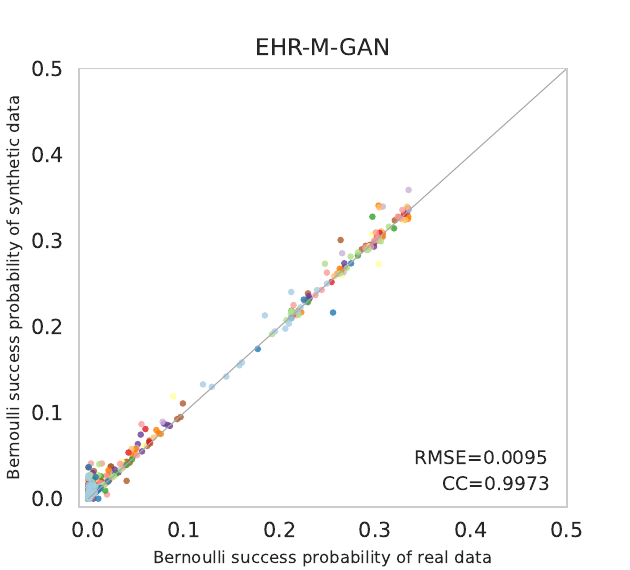}  
	\end{subfigure}
	\quad
	\begin{subfigure}[b]{0.16\textwidth}  
		\centering 
		\includegraphics[width=1.05\linewidth]{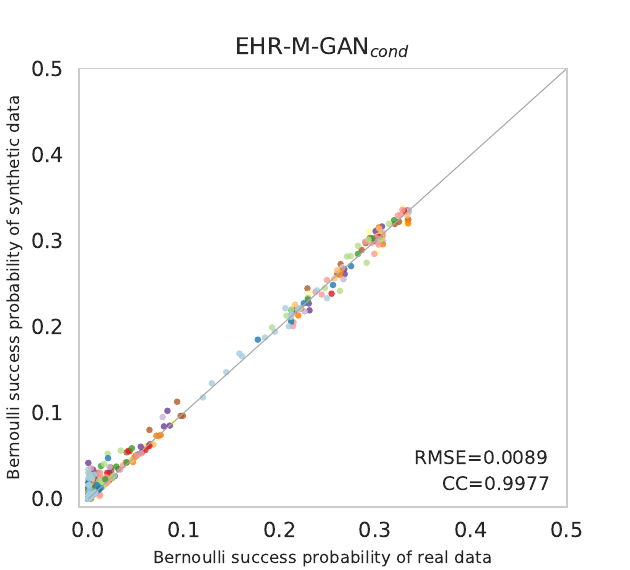}   
	\end{subfigure}	
	
	\caption{\textbf{Scatterplot of the dimension-wise probability test on MIMIC-III dataset.} \red{Dimension-wise probability calculates the Bernoulli success probability of each dimension, i.e., the probability of the treatment being active at a particular time. The x-axis and y-axis represent dimension-wise probability for the real data and synthetic data generated from different models, respectively.} The same color indicates the same treatment (but with varying timestamps). The optimal performance appears along the diagonal line. 
		The corresponding CCs ($[0, 1]$, the higher the better) and RMSEs ($\left[ 0, +\infty \right) $, the lower the better) are also calculated to quantify the probability distribution similarities between the real and synthetic EHRs timeseries. 
		Dimension-wise probability plot for eICU and HiRID dataset can be found in Supplementary materials (see S.4.A).}\label{fig_pairwise_prob}
\end{figure*}

\subsection{Patient trajectories}
\red{We compare the distribution of patient trajectories per timepoint between the real data and synthetic data generated by EHR-M-GAN for theMIMIC-III dataset. 
	Five commonly measured vital sign and laboratory  features --- \textit{Oxygen Saturation}, \textit{Systolic Blood Pressure}, \textit{Respiratory Rate}, \textit{Heart Rate}, \textit{Temperature}, as well as two medical intervention features --- \textit{Mechanical Ventilation} and \textit{Vasopressor}, are considered and compared as an exemplar in Fig. \ref{fig_patient_traj}. 
	It can be inferred that the proposed model can accurately capture the statistical distribution (mean and standard deviation) of both continuous-valued and discrete-valued features. The temporal dynamics are well-preserved in the synthetic timeseries. For example, the variance of \textit{Oxygen Saturation} gradually increases towards the ICU endpoints in the real data, and is closely reflected in the synthetic timeseries. Furthermore, EHR-M-GAN$_{\mathtt{cond}}$ shows superior performance as it can generate correct trajectories with specific patient conditions (see section S.4.D in Supplementary Materials for results).}

\subsection{Discriminative score}
For both continuous-valued and discrete-valued data, the discriminative score is measured as the accuracy of a discriminator trained post-hoc to separate real from generated samples. Synthetic data are generated with the same amount of the hold-out test set from the original data, and are labeled as \textit{synthetic} and \textit{real} correspondingly to train the binary classifier. 
\red{In this study, the classifier (critic) is implemented with a single-layered Bi-directional Long Short-Term Memory (Bi-LSTM) model (i.e., \textit{many-to-one}), with its parameters randomly initialized (as opposed to critic built upon representations from the trained generative model \cite{zhang2022keeping}).
	The critic trained from the supervised learning task can be used to characterize the temporal correlations across the patient EHR timeseries.}

\begin{table}[h]
	\caption{\label{tb_discriminative_score}\textbf{Discriminative score of synthetic data}. A discriminative model is trained post-hoc to discriminate between synthetic samples and real samples. The accuracy from the discriminative classifier is used as the discriminative score, where the lower value indicates better performance. The result is bounded by 0.5 when the classifier cannot distinguish between two distributions.}
	\begin{adjustbox}{width=0.75\textwidth,center}
		
		\begin{tabular}{lllll}
			\hline
			& Method & MIMIC-III & eICU & HiRID \\ \hline
			\multicolumn{1}{c}{\multirow{7}{*}{\begin{tabular}[c]{@{}c@{}}
						Continuous-valued \\ synthetic data\end{tabular}}} & 
			C-RNN-GAN &
			$0.825 \pm 0.013 $ & 
			$0.876 \pm 0.010 $ & 
			$0.774 \pm 0.022 $\\
			\multicolumn{1}{c}{} 
			& R(C)GAN &           
			$0.833 \pm 0.028 $ & 
			$0.850 \pm 0.021 $ & 
			$0.742 \pm 0.016 $\\
			\multicolumn{1}{c}{} 
			& TimeGAN &
			$0.763 \pm 0.018 $ & 
			$0.790 \pm 0.013 $ & 
			$\textBF{0.716} \pm \textBF{0.024} $ \\
			\multicolumn{1}{c}{}                          
			& GAN$_{\mathtt{Unified}}$& 
			$0.809 \pm 0.023 $ & 
			$0.863 \pm 0.027 $ & 
			$0.749 \pm 0.014 $ \\
			\multicolumn{1}{c}{}                          
			& GAN$_{\mathtt{VAE}}$ & 
			$0.842 \pm 0.020 $ & 
			$0.871 \pm 0.014 $ & 
			$0.802 \pm 0.017 $ \\
			\multicolumn{1}{c}{}                          
			& GAN$_{\mathtt{SL}}$ &
			$0.786 \pm 0.016 $ & 
			$0.813 \pm 0.023 $ & 
			$0.752 \pm 0.021 $ \\
			\multicolumn{1}{c}{} 
			& EHR-M-GAN &
			$\textBF{0.746} \pm \textBF{0.018} $ & 
			$\textBF{0.776} \pm \textBF{0.015} $ & 
			$0.724 \pm 0.015 $ \\
			\multicolumn{1}{c}{} & EHR-M-GAN$_{\mathtt{cond}}$ & 
			$\textBF{0.729} \pm \textBF{0.025} $ & 
			$\textBF{0.745} \pm \textBF{0.017} $ & 
			$\textBF{0.693} \pm \textBF{0.012} $ \\ 
			\hline
			\multicolumn{1}{c}{\multirow{7}{*}{\begin{tabular}[c]{@{}c@{}}
						Discrete-valued \\ synthetic data\end{tabular}}} 
			& medGAN &
			$0.903\pm0.027$ & 
			$0.915\pm0.034$ & 
			$0.896\pm0.021$ \\
			& seqGAN & 
			$0.937\pm0.025$ & 
			$0.924\pm0.023$ & 
			$0.913\pm0.027$ \\
			& SynTEG &
			$0.879 \pm 0.021$ & 
			$0.902 \pm 0.030$ & 
			$0.878 \pm 0.025$ \\
			& DualAAE &
			$0.847\pm0.029$ & 
			$0.860\pm0.033$ & 
			$0.829\pm0.024$ \\
			& \hl{PrivBayes} &
			\hl{$0.859 \pm 0.036$} & 
			\hl{$0.883 \pm 0.034$} & 
			\hl{$0.832 \pm 0.017$} \\
			\multicolumn{1}{c}{}                          
			& \hl{GAN$_{\mathtt{Unified}}$} & 
			\hl{$0.890 \pm 0.022 $} & 
			\hl{$0.907 \pm 0.026 $} & 
			\hl{$0.849 \pm 0.015 $} \\
			& GAN$_{\mathtt{VAE}}$  &
			$0.862\pm0.024$ & 
			$0.881\pm0.029$ & 
			$0.824\pm0.018$ \\
			& GAN$_{\mathtt{SL}}$   &
			$0.829\pm0.032$ & 
			$0.844\pm0.028$ & 
			$0.816\pm0.025$ \\
			& EHR-M-GAN &
			$\textBF{0.813}\pm\textBF{0.026}$ & 
			$\textBF{0.831}\pm\textBF{0.024}$ & 
			$\textBF{0.802}\pm\textBF{0.020}$ \\
			& EHR-M-GAN$_{\mathtt{cond}}$ & 
			$\textBF{0.784}\pm\textBF{0.024}$ & 
			$\textBF{0.803}\pm\textBF{0.022}$ & 
			$\textBF{0.778}\pm\textBF{0.019}$ \\ 
			\hline
		\end{tabular}
		
	\end{adjustbox}
\end{table}

\red{As indicated from the results in Table \ref{tb_discriminative_score}, it appears that EHR-M-GAN and EHR-M-GAN$_\texttt{cond}$ can produce synthetic data that are less distinguishable from real data than the benchmarked models.}
Especially for EHR-M-GAN$_\texttt{cond}$, it achieves the optimal discriminative scores consistently against other benchmarks for both continuous-valued and discrete-valued timeseries. 
\hl{For discrete-valued data generation, EHR-M-GAN-generated samples achieve the discriminative score of 0.813 on the MIMIC-III dataset, which has a 4\% statistically significant improvement over the best performing benchmark (EHR-M-GAN vs. DualAAE: 0.813 to 0.847, $p<0.05$). 
The overall discriminative scores produced by PrivBayes on three ICU databases are comparable with the GAN models such as SynTEG and DualAAE.}
For continuous-valued timeseries generation, the discriminative score of TimeGAN on HiRID dataset outperforms the other models as well as EHR-M-GAN, though not statistically significant (EHR-M-GAN vs. TimeGAN: 0.724 to 0.716, $p=0.4374$). By leveraging the additional information from the conditional inputs, EHR-M-GAN$_\texttt{cond}$ can provide significantly better result than TimeGAN (EHR-M-GAN$_\texttt{cond}$ vs. TimeGAN: 0.693 to 0.716, $p<0.05$). 

The ablation study has proved the effectiveness of EHR-M-GAN for generating high quality EHR timeseries. The shared latent space representation learning in the \textit{dual-VAE} (i.e., GAN$_{\mathtt{SL}}$) have shown remarkable success  as making the synthetic data more realistic than separately generating the latent embeddings based on VAEs (as in GAN$_{\mathtt{VAE}}$). The \textit{sequentially coupled generator} further improves the model by capturing the dynamics between mixed-type data and iterating over time, therefore enabling the synthetic timeseries to become more indistinguishable from the original. 
\hl{Further compared with GAN$_{\mathtt{Unified}}$ that models the mixed-type data in a unified network, our proposed model enables effective learning for the marginal distributions from each data type. More importantly, EHR-M-GAN can leverage its dependency learning components to explicitly capture the correlations between heterogeneous types of data.}

%\subsection*{Pearson pairwise correlation}
\subsection{Interdependency characteristics}
In this section, we first employ Pearson pairwise correlation (PPC), which ranges from -1 to 1, to evaluate how closely the synthetic data can model the correlations between continuous-valued and discrete-valued timeseries. Timestamps of the patient trajectories are extracted with every 3 hours interval in a total of 24 hours ICU stay, to explore the temporal dependencies within different variables. \red{To further quantitatively measure the difference between heatmaps generated from real and synthetic samples, we calculate the mean value of the absolute difference between the two PCC matrices ($\mu_{abs}$). We also adopted correlation accuracy (\textit{CorAcc}) \cite{tao2021benchmarking} which quantifies the similarity of two heatmaps within the range of 0 to 1. We discretize the correlation coefficients into 6 correlation levels: \textit{strong negative} ($[-1, -0.5)$), \textit{middle negative} ($[-0.5, -0.3)$), \textit{low negative} ($[-0.3, -0.1)$), \textit{no correlation} ($[-0.1, 0.1)$), \textit{low positive} ($[0.1, 0.3)$), \textit{middle positive} ($[0.3, 0.5)$), and \textit{strong positive} ($[0.5, 1)$). Then, \textit{CorAcc} can be calculated as the percentage of pairs where the real and synthetic data is assigned to the same correlation level.}

\begin{wrapfigure}{l}{0.5\textwidth}
	\centering
	\begin{minipage}[l]{.5\textwidth}
		\includegraphics[width=1\textwidth]{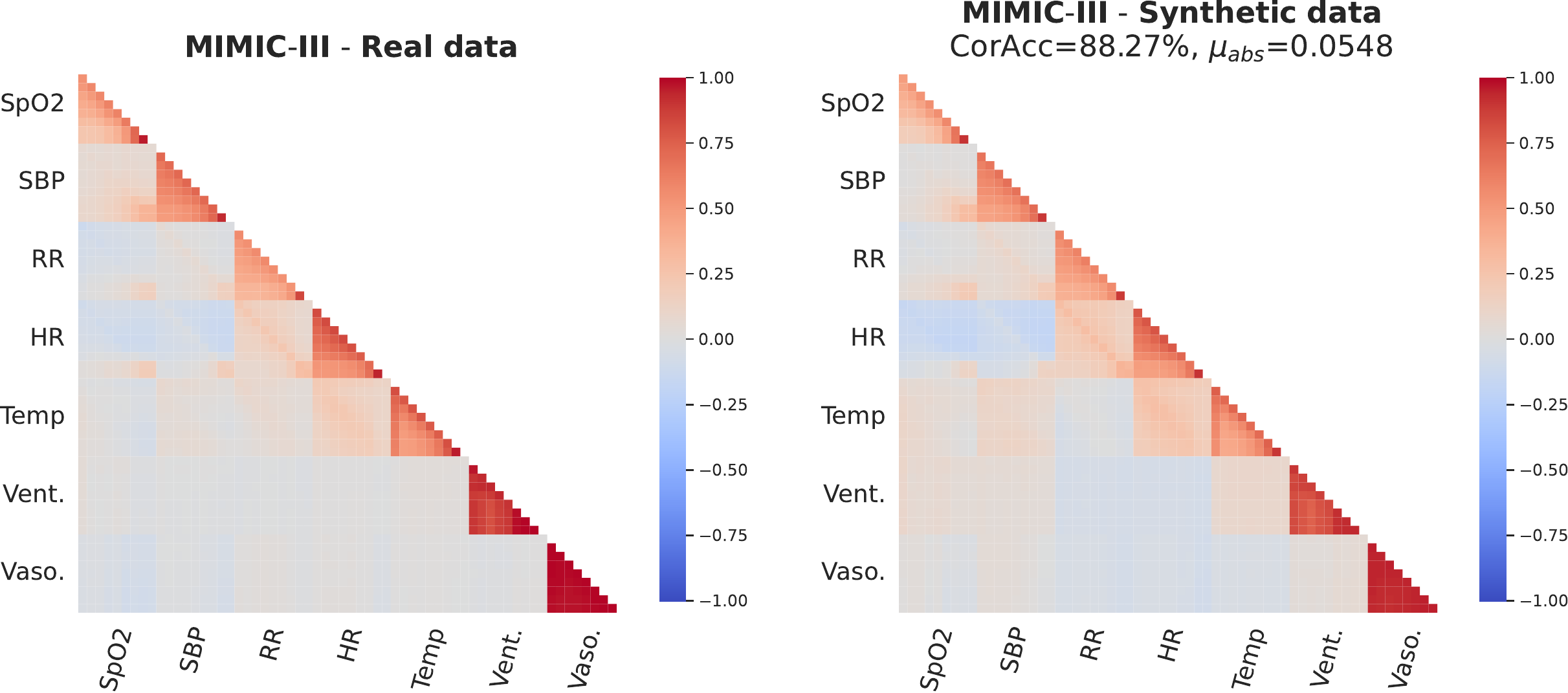}
	\end{minipage}
	\vskip5pt
	\begin{minipage}[l]{.5\textwidth}
		\includegraphics[width=1\textwidth]{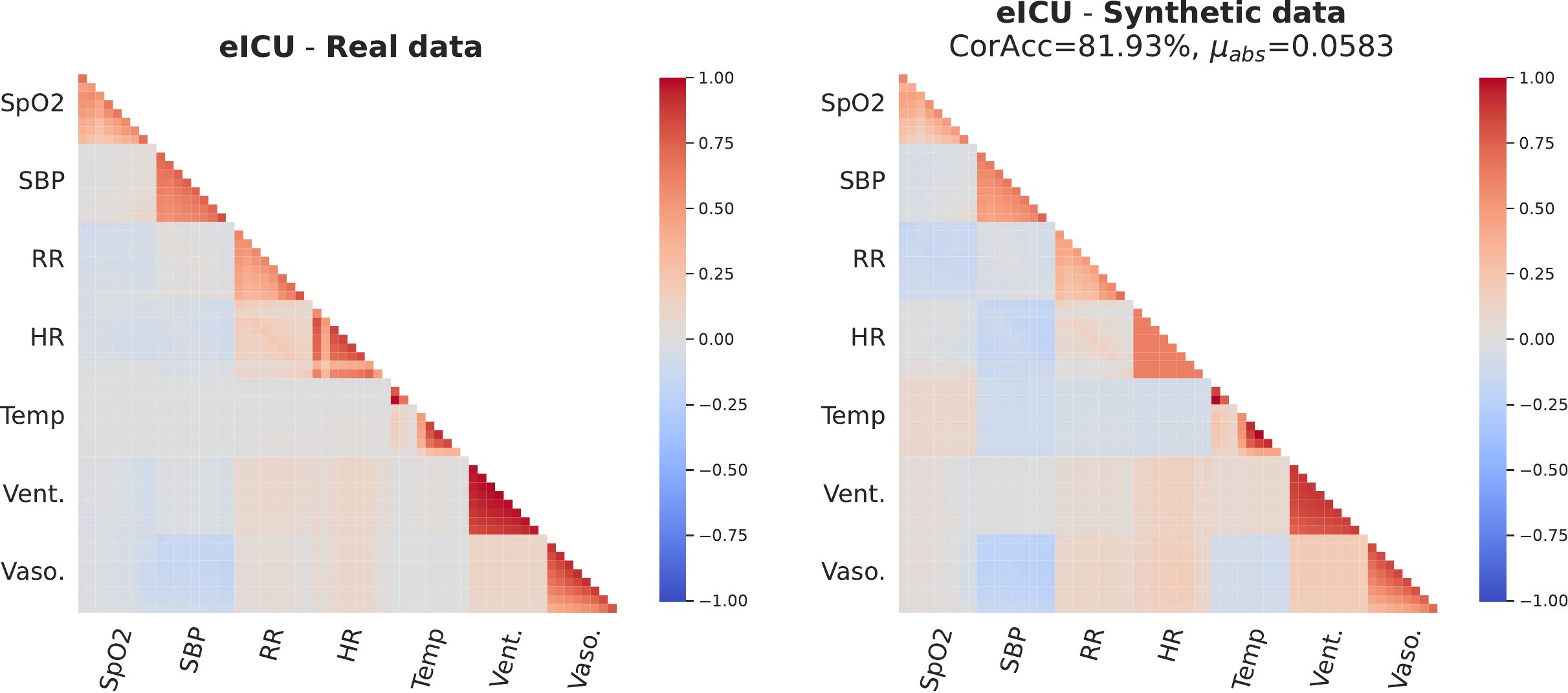}
	\end{minipage}
	\vskip5pt
	\begin{minipage}[l]{.5\textwidth}
		\includegraphics[width=1\textwidth]{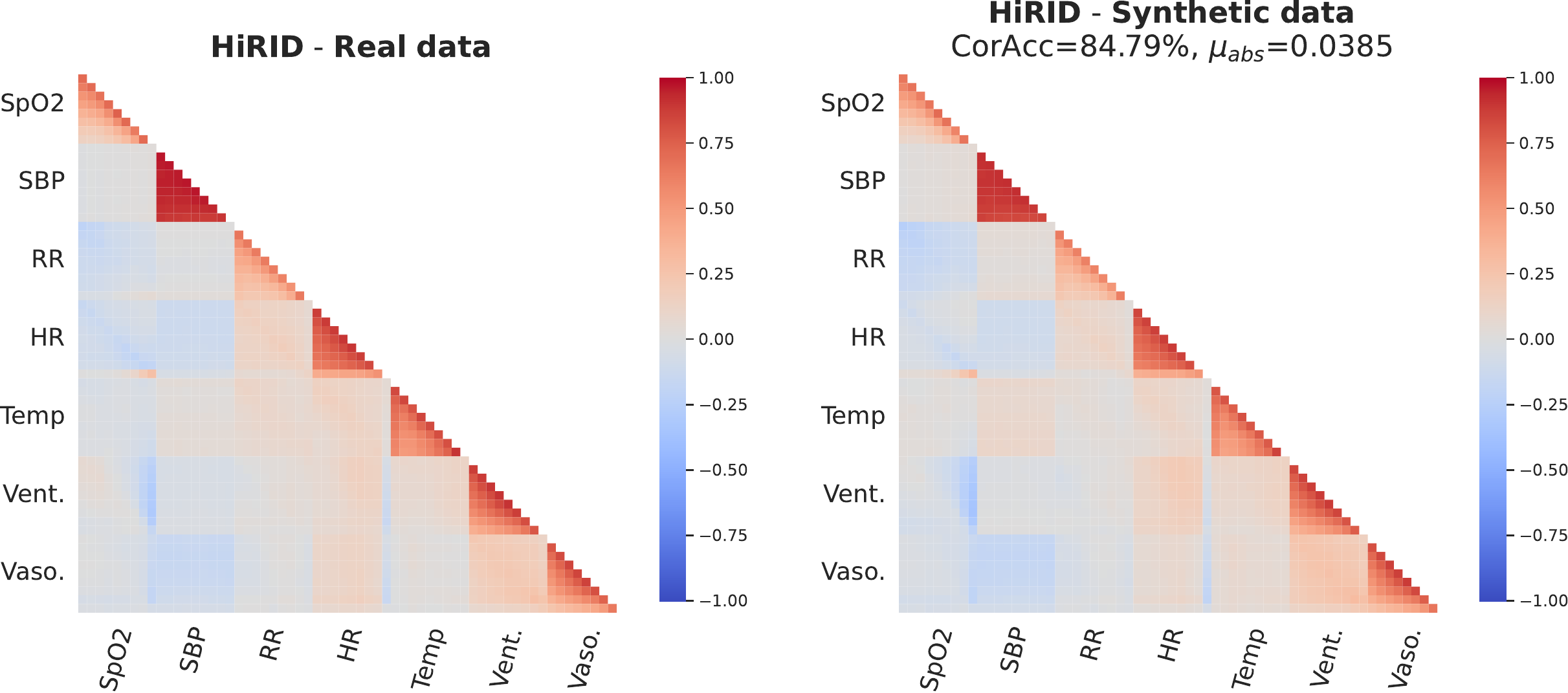}
	\end{minipage}
	\caption{\textbf{Pearson pairwise correlation (PPC) between continuous-valued and discrete-valued timeseries.} The plots contrast the PPC calculated within the real data (left column) and the synthetic data generated by EHR-M-GAN (right column). \red{Besides the visual inspection, the similarity between two heatmaps are quantified by \textit{CorAcc} and $\mu_{abs}$. These metrics indicate how well the synthetic data reconstruct the correlations observed in the real patient trajectories.} As shown in this figure, \textit{SpO2}, \textit{SBP}, \textit{RR}, \textit{HR}, \textit{Temp} represents \textit{Oxygen Saturation}, \textit{Systolic Blood Pressure}, \textit{Respiratory Rate}, \textit{Heart Rate}, \textit{Temperature},  respectively. And \textit{Vent.} and \textit{Vaso.} corresponds to \textit{Vasopressor} and \textit{Mechanical Ventilation}. PPC is calculated every 3 hours over the total 24 hours of ICU stay (ticks of the timestamps are omitted).}
	\label{fig_heatmap}
\end{wrapfigure}

As observed, correlation trends over distinctive features are closely reflected by the synthetic data, \red{with the quantitative measure \textit{CorAcc} 
consistently exceed 0.8 on three critical care databases. It is also worth noticing that EHR-M-GAN can successfully recover temporal dependencies with a high granularity from real patient trajectories.
For example, synchronized correlations across timestamps are observed between \textit{Respiratory Rate} and \textit{Heart Rate} in the MIMIC-III dataset. Such trends are preserved in synthetic data.
This can be explained by the common regulation of these two features by the autonomic nervous system and their synchronized increase in cases of physiological stress, such as hypoxemia. 
In summary, the proposed EHR-M-GAN can reconstruct the temporal dynamics and correlations between features in the real data, which is valuable for downstream ML-based classification and prediction applications.}

Then, autocorrelation functions (ACF) \cite{benedetti2020practical} and the corresponding root-mean-square errors (RMSEs) are calculated to show how EHR-M-GAN can capture the temporal correlations among the timeseries.
ACF measures the relationship between the timeseries and its lagged version. Fig. S6 - S8 in the Supplementary materials shows the ACF calculated for selected continuous-valued and discrete-valued variables (same as Pearson pairwise plot) on real and synthetic timeseries. The time lags are specified as the hourly intervals up to 24 hours before patients' ICU endpoints (ICU discharge or death). Additionally, RMSEs are calculated to quantitatively evaluate the similarity between the corresponding two curves produced by real data and synthetic data. 

Similar patterns are presented between the ACF calculated for real data and their synthetic counterparts, while the quantitative statistics also correspond with the observation. Moreover, overlapping confidence intervals indicate that the synthetic data is able to consistently capture the underlying temporal distributions within the real timeseries. 
%%The 24-hour timeseries data did not exhibit significant seasonal or periodic variation, mainly because these are short-timescale trajectories extracted from EHR data in the critical care setting.
For variables such as \textit{Heart Rate}, \textit{Oxygen Saturation}, and \textit{Systolic Blood Pressure}, the positive ACF coefficients rapidly decrease within the period of first few hours, followed by the growing trends of negative temporal correlation.
The lag with the lowest correlation coefficient is identified at approximately 4 hours. %% This could be related to ....
Specifically, global peaks appear roughly at the 12-hour ticks of \textit{Temperature} for both real and synthetic data on three critical databases. Meanwhile, the negative correlation strengthens as the time lag increase for \textit{Mechanical ventilation} in the original timeseries. Since these behaviours can be reproduced by EHR-M-GAN, therefore they demonstrate that our model can effectively capture the temporal characteristics in the original timeseries.

%Sample trajectories generated by EHR-M-GAN and EHR-M-GAN$_{\mathtt{cond}}$ are shown in Fig. \ref{fig_patient_traj} for MIMIC-III dataset (see Fig. S12, S13, S14 in Supplementary Materials for more results). By visual inspection, we can see that the proposed models can capture patterns within substantially different time-varying signals. Temporal dynamics in both rapidly fluctuating vital signs (e.g., \textit{Oxygen Saturation}) and infrequently changing intervention (e.g., \textit{Mechanical Ventilation}) can be well-preserved. The variability of the synthetic signals suggests that rather than just intelligently ``memorizing'' the training data, EHR-M-GAN is capable of modelling the underlying true data distributions in order to produce \textit{genuine} samples.

%From the instance generated by EHR-M-GAN$_{\mathtt{cond}}$ under the conditional information of \textit{ICU mortality} in Fig. \ref{fig_patient_traj}, it is possible to observe the variations in physiological signals which suggest that the patient suffers from severe deterioration towards the end of the clinical endpoint. Meanwhile, temporal dependencies are observed within different physiological signals along with medical intervention. Synchronized correlations are shown between \textit{Oxygen Saturation}, \textit{Systolic Blood Pressure}, and \textit{Heart Rate} from approximately 18 hr to 21 hr, followed by the presence of \textit{Mechanical Ventilation} for providing the respiratory support for the patient at 22 hr.

\begin{figure}[h!]
	\centering
	\begin{subfigure}[b]{0.135\textwidth}
		\centering
		\includegraphics[width=1\linewidth]{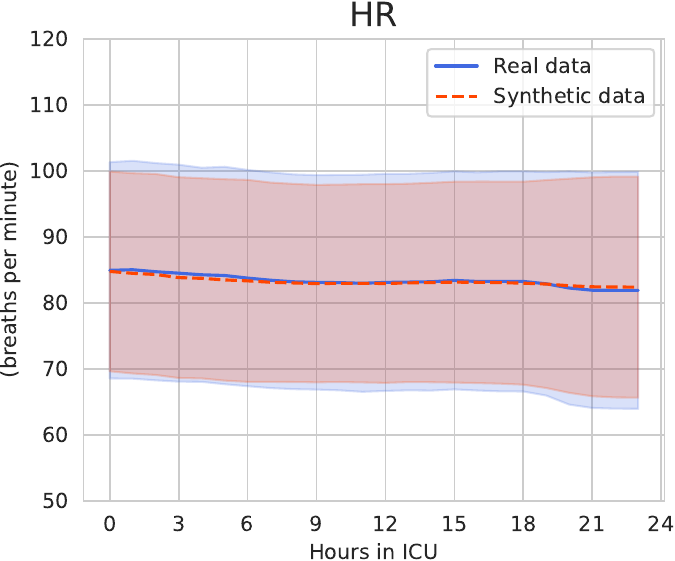} 
	\end{subfigure}
	%\hspace{0.25em}
	\begin{subfigure}[b]{0.135\textwidth}  
		\centering 
		\includegraphics[width=1\linewidth]{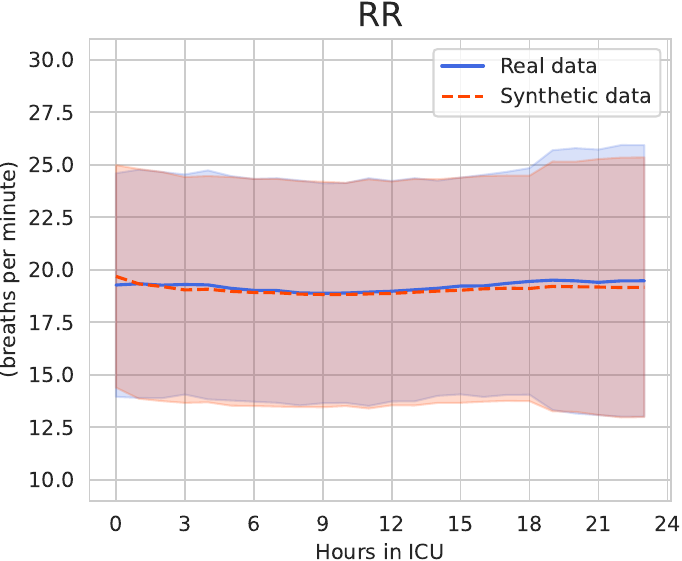} 
	\end{subfigure}
	%\hspace{0.25em}
	\begin{subfigure}[b]{0.135\textwidth}  
		\centering 
		\includegraphics[width=1\linewidth]{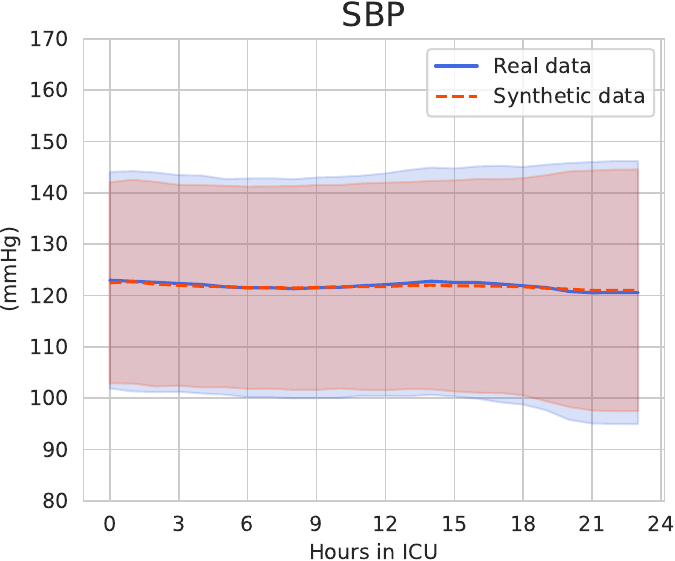} 
	\end{subfigure}
	%\hspace{0.25em}
	\begin{subfigure}[b]{0.135\textwidth}  
		\centering 
		\includegraphics[width=1\linewidth]{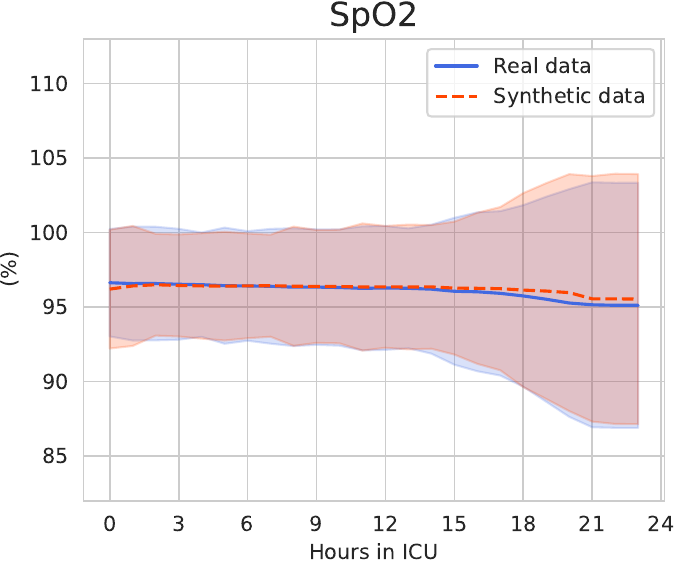} 
	\end{subfigure}
	%\hspace{0.25em}
	\begin{subfigure}[b]{0.135\textwidth}  
		\centering 
		\includegraphics[width=1\linewidth]{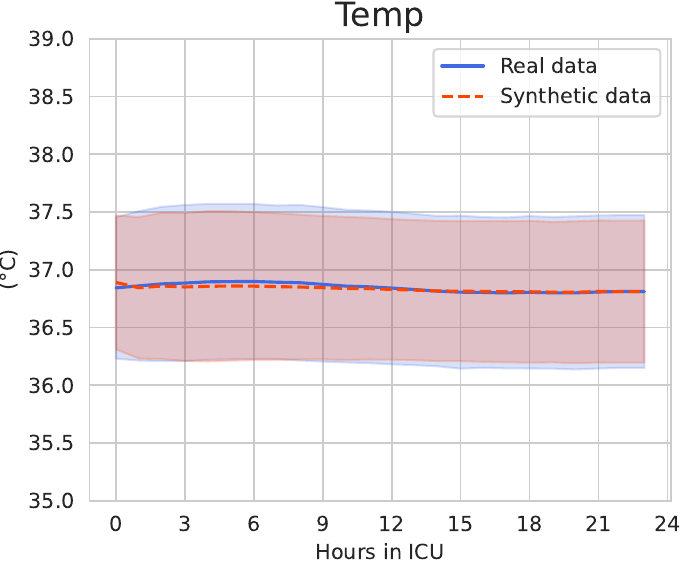} 
	\end{subfigure}
	%\hspace{0.25em}
	\begin{subfigure}[b]{0.135\textwidth}  
		\centering 
		\includegraphics[width=1\linewidth]{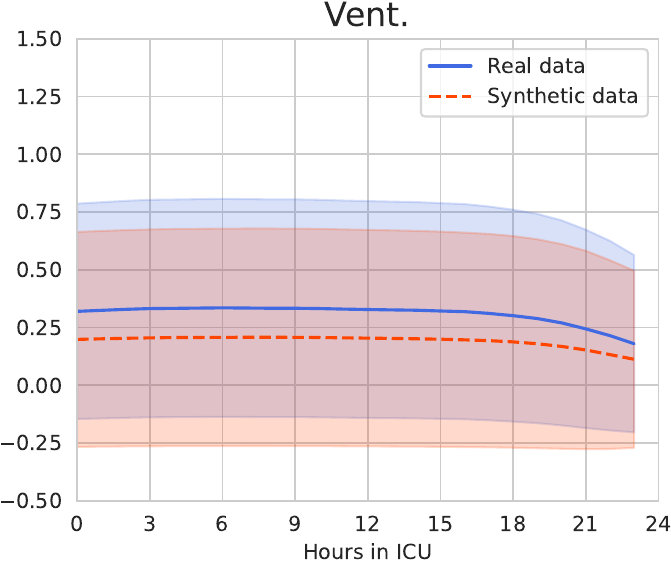} 
	\end{subfigure}
	%\hspace{0.25em}
	\begin{subfigure}[b]{0.135\textwidth}  
		\centering 
		\includegraphics[width=1\linewidth]{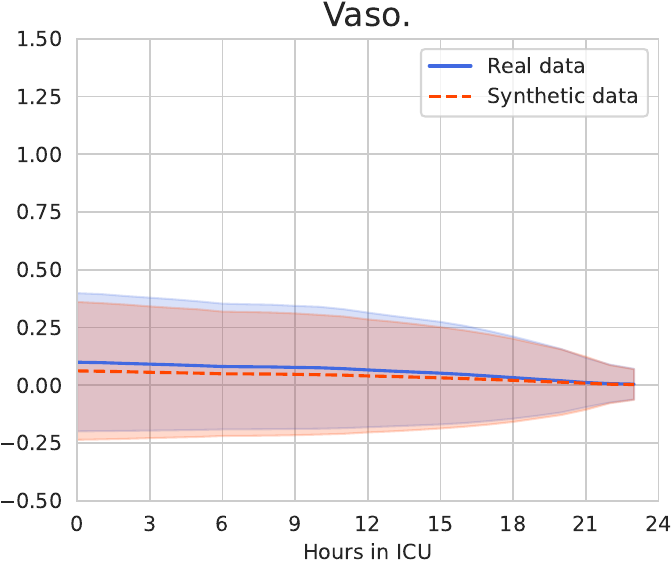} 
	\end{subfigure}
	\caption{\red{\textbf{Comparison of the distribution of values at each timepoint (mean and standard deviation) between real and synthetic patient trajectory produced by EHR-M-GAN.} 
			Multivariate timeseries 24 hours before patients' ICU endpoints are generated, including
			\textit{Heart Rate}, \textit{Respiratory Rate}, \textit{Systolic Blood Pressure}, \textit{Oxygen Saturation}, \textit{Temperature}, \textit{Mechanical Ventilation} and \textit{Vasopressor}. 
			The mean value of the real/synthetic feature at each timepoint is plotted by the solid/dotted line, with the shaded area indicating $\pm 1$ standard deviation.
			For \textit{Mechanical Ventilation} and \textit{Vasopressor}, the y-axis indicates the probability distribution of such intervention being applied ("On") at a given time. The synthetic patient trajectories generated by EHR-M-GAN$_{\mathtt{cond}}$ under different conditions can be found in Supplementary materials.}}
	\label{fig_patient_traj}
\end{figure}

\subsection{Downstream tasks}
% why we choose the task of treatment recommendation
As previously discussed, one of the most prominent goals for GANs is to benefit the future downstream analyses in the real clinical application. A relevant question in the ICU is whether specialized medical treatments, such as therapeutic interventions or organ support, are required for critically ill patients during the admission. Accurate predictions on such tasks can help clinicians to provide actionable, in-time interventions in the resource-intensive ICU. Therefore in this section, \textbf{clinical intervention prediction}  tasks are implemented to evaluate the potential of EHR-M-GAN and EHR-M-GAN$_{\mathtt{cond}}$ in synthesizing high-fidelity synthetic data to further boost the performance of ML classifiers. 
% how to split the observation window, keep it simple (details are in the key figure)
In line with prior work \cite{wang2020mimic, wu2017understanding, suresh2017clinical}, we establish LSTM-based classifiers to predict the status of \textit{mechanical ventilation} and \textit{vasopressors} using continuous-valued multivariate physiological signals as the predictors. A fixed duration of 12 hours is used for both observation window and prediction window (see Fig. \ref{fig_key}). Four outcomes of medical intervention status are defined as: \textit{Stay on}, \textit{Onset}, \textit{Switch off}, \textit{Stay off} (detailed descriptions can be found in Fig. \ref{fig_key}). 

% idea of evaluation --- two aspects
We partition the dataset as illustrated in Figure \ref{fig_downstream_eva_1}, and the performances are assessed from two aspects (see Figure \ref{fig_downstream_eva_2}): \textbf{(i) Traditional approach}: To explore whether the synthetic data can represent the real data accurately, we compare \textit{Train on Real, Test on Real} (TRTR) with \textit{Train on Synthetic, Test on Real} (TSTR), to show whether the performance of a classifier trained on synthetic data from EHR-M-GAN or EHR-M-GAN$_{\mathtt{cond}}$ can be generalized to real data. \hl{In addition to the proposed models, synthetic data produced by the baseline models are also used to train the downstream classifiers for comparison. Other than a measurement of data utility where the downstream task is to predict discrete-valued medical intervention (described as outcomes in this scenario) using continuous-valued physiological features (denoted as predictors), \textit{TSTR} can also be used to assess data synthesizers' ability to capture the interdependencies between the mixed-type features.}
\textbf{(ii) Data augmentation approach}: As data augmentation is employed as a means of circumventing the issue caused by the under-resourced EHR data, here we explore whether synthetic data can used to improve the existing ML algorithms through data augmentation. Therefore, \textit{Train on Synthetic and Real, Test on Real (TSRTR)} is compared with \textit{TRTR} to measure the improvement of the classifier's performance when trained on the augmented data \cite{esteban2017real, kiyasseh2020plethaugment}. The augmentation ratio $\alpha$ or $\beta$ is applied on sub-train data $A^{\prime}_{Tr}$ or synthetic data $B$, in two different scenarios of \textit{TSRTR}, respectively. Details are explained as follows (also see Figure \ref{fig_downstream_eva_2} for illustration).

% different scenarios of two TSRTR-1, alpha
Firstly, as the dearth of data potentially degrades the performance of downstream classifiers, given that the real data has a limited and fixed sample size, we investigate whether adding synthetic EHR data provided by EHR-M-GAN and EHR-M-GAN$_{\texttt{cond}}$ can improve the training of downstream classifiers. \textbf{Ratio ${\bm{\alpha}}$} indicates the portion of synthetic data %\textbf{with ratio $\pmb{\alpha}$} of its initial size 
(see Figure \ref{fig_downstream_eva_2}) being used to augment the real data to improve the quality and robustness of the downstream classifiers. $\alpha$ is set to be 10\%, 25\%, and 50\%, representing the availability of synthetic samples provided for augmentation. 

% different scenarios of two TSRTR-2, beta
Secondly, the acquisition of healthcare data is generally time-consuming and expensive, therefore another overarching goal for the generative model is to minimize the efforts in collecting data. In this section, we investigate whether high-fidelity synthetic data can offer a viable solution for boosting the downstream classifiers' performance when the availability of real data is limited. 
This allows us to understand if the sample size required for real data collection can be reduced while maintaining sufficient predictive power through the use of synthetic data.
During the experiment, the synthetic data $B$ is given (to emulate the scenario where synthetic datasets are available for a particular clinical research purpose), which further is combined with limited real data (collected during clinical trial), to train the downstream classifiers (i.e., augment synthetic data with limited real data). 
Then by implementing EHR-M-GAN or EHR-M-GAN$_{\texttt{cond}}$ in \textit{TSRTR}, we investigate the proportion of the real data $A^{\prime}_{Tr}$ (\textbf{ratio ${\bm{\beta}}$}) required to maintain the same performance as in \textit{TRTR} based on the entire synthetic dataset $B$ (see Figure \ref{fig_downstream_eva_2}).  

\begin{figure}[h!]
	\begin{subfigure}{\columnwidth}
		\centering
		\includegraphics[width=.7\textwidth]{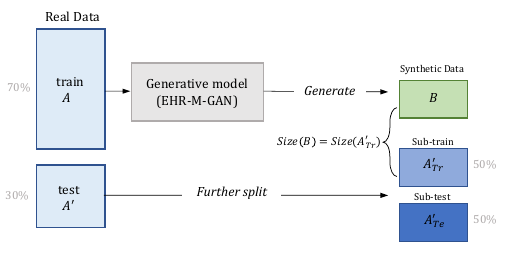}
		\caption{Data splitting. \label{fig_downstream_eva_1}}
	\end{subfigure}
	
	\vspace{5pt}
	
	\begin{subfigure}{\columnwidth}
		\centering
		\includegraphics[width=0.8\textwidth]{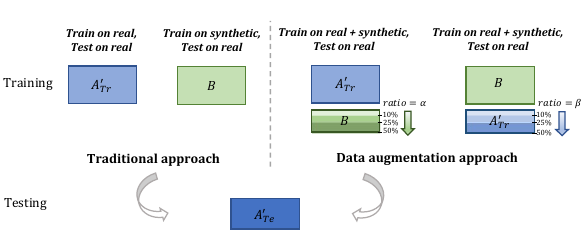}
		\caption{Data augmentation scenarios. \label{fig_downstream_eva_2}}
	\end{subfigure}
	
	\caption{\label{fig_downstream_eva}\textbf{Downstream intervention prediction experimental setup.} \textbf{a. Data splitting.} During training stage, the real data is split into two sets with 70\% training data $A$ and 30\% test data $A^{\prime}$. The test data $A^{\prime}$ is further split into sub-train data $A^{\prime}_{Tr}$ and sub-test data $A^{\prime}_{Te}$ with equal size. Then, the synthetic data $B$, with size equal to the sub-train data $A^{\prime}_{Tr}$, is synthesized by EHR-M-GAN (or EHR-M-GAN$_{\mathtt{cond}}$) trained on the real training data $A$. \textbf{b. Data augmentation scenarios.} Subsequent experiments are trained on set $A^{\prime}_{Tr}$, or $B$, or $A^{\prime}_{Tr} \cup B$ and then tested on $A^{\prime}_{Te}$. In traditional approach, results based on \textit{Train on Real, Test on Real (TRTR)} and \textit{Train on Synthetic, Test on Real (TSTR)} are compared to assess the generalisability of the synthetic data. In data augmentation approach, i.e., \textit{Train on Synthetic and Real, Test on Real (TSRTR)}, we either augment real data $A^{\prime}_{Tr}$ with $\alpha$ (augmentation ratio, 0 to 50\%) of the synthetic samples $B$, or augment synthetic samples $B$ with $\beta$ (0 to 50\%) of the real data $A^{\prime}_{Tr}$.}
\end{figure}

\begin{table*}[h!]
	\caption{\textbf{Downstream task evaluation.} Downstream tasks are evaluated under the training scenarios of \textit{Train on Real, Test on Real (TRTR)} and \textit{Train on Synthetic, Test on Real (TSTR)}. Prediction of two outcomes of interest  -- intervention by \textit{Mechanical ventilation (Vent.)} and \textit{Vasopressors (Vaso.)} are selected as exemplary tasks. Macro-AUROC is used to score the performance of the LSTM-based classifiers on the mutli-class prediction tasks (labeled as \textit{Stay on}, \textit{Onset}, \textit{Switch off}, \textit{Stay off}).}
	\label{tb_downstream_1}
	
	% tableA
	\centering
	\begin{adjustbox}{width=0.9\textwidth,center}
		\begin{tabular}{ll ccc ccc}
			\hline				
			Dataset & 
			Treatments & 
			Real data &
			\hl{GAN$_{\mathtt{Unified}}$} &
			\hl{GAN$_{\mathtt{VAE}}$} &
			\hl{GAN$_{\mathtt{SL}}$} &
			EHR-M-GAN &
			EHR-M-GAN$_{\mathtt{cond}}$ \\ \hline
			
			MIMIC-III & Vent. & 
			$\textBF{0.894}\pm \textBF{0.016} $& 
			\hl{$0.724 \pm 0.015$} &
			\hl{$0.701 \pm 0.018$} &
			\hl{$0.728 \pm 0.010$} &
			$ 0.740 \pm 0.009 $ & 
			$ 0.823\pm 0.020 $ \\
			
			& Vaso. &
			$\textBF{0.841} \pm \textBF{0.009} $& 
			\hl{$0.694 \pm 0.012$} &
			\hl{$0.651 \pm 0.015$} &
			\hl{$0.679 \pm 0.009$} &
			$ 0.725 \pm 0.015 $& 
			$ 0.810 \pm 0.019 $\\
			
			eICU & Vent. & 
			$\textBF{0.868} \pm \textBF{0.015} $&  
			\hl{$0.697 \pm 0.014$} &
			\hl{$0.702 \pm 0.009$} &
			\hl{$0.718 \pm 0.012$} &
			$ 0.745 \pm 0.008 $&  
			$ 0.795 \pm 0.015 $\\
			
			& Vaso. & 
			$\textBF{0.813} \pm \textBF{0.018} $& 
			\hl{$0.648 \pm 0.011$} &
			\hl{$0.657 \pm 0.012$} &
			\hl{$0.665 \pm 0.014$} &
			$ 0.706 \pm 0.014 $&    
			$ 0.748 \pm 0.017 $\\
			
			HiRID & Vent. & 
			$\textBF{0.867} \pm \textBF{0.012} $&   
			\hl{$0.765 \pm 0.014$} &
			\hl{$0.747 \pm 0.013$} &
			\hl{$0.803 \pm 0.008$} &         
			$ 0.825 \pm 0.019 $&    
			$ 0.856 \pm 0.033 $\\
			
			& Vaso. & 
			$\textBF{0.878} \pm \textBF{0.010} $&   
			\hl{$0.754 \pm 0.018$} &
			\hl{$0.752 \pm 0.020$} &
			\hl{$0.779 \pm 0.013$} &
			$ 0.814 \pm 0.015 $&    
			$ 0.844 \pm 0.018 $\\ \hline
			
		\end{tabular}
	\end{adjustbox}

\end{table*}

\begin{table*}[h!]
	\caption{\textbf{Downstream task evaluation with data augmentation.} Downstream tasks are evaluated under the training scenarios of \textit{Train on Synthetic and Real, Test on Real (TSRTR)}. The predictive tasks and evaluation metrics are in accordance with Table \ref{tb_downstream_1}. The upper arrow ($\uparrow$) indicates that the AUROC value under \textit{TSRTR} is higher than \textit{TRTR} in Table \ref{tb_downstream_1} for the corresponding task, while the bold arrow 
		%(\boldsymbol $\uparrow$) 
		\red{(\protect \contour{black}{$\uparrow$})}
		indicates that the value is significantly improved using \textit{t}-test (p$\leq$0.05).	
	}\label{tb_downstream_2}
	
	% tableB
	\begin{subtable}[t]{\textwidth}
		\caption{\label{tb_enhance1}Performance of the downstream LSTM-based classifier under \textit{TSRTR} with data augmentation ratio $\alpha$. All data from sub-train data $A^{\prime}_{Tr}$ concated with $\alpha$ of the synthetic data $B$ (augmentation ratio $\alpha$ = 10\%, 25\% or 50\%) is used as the training set.}
		\begin{adjustbox}{width=1.\textwidth,center}
			\begin{tabular}{llcccccc}
				\hline 
				& & & EHR-M-GAN & & & EHR-M-GAN$_{\mathtt{cond}}$ & \\ 
				\cmidrule(lr){3-5} \cmidrule(lr){6-8}
				
				Dataset & Treatments & 
				$\alpha= 10\%$ & $\alpha= 25\%$ & $\alpha= 50\%$ & 
				$\alpha= 10\%$ & $\alpha= 25\%$ & $\alpha= 50\%$ \\
				
				\cmidrule(lr){1-5} \cmidrule(lr){6-8}
				
				MIMIC-III & Vent.&
				$0.828 \pm 0.013$ &
				$0.877 \pm 0.014$ &
				$0.912 \pm 0.015$ (\contour{black}{$\uparrow$}) & %0.0359
				$0.845 \pm 0.022$ &
				$0.896 \pm 0.013$ ($\uparrow$) & %0.7878
				$\textBF{0.923}\pm\textBF{0.018}$ (\contour{black}{$\uparrow$}) \\ %0.0043
				
				& Vaso. &     
				$0.816 \pm 0.015 $ &
				$0.834 \pm 0.023 $ &    
				$0.859 \pm 0.013 $ (\contour{black}{$\uparrow$}) &  %0.0062
				$\mspace{20mu}$ $0.848 \pm 0.012 $ ($\uparrow$) & %0.2080
				$0.876 \pm 0.017 $ (\contour{black}{$\uparrow$}) &   %0.0001
				$\textBF{0.896}\pm\textBF{0.015}$ (\contour{black}{$\uparrow$}) \\ %0.0001
				
				eICU & Vent. &
				$0.858 \pm 0.008 $&
				$0.862 \pm 0.012 $&   
				$0.873 \pm 0.014 $ ($\uparrow$) &   %0.5019
				$0.865 \pm 0.009 $&
				$0.879 \pm 0.014 $ ($\uparrow$) &   %0.1517 
				$\textBF{0.883}\pm\textBF{0.016}$ ($\uparrow$) \\ %0.0735
				
				& Vaso. &       
				$0.798 \pm 0.015 $&
				$0.805 \pm 0.020 $&  
				$0.821 \pm 0.028 $ ($\uparrow$) &   %0.5077   
				$\mspace{20mu}$ $0.813 \pm 0.016 $ ($\uparrow$) & %1.0000
				$0.834 \pm 0.019 $ (\contour{black}{$\uparrow$}) & %0.0396
				$\textBF{0.839} \pm \textBF{0.014} $ (\contour{black}{$\uparrow$}) \\ %0.0061
				
				HiRID & Vent. &
				$\mspace{20mu}$ $0.871 \pm 0.025 $ ($\uparrow$) & % 0.6895
				$\mspace{20mu}$ $0.882 \pm 0.021 $ ($\uparrow$) & % 0.1013
				$0.913 \pm 0.019 $ (\contour{black}{$\uparrow$}) & % 0.0001
				$\mspace{20mu}$ $0.894 \pm 0.015 $ (\contour{black}{$\uparrow$}) & % 0.0014
				$0.906 \pm 0.018 $ (\contour{black}{$\uparrow$}) & % 0.0001
				$\textBF{0.923} \pm \textBF{0.021} $ (\contour{black}{$\uparrow$}) \\ % 0.0001
				
				& Vaso. &     
				$0.850 \pm 0.016 $&
				$0.874 \pm 0.022 $&     
				$0.894 \pm 0.018 $ (\contour{black}{$\uparrow$}) & % 0.0453   
				$\mspace{20mu}$ $0.883 \pm 0.017 $ ($\uparrow$) & % 0.4851
				$0.908 \pm 0.024 $ ($\uparrow$) &  %  0.0057
				$\textBF{0.913} \pm \textBF{0.019} $ (\contour{black}{$\uparrow$}) \\ % 0.0004
				
				\hline
			\end{tabular}
		\end{adjustbox}
	\end{subtable}

	\vspace*{5mm}
	%%\centering
	
	% tableC
	\begin{subtable}[t]{\textwidth}
		\caption{\label{tb_enhance2}Performance of the downstream LSTM-based classifier under \textit{TSRTR} with data augmentation ratio $\beta$. All data from synthetic data $B$ concated with $\beta$ of the sub-train data $A^{\prime}_{Tr}$ (augmentation ratio $\beta$ = 10\%, 25\% or 50\%) is used as the training set. }
		\begin{adjustbox}{width=1.\textwidth,center}
			\begin{tabular}{llcccccc}
				\hline 
				& & & EHR-M-GAN & & & EHR-M-GAN$_{\mathtt{cond}}$ & \\ 
				\cmidrule(lr){3-5} \cmidrule(lr){6-8}
				
				Dataset & Treatments & 
				$\beta= 10\%$ & $\beta= 25\%$ & $\beta= 50\%$ & 
				$\beta= 10\%$ & $\beta= 25\%$ & $\beta= 50\%$ \\
				
				\cmidrule(lr){1-5} \cmidrule(lr){6-8}
				
				MIMIC-III & Vent. &
				$0.757 \pm 0.016$ &
				$0.824 \pm 0.010$ &
				$0.885 \pm 0.009$ &
				$0.847 \pm 0.017$ &
				$\mspace{20mu}$ $0.903 \pm 0.014$ ($\uparrow$) & % 0.2511
				$\textBF{0.915} \pm \textBF{0.009}$ (\contour{black}{$\uparrow$}) \\ % 0.0060

				& Vaso. &     
				$0.786 \pm 0.019$ &
				$0.810 \pm 0.020$ & 
				$\mspace{20mu}$ $0.849 \pm 0.017$ ($\uparrow$) &  %0.2591
				$0.823 \pm 0.014$ & 	
				$\mspace{20mu}$ $0.851 \pm 0.019$ ($\uparrow$) &  %0.1999
				$\textBF{0.873} \pm \textBF{0.017}$ (\contour{black}{$\uparrow$})\\ %0.0003
				
				eICU & Vent. &
				$ 0.761 \pm 0.011$ &
				$ 0.822 \pm 0.012$ &    
				$\mspace{20mu}$ $ 0.870 \pm 0.019$ ($\uparrow$) & %0.8186
				$ 0.816 \pm 0.016$ &
				$ 0.845 \pm 0.018$ &   
				$\textBF{0.872} \pm \textBF{0.020}$ ($\uparrow$) \\ %(beta = 50%) %0.6578
				
				& Vaso. &     
				$0.742 \pm 0.014$ &
				$0.797 \pm 0.013$ &  
				$\mspace{20mu}$ $0.846 \pm 0.018$ (\contour{black}{$\uparrow$}) & % 0.0025
				$0.785 \pm 0.022$ &
				$\mspace{20mu}$ $0.819 \pm 0.021$ ($\uparrow$) & % 0.5493 
				$\textBF{0.834} \pm \textBF{0.013}$ (\contour{black}{$\uparrow$}) \\ %0.0181
				
				HiRID & Vent. &
				$0.856 \pm 0.012$ & 
				$\mspace{20mu}$ $0.879 \pm 0.019$ ($\uparrow$) &   % 0.1532  
				$\mspace{20mu}$ $0.895 \pm 0.021$ (\contour{black}{$\uparrow$}) &   % 0.0055
				$\mspace{20mu}$ $0.874 \pm 0.016$ ($\uparrow$) &   % 0.3390
				$\mspace{20mu}$ $0.896 \pm 0.018$ (\contour{black}{$\uparrow$}) &   % 0.0020
				$\textBF{0.904}\pm\textBF{0.012}$ (\contour{black}{$\uparrow$}) \\  % 0.0001
				
				& Vaso. &    
				$0.826 \pm 0.024$ &
				$0.859 \pm 0.013$ &     
				$\mspace{20mu}$ $0.893 \pm 0.018$ ($\uparrow$) &  % 0.0585
				$0.865 \pm 0.025$ & 
				$\mspace{20mu}$ $0.897 \pm 0.021$ (\contour{black}{$\uparrow$}) &  % 0.0366   
				$\textBF{0.914}\pm\textBF{0.018}$ (\contour{black}{$\uparrow$})\\ % 0.0002
				
				\hline
			\end{tabular}
		\end{adjustbox}
	\end{subtable}

\end{table*}

% result1: TSTR
\textBF{Traditional approach.} Table \ref{tb_downstream_1} compares the classification performances of predicting forthcoming medical interventions in the ICUs under the experimental setting of \textit{TRTR} and \textit{TSTR}. It is expected that the optimal AUROCs are achieved by the classifiers that are trained on real data. In comparison, the classifiers trained on the synthetic data provided by proposed models can achieve similar performances. More specifically, synthetic data generated by EHR-M-GAN$_{\texttt{cond}}$ demonstrates better generalisability when compared with EHR-M-GAN in the downstream application, such as the task of predicting \textit{mechanical ventilation} on the HiRID dataset (TRTR vs. TSTR from EHR-M-GAN$_\texttt{cond}$: 0.867 to 0.856, with \textit{p}=0.3906).

\hl{Compared with the baseline models, the proposed EHR-M-GAN shows improved performance in \textit{TSTR}, as it can model the distribution of mixed-type EHRs more accurately, while preserving the temporal correlations in the heterogeneous timeseries through the dependency learning components. The results indicate that interdependency between the mixed-type EHRs is weakly captured by GAN$_{\texttt{VAE}}$, as the two streams of inputs are trained in parallel and separately. GAN$_{\texttt{Unified}}$ attempts to capture the temporal correlations of mixed-type EHRs through jointly modeling their underlying distribution in a unified network. However, its unified architecture limits the model's capacity to learn the marginal distribution of each data type, the resulted quality of the synthetic EHRs is impaired and so is its performance in \textit{TSTR}. 
}

% Result2: enhance in clinical setting-1
\textBF{Data augmentation approach (with ratio ${\bm{\alpha}}$).} The results in Table \ref{tb_enhance1} demonstrate that classifiers boosted by EHR-M-GAN can consistently outperform TRTR (see Table \ref{tb_downstream_1}) at the augmentation ratio of 50\%. In comparison, only 25\% of augmentation ratio is needed to achieve improved results for EHR-M-GAN$_{\texttt{cond}}$. For example, the classifier trained on MIMIC-III to predict the status of \textit{Vasopressor} with augmentation ratio $\alpha$ set as 50\%, significantly increase the AUROC by 6\% when compared to the classifier trained using only the real data (EHR-M-GAN$_{\texttt{cond}}$ vs. TRTR: 0.896 to 0.841, $p<0.05$).  Our experiment results have demonstrated that the proposed models can be used for data augmentation to overcome the issue of data scarcity and subsequently improve the classifiers' performance. %applied to generate EHR data that imp Therefore, we can conclude that the degraded performance owing to the clinical data scarcity can be mitigated by implementing the proposed EHR-M-GAN and EHR-M-GAN$_{\texttt{cond}}$ for data enhancement.

% Result3: enhance in clinical setting-2
\textBF{Data augmentation approach (with ratio ${\bm{\beta}}$).} On the other hand, as shown in Table \ref{tb_enhance2}, by augmenting with the synthetic data provided by EHR-M-GAN, only approximately 50\% of the real data is required to keep the classification AUROCs on par with, or even significantly better than fully exploiting the real data under \textit{TRTR}. For EHR-M-GAN$_{\mathtt{cond}}$, the ratio needed for real data to maintain the comparable predictive power is further reduced to 25\%, which equivalently indicates a 75\% reduction of sample size required in real data collection. Overall, results presented in Table \ref{tb_enhance2} demonstrate that by exploiting only a limited ratio of the real data, EHR-M-GAN and EHR-M-GAN$_{\texttt{cond}}$ can robustly maintain the level of prediction performance, therefore alleviating the necessity for acquiring clinical data at scale.

\subsection{Privacy risk evaluation}
Patient privacy is a major concern with regards to sharing electronic health records in any means. 
Generative models can overcome the explicit one-to-one mapping towards the underlying original data in contrast to data anonymisation. \red{However, GAN could potentially raise privacy concerns of information leakage if they simply ``memorise'' the training data, or synthesize samples nearly identical to the real samples (often due to mode collapse).} In that case, sensitive medical information (e.g. national insurance number) belonging to a specific patient used in training GANs can be retrieved during the generative stage, thus posing challenges for preserving privacy in downstream applications. 

In this section, we first quantify the vulnerability of EHR-M-GAN to adversary's \textbf{membership inference attacks}, also known as presence disclosure \cite{hayes2019logan, chen2020gan}. The threat model is implemented under the membership inference for GANs in the \textit{black-box settings} \cite{hayes2019logan}. The attacker is assumed to possess complete knowledge of all the patient records set $P$, where a subset from $P$ further is used to train GANs. During the experiment, the number of samples in the subset for training EHR-M-GAN are varied to investigate the impact of the availability of training data on the success of the attacker (see Figure \ref{fig_privacy1}). By observing the synthetic patient records from EHR-M-GAN, the adversary's goal is to determine whether a single known record $x$ in the patient record set $P$ is from the data used in training EHR-M-GAN. 
\red{If EHR-M-GAN simply "memorises" the training data and can only generate synthetic samples (nearly) identical to the real samples, it would be straightforward for the adversary to identify samples that are used as training data. }
Determined by whether the attacker can correctly infer a given record is \textit{in} or \textit{not in} GAN's training, the accuracy and recall can be calculated.

\begin{figure}[htbp]
	\centering
	\begin{subfigure}[h]{0.43\textwidth}
		\includegraphics[width=1\textwidth]{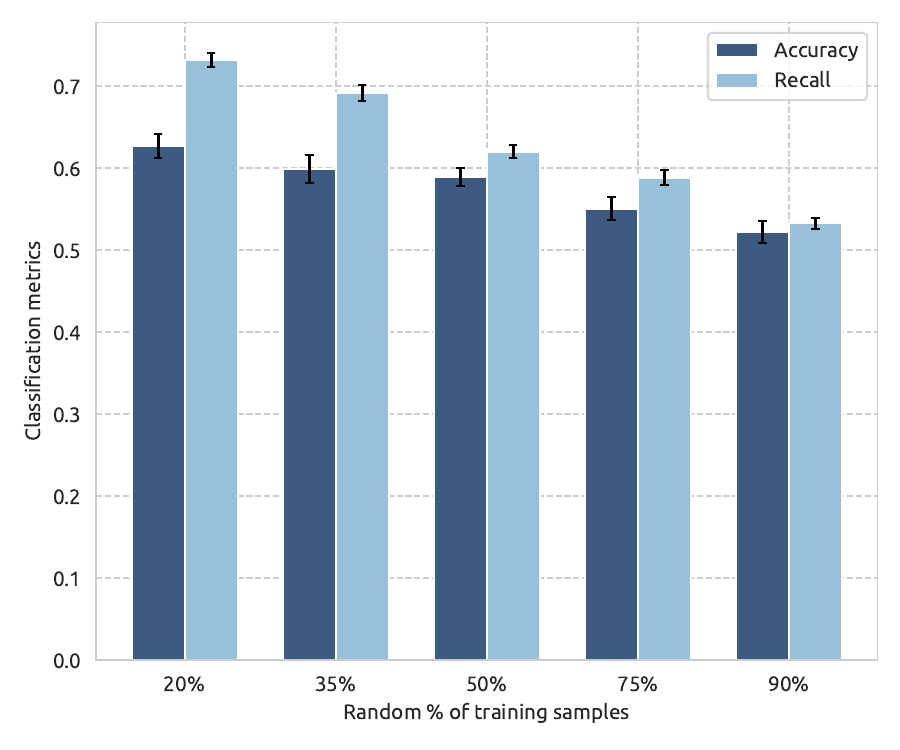}
		\caption{Membership inference attack. \label{fig_privacy1}}
	\end{subfigure}
	\qquad
	\begin{subfigure}[h]{0.43\textwidth}
		\includegraphics[width=1\textwidth]{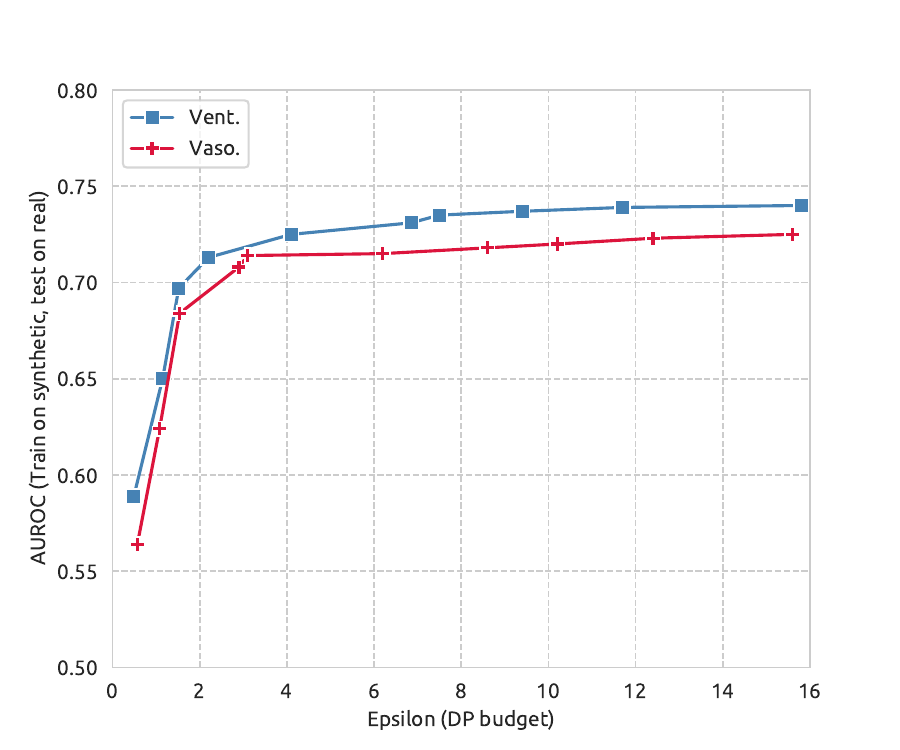}
		\caption{Differential privacy. \label{fig_privacy2}}
	\end{subfigure}
	\caption{\label{fig_privacy}\textbf{Privacy risk evaluation of EHR-M-GAN on MIMIC-III dataset.} \textbf{a. Membership inference attack.} Membership inference attack against EHR-M-GAN vs. the percentage of the training data. Accuracy and recall are used to evaluate the success rate of such attacks. Lower accuracy or recall indicates less privacy information disclosed by the attacker from the generative model (0.5 can be seen as the \textit{random guess} baseline where strong privacy guarantees are provided by GANs). Recall indicates the ratio of samples that are successfully claimed by the attacker among all the real data that are used in training GAN models. \textbf{b. Differential privacy.} Performance of medical intervention prediction tasks, under various differential privacy (DP) budgets, measured by Macro-AUROC. }
\end{figure}

% membership attack
As shown in Figure \ref{fig_privacy1}, when 90\% of the training data is used for developing EHR-M-GAN, the attacker had a recall of 0.533 and accuracy of 0.527 to recover which training data are considered. This is eminently close to flipping a coin with random guess (i.e., 0.5), indicating EHR-M-GAN is sufficiently robust against the membership inference attack. 
\red{In other words, patient samples used in EHR-M-GAN's training are not recoverable by the threat model. }
On the other hand, as the percentage of the training data reduces, both accuracy and recall for membership inference attacks rise. An accuracy of 0.624 and recall of 0.732 are reached with 20\% of training data. This offers a guideline for future application in developing GANs that incorporating more training data can make the generator less susceptible to such attack. This is also consistent with the conclusion derived from the experiment on membership inference attacks in the prior research \cite{lin2020using}.

The concept differential privacy (DP) \cite{dwork2008differential}, which is a rigorous mathematical definition of privacy, has emerged to be the prevailing notion in the context of statistically analyzing data privacy. The $(\epsilon, \delta){\text -}$differential privacy is guaranteed for model $M$, if given any pair of \textit{adjacent} datasets $D$ and $D^{\prime}$ (differing on a single patient record), it holds: $ P[\mathcal{M}(D) \in S] \leq e^{\epsilon} P\left[\mathcal{M}\left(D^{\prime}\right) \in S\right]+\delta$. In our case, $\mathcal{M}(\cdot)$ is the GAN model trained based on $D$ or $D^{\prime}$, and $S$ is the subset of any possible outcomes of the generative process. By perturbing the underlying data distribution, DP bounds the maximum variations of the algorithm when \textit{any} single individual is included or excluded from the dataset. In practice, recent works on developing differentially private deep learning models have benefited from differential private stochastic gradient descent (DP-SGD) algorithm. DP-SGD operates DP by gradient clipping and noise adding during SGD, thereby ensuring that the impact of single record in the training dataset on algorithm parameters is limited within DP's extend. In this section, $(\epsilon, \delta){\text -}$differential privacy is implemented in EHR-M-GAN using TensorFlow Privacy\footnote{https://github.com/tensorflow/privacy}. We then perform the same downstream tasks on medical intervention prediction using synthetic data generated from DP-guaranteed EHR-M-GAN, and compare its performance with \textit{TSTR} (as shown in Table \ref{tb_downstream_1}).

Figure \ref{fig_privacy2} shows the \textit{TSTR} performance of EHR-M-GAN under differential privacy guarantee with varying budgets $\epsilon$ ($\delta$ fixed at $\leq 0.001$). The value $\epsilon$ determines how strict the privacy is, where the smaller value indicates a stronger privacy restriction. As suggested in Figure \ref{fig_privacy2}, the performance of the downstream tasks operated based on the synthetic data generated by EHR-M-GAN improves as the DP budget relaxes ($\epsilon$ increases). We observe that the AUROC of DP-bounded EHR-M-GAN can maintain at an acceptable level even under strict privacy setting. For example, the AUROC for predicting the treatment of \textit{Vasopressor} can maintain at 0.714 (AUROC = 0.725 under \textit{TRTR}) even when the $\epsilon$ decrease to 4, which is an empirically reasonable value for implementing DP in practice \cite{2017LearningWP}. Future work that focuses on privacy-preserving GAN under DP-guarantee is expected, where the fidelity of the synthetic data can be restored without compromising its privacy.

\section{Discussion and conclusions}
In this study, we propose a generative adversarial network entitled EHR-M-GAN, aiming at mitigating the challenge of synthesizing longitudinal EHR with mixed data types. 
%%To better capture the correlations between the continuous-valued and discrete-valued timeseries, shared latent representations are learnt based on the proposed \textit{dual-VAE}, where the dimensionality for the subsequent adversarial learning is reduced. Then the proposed \textit{sequentially coupled generator} built based on the architecture of RNNs enables EHR-M-GAN to model the temporal dependencies between heterogeneous data. 
\red{A comprehensive list of evaluation metrics is introduced for the systematic assessment, in terms of the fidelity, correlation, utility, and privacy of the synthesis model.}
%During the quantitative and qualitative evaluations of the proposed model, 
First, both EHR-M-GAN and its conditional version, EHR-M-GAN$_\texttt{cond}$, demonstrate consistent improvements against the state-of-the-art benchmark GANs in synthesizing timeseries data with high-fidelity. 
\red{This indicates that the distributional characteristics of the EHR timeseries can be well-preserved in the synthetic data provided by EHR-M-GAN, therefore ensuring its usability during clinical data sharing.} 
Second, as opposed to previous models which were confined to synthesizing only one specific type of data, EHR-M-GAN can produce mixed-type timeseries and successfully capture the temporal dynamics \red{and correlation between features.}
\red{By accurately reconstructing the interdependencies and complex clinical relationships between features, downstream studies such as association analysis and outcome prediction can be supported.}
\red{Notably, the proposed models also outperform the GAN variants that allow mixed-type inputs in the ablation study, indicating that the components in EHR-M-GAN are effective in synthesizing mixed-type timeseries with high fidelity, while successfully reconstructing the interdependencies between them.}
Then, during downstream task evaluation, given the prediction of medical interventions in fast-paced critical care environments as an exemplar, \red{the results demonstrate the broad applicability of our model in developing ML algorithm-based decision support tools by data augmentation.}
\red{Lastly, the assessment of privacy risks further demonstrates the synthetic data provided by EHR-M-GAN can preserve the sensitive information in patient records while maintaining an acceptable level of data utility.}

%% implications
\red{The results in our study have several notable implications with respect to the synthesis of EHR data. First, as the proposed model can be used to provide synthetic longitudinal EHRs for various data types while preserving their underlying correlations, it is now feasible to use the synthesized data to improve the performance of ML models for downstream applications such as the prediction of next intervention, or understanding the disease dynamics and patient phenotyping, based on both the continuous and discrete components of EHR timeseries \cite{alaa2019attentive, lee2020temporal}.
	Second, the experimental results indicate that the quality of the synthetic EHR data can be improved by the integration of mixed-type information, in contrast to the benchmarks that utilize single-type data for learning. This also enables us to mimic how information is presented in clinical practices. 
	Furthermore, we can generate condition/outcome-specific patient trajectories along with corresponding interventions, to facilitate clinical prediction and decision-making. 
	% With the ability to generate realistic mixed-type time-series, this will assist clinicians in their decision making. 
}%Furthermore, the findings from our ablation study suggest that, it is not adequate to change the activation functions only or combine the network architectures to accurately learn the distributions of heterogeneous data types. Therefore, we conclude that the proposed model benefits from its novel components that are effective in synthesizing mixed-type timeseries with high fidelity, while successfully reconstructing the interdependencies between them. 
\red{Third, though facing the privacy-utility tradeoffs, the synthetic EHRs data provided by the proposed model leads to negligible privacy risks under the membership inference attacks. This paves the way for a series of applications in clinical research, including but not limited to, enabling the development of ML models by accessing the synthetic data, overcoming the paucity of medical data and improving the robustness of ML algorithms through data augmentation.}

%%% scalability 
\red{Due to the heterogeneous nature of EHR data, besides the ICU setting in our empirical evaluation, there are needs for synthesizing mixed-type EHR timeseries in various clinical scenarios. For example, patients' encounters in hospitals are documented as structured EHRs recorded in the temporal order. Each visit is typically associated with the corresponding medical events presented in the form of discrete-valued ICD codes \cite{zhang2021synteg}, and continuous-valued measurements. These mixed-type EHR timeseries  capture a patient's health status and better align with clinical decision-making process than those using the single-type data alone. Therefore, developing GANs targeting mixed-type EHRs generation have the potential to pave the way for complex deep-learning systems that are capable of integrating information from various sources. However, it is worth noting that the validation of our proposed model is based on critical care settings with limited feature dimensions, can only serve as a proof of concept. When extending the proposed model to other clinical settings, such as synthesizing ICD codes with hundreds or thousands of feature dimensions \cite{zhang2021synteg}, the scalability and utility  of our proposed model when dealing with the enlarged, sparse feature space needs further investigation.}

% limitations-1
There are limitations in the current work. 
\red{First, data curation strategies on clinical timeseries, including truncating, smoothing and imputation, are applied before the EHR timeseries are used for the training of generative models.}
As during the data preprocessing, we first extract the timeseries with a fixed duration (i.e., 24 hours before the ICU clinical endpoints), and then hourly aggregate patients' physiological and intervention signals based on their mean statistics, followed by completing the missing value in the timeseries through the ``Simple Imputation'' approach \cite{che2018recurrent}. 
%%%%
Although these preprocessing steps are commonly used in clinical research under the critical care settings \cite{wang2020mimic}, the proposed model cannot model the irregular time intervals between signals nor missing values within the timeseries. 
%%%%
However, dealing with irregularity of the timestamps when synthesizing clinical events in EHRs could be useful for predicting outcomes that are time-aware in the downstream tasks \cite{zhang2021synteg}. Modeling such time intervals could be non-trivial as the determinative perspectives sometimes go beyond the scope of inferring patients physiological status such as resource allocations within hospitals. Also, synthesizing timeseries while incorporating the missing values could also be beneficial in the real-world application scenarios. As ML models are sometimes sensitive to the data missingness, imputing the incomplete data in EHRs using generative approaches could improve the performance of ML models, and has become an area of active research \cite{yoon2018gain}.
%%%%
\red{Furthermore, as evaluations are performed based on clinical timeseries with a fixed length, no comparisons are made between the model's scalability when dealing with timeseries with varying lengths. Recent studies have found the quality of the synthetic longitudinal data degenerates over time, also called as the ``drift problem'' \cite{zhang2022keeping}. 
	Such problems when dealing with long sequences should be recognised and mitigated with techniques such as conditional fuzzing and regularization methods \cite{zhang2022keeping}, in both the generation and evaluation steps.}

\red{The evaluation of GANs is still a challenging task.
	Recent findings have suggested that systematical assessment for EHR synthesizers is critical before their applications in different use cases \cite{yan2022multifaceted}.
	In this study, a comprehensive evaluation list is provided with regards to the fidelity, correlation, utility and privacy of the synthesis models.
	It is also worth noting that evaluation metrics should be properly chosen and implemented based on the purpose of the task, otherwise may lead to biased results. 
	For example, recent findings \cite{zhang2022keeping} have reported that the traditional implementation of the discriminative score which trains the critic using the randomly initialised parameters, though widely used \cite{yoon2019time}, may lead to unreliable results. Improvement has been made to this evaluation metric for a more robust assessment, where the parameters of the trained generative model can be used for the critic's initialization.}

% limitation-3
Finally, the conditional aspect of our model is currently limited as it can not generate patient-specific EHRs conditioning on information at a more granular level.
Even though the proposed conditional GANs can synthesize a subgroup of patients with target outcomes or statuses that clinicians specify, it is still limited in incorporating personalised information during the conditional generation.
Future work for developing GANs in healthcare data can be extended to patient-level EHRs generation, such as synthesizing counterfactual information of a target patient for treatment effect estimation \mbox{\cite{yoon2018ganite, qian2021synctwin}}. Ultimately, by constructing the ``synthetic twin'' of patients using GANs, the synthesis tool can become more generalisable for precision medicine and support the clinical decision making in delivering personalized healthcare. 

%%Finally, although our models successfully extend the scope of synthetic data generation into mixed-type timeseries, they are still limited in single modality (i.e., structured EHR data). However, the adoption of digitized healthcare information systems provide a vast range of other --- heterogeneous databases with mixed modalities \cite{huang2020fusion}. Open-access data with various modalities are provided across the computational health community, including medical image data (such as magnetic resonance imaging (MRI) scans from BraTS \cite{menze2014multimodal}), physiological waveform signals (such as ECG and PPG data from MIMIC-III Waveform \cite{moody2020mimic}), as well as unstructured natural language information (such as free-text clinical notes from i2b2 platform \cite{n2c2link}). Future work can be investigated in the topic of generating synthetic data with multi-modalities, as integrating such data will increase the value of data-driven ML models in the real world clinical applications.

% conclusion
Synthetic data provides an alternative to sharing real patient data while preserving patient privacy. Results in our study demonstrate that the proposed EHR-M-GAN and EHR-M-GAN$_\texttt{cond}$ can generate realistic longitudinal EHR timeseries with mixed data types. By providing synthetic EHR data with higher fidelity and more variety, the proposed model can therefore enable faster development in AI-driven clinical tools with increased robustness and adaptability. In addition to the improved performance against the existing state-of-the-art benchmark models, augmentation provided by synthetic data during training boosts the predictive performance in downstream clinical tasks. EHR-M-GAN can help eliminate the barriers to data acquisition for healthcare studies, therefore overcoming the challenges posed by the paucity of medical data available and approved for research use.
Despite the novelty of this study in filling the research gap for synthesizing longitudinal EHRs in mixed-type settings, we acknowledge that there is still a gap between the real EHRs data and its synthetic counterparts produced by current generative methods. Therefore developing advanced EHR synthesizers especially in mixed-type settings still requires active research in the future study.

\bibliographystyle{agsm}
\bibliography{ref}

\end{document}

\maketitle
\section{Methodology.}
\subsection{Related work.}\label{meth:related_work}

Generative adversarial networks (GANs) have been used in EHR data synthesis, which can augment limited clinical data or even replace sensitive patient information.  
As longitudinal EHR data can capture patients' status over time, generative approaches for EHR data synthesis in the previous literature have been extended from static data to clinical timeseries generation. 
EHRs consist of a set of heterogeneous data types, such as continuous-valued and discrete-valued features.
%% continuous
When generating continuous-valued timeseries such as heart rate and respiratory rate in the critical care database, models such as C-RNN-GAN \mbox{\cite{mogren2016c}}, R(C)GAN \mbox{\cite{esteban2017real}} and TimeGAN \mbox{\cite{yoon2019time}} can be adopted. In order to model the temporal dynamics in the real-valued timeseries, recurrent neural networks (RNNs) such as long short-term memory (LSTM) are used as the generator and discriminator in their architectures. 
%% discrete
For synthesizing discrete-valued timeseries data such as diagnostic ICD-codes, GANs variants such as SynTEG \mbox{\cite{zhang2021synteg}}, \red{LS-EHR \cite{zhang2022keeping},} and DualAAE \mbox{\cite{lee2020generating}} models are proposed. For example, SynTEG generates time-stamped clinical events across patients' multiple visits. \red{Its amended version --- LS-EHR \cite{zhang2022keeping} model enhances the longitudinal EHR data synthesis by overcoming the performance drift through feedback mechanisms (including condition fuzzing, regularization and rejection sampling).}

\begin{figure*}[h!]
\captionsetup{justification=centering}
	\includegraphics[width=1.0\linewidth]{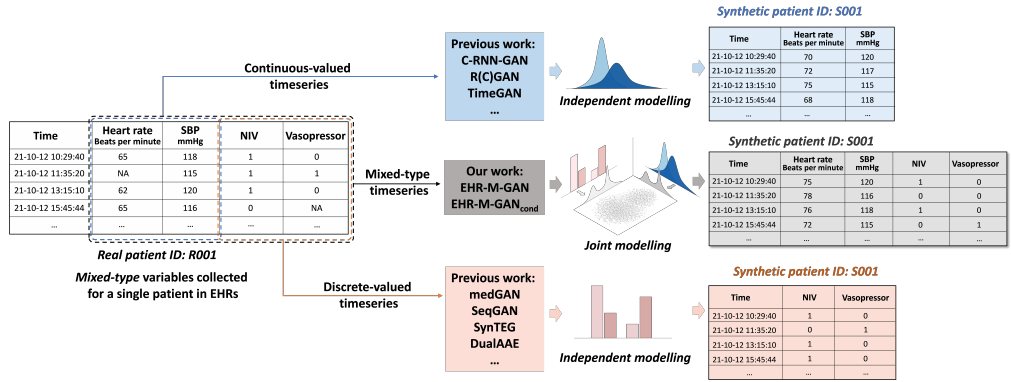}
	\caption[]{\textbf{A comparison of the proposed model with existing generative models.}} \label{fig_previous_work}
\end{figure*}

\red{As EHRs is an amalgamation of heterogeneous data types, previous work has demonstrated the importance of synthesizing mixed-type EHRs for various clinical applications. Several models have been proposed to generate static EHRs of mixed data types, such as discrete-valued medical concepts and continuous-valued measurements \cite{chin2019generation, yan2020generating, kroes2022generating}.}
\red{However, for synthesizing clinical timeseries,} most of the proposed models have been capable of synthesizing only a single data type (either continuous or discrete-valued timeseries separately). Consequently, previous work has tended to ignore the inter-dependencies among different data types, as shown in Fig. \mbox{\ref{fig_previous_work}}. 
In contrast, our proposed model can simultaneously generate both continuous-valued and discrete-valued timeseries, while capturing the inter-dependencies between the mixed-type data.

\subsection{Implementation of GANs using LSTMs}\label{meth:preli}
GAN consists of two networks that are adversarially trained to compete against each other. The recurrent neural networks (RNNs) are instantiated considering simulating the temporal structure for generating the sequential data.
As shown in Fig. 1 (b. Network architecture) in the main article, the generator $G$ accepts $\bm{\upsilon}_{1:T} \in \mathcal{T} \times \mathcal{V} $ as the input, which is a sequence of length $T$ sampled independently from a prior distribution \cite{esteban2017real}, \red{such as Gaussian distribution or uniform distribution. In this study, uniform distribution on the unit interval is chosen as the prior for sampling the random noise.} Then $G$ is optimized to approximate the distribution of true data, $p_{\mathbf{x}}$, by generating samples $\hat{\mathbf{x}}_{1:T}$ that are hard for the discriminator to distinguish from. 
Meanwhile, the discriminator $D$ is optimized to distinguish real samples $\mathbf{x}_{1:T}$ from synthetic samples $\hat{\mathbf{x}}_{1:T}$. Overall, the training of GAN is a minmax game with the following objective function:
\begin{equation}\label{eq1}
	\min _{G} \max _{D} V_{\mathrm{GAN}}= 
	\mathbb{E}_{ \mathbf{x} \sim p_{ \mathbf{x} } } [\log D( \mathbf{x} )] +\mathbb{E}_{\bm{\upsilon} \sim p_{\bm{\upsilon}}  }[\log (1-D(G(\bm{\upsilon})))]
\end{equation}

Conditional GAN is the extension case of GAN, where both the generator $G$ and discriminator $D$ receive conditional information $\mathbf{y} \in \mathcal{L} = \{1, 2, ..., |L|\}$ from $|L|$ classes \cite{esteban2017real}. 
In other words, the inputs are augmented by being concatenated with $\mathbf{y}$ at each timestamp, i.e., $\mathbf{x}_{1:T} \to [\mathbf{y}; \mathbf{x}_{1:T}]$.
This formulation allows $G$ to generate samples conditioned on the auxiliary information of $|L|$-dimensional categorical labels.  In this case, the objective function becomes:
\begin{equation}\label{eq2}
	\begin{split}
	\min _{G} \max _{D} V_{\mathrm{CGAN}} = 
	&\mathbb{E}_{\mathbf{y},\mathbf{x} \sim p_{ \mathbf{y},\mathbf{x} }}
	[\log D(\mathbf{x} | \mathbf{y})]\\
	+ & \mathbb{E}_{\mathbf{y} \sim p_{\mathbf{y}}, \bm{\upsilon} \sim p_{{ \bm{\upsilon} }}   } 
	[\log (1-D( G(\mathbf{y}, \bm{\upsilon}) | \mathbf{y} ))]
	\end{split}
\end{equation}

\subsection{Shared latent space learning using dual-VAE.}
\begin{figure*}[h!]
	\centering
	\captionsetup{justification=centering}
	\includegraphics[width=0.63\linewidth]{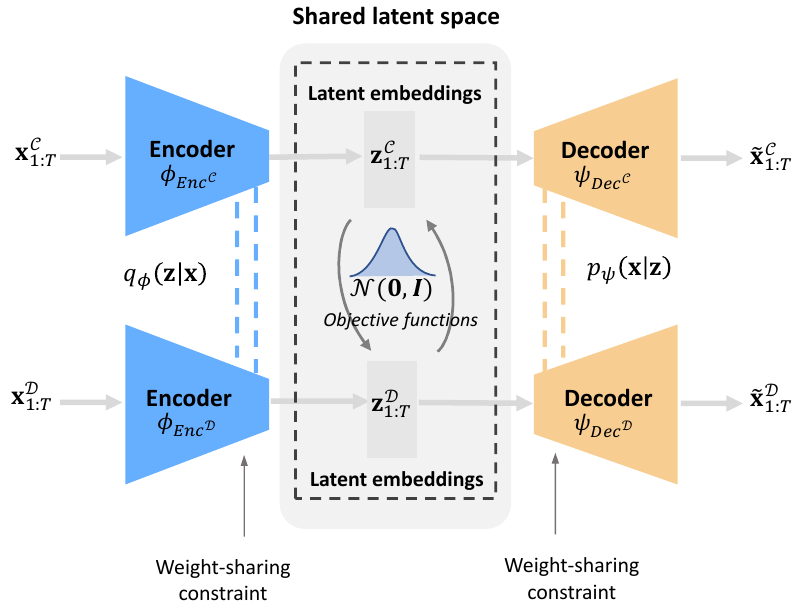}
	\caption{\textbf{The network architecture of dual-VAE during the pretraining stage.}} \label{fig_sl}
\end{figure*}

As shown in Fig. \ref{fig_sl}, the shared latent space is learnt by a dual-VAE network, which contains a pair of encoders (parameterized as $\smash{\phi_{\mathit{Enc}^{\mathcal{C}}}}$ and $\smash{\phi_{\mathit{Enc}^{\mathcal{D}}}}$), and a pair of decoders (parameterized as $\smash{\psi_{\mathit{Dec}^{\mathcal{C}}}}$ and $\smash{\psi_{\mathit{Dec}^{\mathcal{D}}}}$) of VAE networks, one for each type of timeseries. 
\red{We found VAE preferable to vanilla autoencoder in our case, considering that (1) the KL regularization in VAE strengthens the learning of the compressed latent representations, which further narrows the domain gap for mixed-type features \cite{wan2020old}; (2) VAE can be easily extended to the conditional learning scenario in EHR-M-GAN$_{\mathtt{cond}}$.} The encoders map the observations into the latent space with $ \mathit{Enc}(\mathbf{x}) \triangleq q_{\phi}(\mathbf{z}|\mathbf{x}) $, while the decoders further map the representations into the reconstructed input with $ \mathit{Dec}(\mathbf{z}) \triangleq p_{\psi}(\mathbf{x}|\mathbf{z}) $.
During the implementation, we found that except for pretraining the dual-VAE, integrating the optimization for decoders during the joint training stage also benefit the generative model from learning an improved representations in the shared latent space. 

In dual-VAE, we enforce a weight-sharing constraint \cite{liu2017unsupervised} across certain layers within both the encoders pairs and decoders pairs to further eliminate the gap between domains (see Fig. \ref{fig_sl}). To be specific, only weights of the last few layers of the encoders and the first few layers of the decoders are shared \cite{liu2016coupled}. This forces the encoders to derive the same high-level representations while maintaining different low-level realizations. Meanwhile, it forces the decoders to share the same high-level semantics and decode them into different low-level feature space observations.

\subsection{Comparison between LSTM and Bilateral-LSTM} 
% The SRN is implemented with Long short-term memory (LSTM) network, which is capable of capturing temporal dynamics among the patient trajectories.
% Each SRN consists of three layers: (i) At each timestamp $t \in \{1, 2, ..., T \}$, the \textit{input layer} takes the random noise $\bm{\upsilon}_{t}$ as the input, which is normally sampled from Gaussian distribution or uniform distribution (e.g., $\bm{\upsilon}_{t} \in \mathcal{U}(0,1)$), and then feed it into the following LSTM network;
% (ii) The \textit{recurrent layer} $f_{\mathtt{rec}}$ learns a mapping from $\bm{\upsilon}_{t}$ and previous hidden state $\mathbf{h}_{t-1}$ to $\mathbf{h}_{t}$ at current stage, by applying the corresponding transition functions;
% (iii) At each stage, the output from the
% LSTM cell is fed into a \textit{fully connected layer} $f_{\mathtt{conn.}}$ with weights shared across timesteps to obtain the generated latent codes
% $\hat{\mathbf{z}}_{t}$:
% \begin{equation}{
% 	\begin{aligned}
% 	\mathbf{h}_{t} &= f_{\mathtt{rec}}(
% 	\bm{\upsilon}_{t},
% 	\mathbf{h}_{t-1}
% 	)\\
% 	\hat{\mathbf{z}}_{t} &= f_{\mathtt{conn}}(\mathbf{h}_{t})
% 	\end{aligned}
% }
% \end{equation}

% The abovementioned steps will recursively process the timeseries with the length of $T$. 
To better compare with BLSTMs, we elaborate the architecture of the LSTM network.
LSTM utilizes three gates to control the cell state in order to mitigate the problems of gradient vanishing and exploding that appears in the recurrent neural network (RNN) --- an input gate $\mathbf{i}_{t}$ that controls the amount of input information to be passed along into the memory cell, a forget gate $\mathbf{f}_{t}$ which controls the amount of past information to be neglected, and an output gate $\mathbf{o}_{t}$ which controls the update of the new memory cell.
The range of outputs from $\mathbf{i}_{t}, \mathbf{f}_{t}$ and $\mathbf{o}_{t}$ are limited by $[0,1]$ due to the sigmoid activation function. At each time step $t$, the transition functions in LSTM are as follows:
\begin{equation}\label{eq3}
	\begin{aligned}
	\mathbf{i}_{t} &=\sigma\left(\mathbf{W}_{iv} \bm{\upsilon}_{t}+\mathbf{W}_{ih} \mathbf{h}_{t-1}+\mathbf{b}_{i}\right) \\
	\mathbf{f}_{t} &=\sigma\left(\mathbf{W}_{fv} \bm{\upsilon}_{t}+\mathbf{W}_{fh} \mathbf{h}_{t-1}+\mathbf{b}_{f}\right) \\
	\mathbf{o}_{t} &=\sigma\left(\mathbf{W}_{ov} \bm{\upsilon}_{t}+\mathbf{W}_{oh} \mathbf{h}_{t-1}+\mathbf{b}_{o}\right) \\
	\tilde{\mathbf{c}}_{t} &=\tanh \left(\mathbf{W}_{cv} \bm{\upsilon}_{t}+\mathbf{W}_{ch} \mathbf{h}_{t-1}+\mathbf{b}_{c}\right) \\
	\mathbf{c}_{t} &=\mathbf{f}_{t} \odot \mathbf{c}_{t-1}+\mathbf{i}_{t} \odot \tilde{\mathbf{c}}_{t} \\
	\mathbf{h}_{t} &=\mathbf{o}_{t} \odot \tanh \left(\mathbf{c}_{t}\right)
	\end{aligned}
\end{equation}
where $\mathbf{c}_{t}$ denotes the context vector, $\sigma$ denotes the sigmoid activation function, and $\odot$ denotes the operation of element-wise multiplication.

\begin{figure*}[h!]
	\centering
	\captionsetup{justification=centering}
	\includegraphics[width=0.7\linewidth]{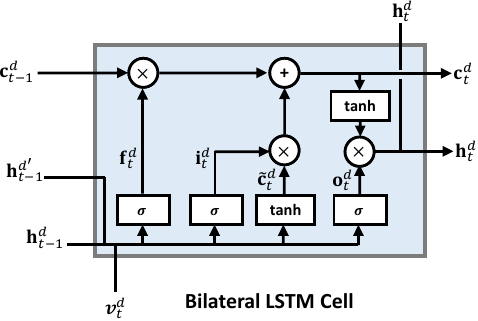}
	\caption{\textbf{Illustration of BLSTM cell.}}\label{fig_lstm1}
\end{figure*}

Based on the basic structure of LSTM, the Bilateral Long Short-Term Memory (BLSTM) network is proposed (see Fig. \ref{fig_lstm1}). Equations that demonstrate the calculation of BLSTM units can be found in \textit{Methodology} section in the main article. 

\newpage
\subsection{Algorithms.}
\begin{algorithm}[h!]
	\begin{algorithmic}[1]
	\State \textbf{Input:} $\mathscr{D} = \{(\mathbf{x}_{i, 1:T}^{\mathcal{C}},     \mathbf{x}_{i, 1:T}^{\mathcal{D}})\}_{i=1}^{N}$, learning rate $\eta_{\mathrm{VAE}}$, scalar loss weights $\beta_0$, $\beta_1$, $\beta_2$, $\beta_3$ (if \texttt{conditional}), minibatch size $n_{mb}$.
	\vspace{1ex}
	\State Initialize parameters: 
	$\phi_{\mathrm{Enc}}^{\mathcal{C}}$,
	$\phi_{\mathrm{Enc}}^{\mathcal{D}}$,
	$\psi_{\mathrm{Dec}}^{\mathcal{C}}$,
	$\psi_{\mathrm{Dec}}^{\mathcal{D}}$
	\vspace{1ex}
	\For{number of pretrain iterations}\\
		\hspace{12bp} \red{Sample a minibatch of $n_{mb}$ data samples:} $\{(\mathbf{x}_{i, 1:T}^{\mathcal{C}}, \mathbf{x}_{i, 1:T}^{\mathcal{D}})\}_{i=1}^{n_{mb}}\stackrel{i.i.d.}{\sim} \mathcal{D}$
		\vspace{1ex}
  
		\hspace{-6bp} {\commenttext{Map between features and latent representations:}} \vspace{0.5ex}
		\For{$i=1, 2, ..., n_{mb}, t=1, 2, ..., T$}\\
			\vspace{1ex}
			% encoder get the latent
			\hspace{24bp}
			$(\mathbf{z}_{i, t}^{\mathcal{C}},
			\mathbf{z}_{i, t}^{\mathcal{D}}) = 
			(\mathit{Enc}^{\mathcal{C}}(\mathbf{x}_{i, t}^{\mathcal{C}},
			\mathbf{z}_{i, t-1}^{\mathcal{C}}),
			\mathit{Enc}^{\mathcal{D}}(\mathbf{x}_{i, t}^{\mathcal{D}}, \mathbf{z}_{i, t-1}^{\mathcal{D}}))$\\
			\vspace{1ex}
			\hspace{24bp}		
			$(\tilde{\mathbf{x}}_{i, t}^{\mathcal{C}},
			\tilde{\mathbf{x}}_{i, t}^{\mathcal{D}}) = 
			(\mathit{Dec}^{\mathcal{C}}(\mathbf{z}_{i, t}^{\mathcal{C}}),
			\mathit{Dec}^{\mathcal{D}}(\mathbf{z}_{i, t}^{\mathcal{D}}))$
			\vspace{1ex}
		\EndFor 
  
		\hspace{-6bp} {\commenttext{Estimate the loss terms:}} \vspace{0.5ex}
		\For{$d \in \{\mathcal{C}, \mathcal{D} \} \vspace{1ex}$} \\
				\hspace{24bp}		
				$\mathcal{L}^{\mathrm{ELBO}}_{d}=\frac{1}{n_{m b}} \sum_{i=1}^{n_{m b}}[
				-\mathbb{E}_{q_{\phi}(\mathbf{z}|\mathbf{x})}[\log p_{\psi}(\mathbf{x} | \mathbf{z})] + 
				\beta_{\mathrm{KL}} D_{\mathrm{KL}}(q_{\phi}(\mathbf{z} | \mathbf{x}) \| p_{\psi}(\mathbf{z}))] $
			\vspace{1ex} \\
			\hspace{24bp}
			$\mathcal{L}^{\mathrm{Match}}=
			\frac{1}{n_{m b}} \sum_{i=1}^{n_{m b}}[
			\mathbb{E}_
			{\mathbf{z} \sim p_{\mathbf{z}}		
			}[
			\sum_{t \in \mathcal{T}}
			||
			\mathbf{z}^{\mathcal{C}}_{t}
			-
			\mathbf{z}^{\mathcal{D}}_{t} ||^{2}] ]
			$
			\vspace{1ex} \\ 
			\hspace{24bp}
			$\mathcal{L}^{\mathrm{Contra}}=\frac{1}{2 n_{m b}} \sum^{n_{m b}}_{i^{d}=1}
			\sum^{n_{m b}}_{i^{d^{\prime}}=1}
			[\mathcal{L}_{i^{d}, i^{d^{\prime}}}^{\mathrm{Contra}}+\mathcal{L}_{i^{d^{\prime}}, i^{d}}^{\mathrm{Contra}}]
			$	
			\vspace{1ex} \\
			\hspace{24bp}
			$\mathcal{L}_{d} = \beta_0 \mathcal{L}^{\mathrm{ELBO}}_{d} + \beta_1 \mathcal{L}^{\mathrm{Match}} + \beta_2 \mathcal{L}^{\mathrm{Contra}}$		
			\vspace{1ex}\\
			\hspace{24bp} if \texttt{conditional}:\vspace{1ex}\\
			\hspace{36bp}
			$\mathcal{L}_{d}^{\mathrm{Class}}=
			\frac{1}{n_{m b}} \sum_{i=1}^{n_{m b}}[
			\mathbb{E}_{\mathbf{z}^d \in 
				\mathcal{H}^{\mathcal{S}}
			} \mathrm{CE}\left(f^d_{\mathtt{linear}}(\mathbf{z}^{d}), 
			\mathbf{y}
			\right) ]$\vspace{1ex}\\
			\hspace{36bp}
			$\mathcal{L}_{d} = \beta_0\mathcal{L}^{\mathrm{ELBO}}_{d} + \beta_1\mathcal{L}^{\mathrm{Match}} + \beta_2\mathcal{L}^{\mathrm{Contra}} + \beta_3\mathcal{L}_{d}^{\mathrm{Class}}$
			\EndFor
			\vspace{1ex}
   
			\hspace{8bp} \commenttext{Update the network weights:} \vspace{0.5ex} \\
			\hspace{24bp} 
			$\phi_{\mathrm{Enc}}^{\mathcal{C}}=\operatorname{Adam}\left(\frac{\partial \mathcal{L}^{\mathcal{C}}_{\mathrm{VAE}}}{\partial \phi_{\mathrm{Enc}}^{\mathcal{C}}}, \eta_{\mathrm{VAE}}\right)$,		
			$\psi_{\mathrm{Dec}}^{\mathcal{C}}=\operatorname{Adam}\left(\frac{\partial \mathcal{L}^{\mathcal{C}}_{\mathrm{VAE}}}{\partial \psi_{\mathrm{Dec}}^{\mathcal{C}}}, \eta_{\mathrm{VAE}}\right)$ 
			\vspace{1ex}\\
			\hspace{24bp} 
			$\phi_{\mathrm{Enc}}^{\mathcal{D}}=\operatorname{Adam}\left(\frac{\partial \mathcal{L}^{\mathcal{D}}_{\mathrm{VAE}}}{\partial \phi_{\mathrm{Enc}}^{\mathcal{D}}}, \eta_{\mathrm{VAE}}\right)$,	
			$\psi_{\mathrm{Dec}}^{\mathcal{D}}=\operatorname{Adam}\left(\frac{\partial \mathcal{L}^{\mathcal{D}}_{\mathrm{VAE}}}{\partial \psi_{\mathrm{Dec}}^{\mathcal{D}}}, \eta_{\mathrm{VAE}}\right)$	
	\EndFor
	\vspace{1ex}
	% return the output
	\State \textbf{Return:} ${\psi_{\mathrm{Dec}}^{\mathcal{C}}}$, $\psi_{\mathrm{Dec}}^{\mathcal{D}}$
		
	\end{algorithmic}
	\caption{Algorithm of dual-VAE for pretraining.}
\end{algorithm}

\newpage

\begin{algorithm}[h!]
	\begin{algorithmic}[1]
	\State \textbf{Input:} {$\mathscr{D} = \{ (\mathbf{x}_{i, 1:T}^{\mathcal{C}},     \mathbf{x}_{i, 1:T}^{\mathcal{D}}) \}_{i=1}^{N}$, pretrained decoder in dual-VAE for both domains $\psi_{\mathrm{Dec}}^{\mathcal{C}}$, 			$\psi_{\mathrm{Dec}}^{\mathcal{D}}$, learning rate $\eta_{\mathrm{GAN}}$, minibatch size $n_{mb}$} \vspace{1ex}
	\State Initialize parameters: $\theta_G^{\mathrm{CRN}}$, $\mu_D^{\mathcal{C}}$, $\mu_D^{\mathcal{D}}$.\vspace{1ex}
	
		\For{number of training iterations}\\
		% sample latent noise
		\hspace{12bp} \red{Sample a minibatch of $n_{mb}$ random noise samples:} $\{( \bm{\upsilon}_{i, 1:T}^{\mathcal{C}},   \bm{\upsilon}_{i, 1:T}^{\mathcal{D}} )\}_{i=1}^{n_{mb}}\stackrel{i.i.d.}{\sim} \mathcal{V}$ \vspace{1ex}
		\For{$i=1, 2, ..., n_{mb}, t=1, 2, ..., T$}\vspace{1ex} 
  
			\hspace{8bp} \commenttext{Generate synthetic latent codes using coupled-generator:} \vspace{0.5ex}\\				
			% put it into the generator			
			\hspace{24bp} $ (\hat{\mathbf{z}}_{i, t}^{{\mathcal{C}}}, \hat{\mathbf{z}}_{i, t}^{{\mathcal{D}}}) = G^{\mathrm{CRN}}((\bm{\upsilon}_{i, t}^{{\mathcal{C}}}, \bm{\upsilon}_{i, t}^{{\mathcal{D}}}), 
			(\mathbf{h}_{i, t-1}^{\mathcal{C}}, \mathbf{h}_{i, t-1}^{\mathcal{D}} )) $ \vspace{1.5ex} 
   
			\hspace{8bp} \commenttext{Decode generated latent codes into observational space}: \vspace{0.5ex} \\		
			\hspace{24bp} $(\hat{\mathbf{x}}_{i, t}^{\mathcal{C}},
			\hat{\mathbf{x}}_{i, t}^{\mathcal{D}}) = 
			(\mathit{Dec}^{\mathcal{C}}(\hat{\mathbf{z}}_{i, t}^{\mathcal{C}}),
			\mathit{Dec}^{\mathcal{D}}(\hat{\mathbf{z}}_{i, t}^{\mathcal{D}}))$ \vspace{1ex}
		\EndFor	\\
		\hspace{12bp} \red{Sample a minibatch of $n_{mb}$ real data samples:} $\{ ({\mathbf{x}}_{i, 1:T}^{\mathcal{C}}, {\mathbf{x}}_{i,  1:T}^{\mathcal{D}})\}_{i=1}^{n_{mb}}\stackrel{i.i.d.}{\sim} \mathcal{D}$, \red{and a minibatch 
  
        \hspace{-6bp} of $n_{mb}$ synthetic data samples} $\{ (\hat{\mathbf{x}}_{i, 1:T}^{\mathcal{C}}, \hat{\mathbf{x}}_{i,  1:T}^{\mathcal{D}})\}_{i=1}^{n_{mb}}\stackrel{i.i.d.}{\sim} \mathcal{D}$\vspace{2ex}	
  
		\hspace{-6bp} \commenttext{Distinguish real and fake samples using discriminators and estimate loss}: \vspace{1ex}\\
        \hspace{12bp} ${\mathcal{L}}_{\mathrm{GAN}}=\frac{1}{n_{m b}} \sum_{n=1}^{n_{m b}}[
        \log (D^{\mathcal{C}}({\mathbf{x}}_{i}^{\mathcal{C}})) +
        \log (D^{\mathcal{D}}({\mathbf{x}}_{i}^{\mathcal{D}}))]
        + $\\\vspace{0.5ex}
        $
        \hspace{86bp} [\log (1- D^{\mathcal{C}}(\hat{\mathbf{x}}_{i}^{\mathcal{C}}) ) +
        \log(1-D^{\mathcal{D}}(\hat{\mathbf{x}}_{i}^{\mathcal{D}}) ] 
        $\vspace{2ex}

		\hspace{-6bp} \commenttext{Update network weights via Adam optimizer}: \vspace{1ex}\\		
		\hspace{12bp}  $\theta_G^{\mathrm{CRN}}=\operatorname{Adam}\left(\frac{\partial \mathcal{L}_{\mathrm{GAN}}}{\partial \theta_G^{\mathrm{CRN}}   }, \eta_{\mathrm{GAN}}\right)$ \vspace{1ex} \\
		\hspace{12bp}  $\mu_D^{\mathcal{C}}=\operatorname{Adam}\left(\frac{\partial \mathcal{L}_{\mathrm{GAN}}}{\partial \mu_D^{\mathcal{C}}}, \eta_{\mathrm{GAN}}\right)$, $\,$ 	
		$\mu_D^{\mathcal{D}}=\operatorname{Adam}\left(\frac{\partial \mathcal{L}_{\mathrm{GAN}}}{\partial \mu_D^{\mathcal{D}} }, \eta_{\mathrm{GAN}}\right)$					
		\EndFor	
	% generate real data
	\vspace{1ex}
 
	\hspace{-20bp} \commenttext{Synthesize M pairs of coupled mixed-types of features for M patients:}\\
	% sample latent noise
	Sample $\{( \bm{\upsilon}_{i, 1:T}^{\mathcal{C}},   \bm{\upsilon}_{i, 1:T}^{\mathcal{D}} )\}_{i=1}^{M}\stackrel{i.i.d.}{\sim} \mathcal{V}$ 
	\vspace{1ex}
	\For{$i=1, 2, ..., M, t=1, 2, ..., T$ \vspace{1ex}}	\\	
		\hspace{12bp} 			
		$ (\hat{\mathbf{z}}_{i, t}^{{\mathcal{C}}}, \hat{\mathbf{z}}_{i, t}^{{\mathcal{D}}}) = G^{\mathrm{CRN}}((\bm{\upsilon}_{i, t}^{{\mathcal{C}}}, \bm{\upsilon}_{i, t}^{{\mathcal{D}}}), 
		(\mathbf{h}_{i, t-1}^{\mathcal{C}}, \mathbf{h}_{i, t-1}^{\mathcal{D}} )) $ \vspace{1ex}\\	
		\hspace{12bp} 	
		$(\hat{\mathbf{x}}_{i, t}^{\mathcal{C}},
		\hat{\mathbf{x}}_{i, t}^{\mathcal{D}}) = 
		(\mathit{Dec}^{\mathcal{C}}(\hat{\mathbf{z}}_{i, t}^{\mathcal{C}}),
		\mathit{Dec}^{\mathcal{D}}(\hat{\mathbf{z}}_{i, t}^{\mathcal{D}}))$ \vspace{1ex}
	\EndFor
	
	% return the output
	\State \textbf{Return:} $\hat{\mathscr{D}} = \{ (\hat{\mathbf{x}}_{i, 1:T}^{\mathcal{C}}, \hat{\mathbf{x}}_{i, 1:T}^{\mathcal{D}}) \}_{i=1}^{M}$

	\end{algorithmic}
\caption{Algorithm of EHR-M-GAN.}
\end{algorithm}

\newpage
\section{Datasets.}
We construct the pipeline of data preprocessing based on the work of MIMIC-Extract \cite{wang2020mimic}. Three large-scale, publicly available datasets --- MIMIC-III, eICU, and HiRID are processed based on the standard pipeline. The complete steps for data preprocessing include:
\begin{itemize}
	\item Cohort selection: In cohort selection, patients in three ICU databases are selected based on the same predefined criteria (see Section 2.\ref{sec:cohort} for details). 
	
	\item Timeseries features extraction: Then, the timeseries features are extracted based on the lists provided in Section 2.\ref{sec:featurelist}. Both continuous-valued and discrete-valued features are selected accordingly. 
	
	\item Unit conversion and outlier filtering: Due to the fact that clinical data is often measured in different units, unit conversion are applied (such as converting Fahrenheit to Celsius for \textit{Temperature}). For outlier filtering, a reasonable physiologically valid range are applied for different measurements (see \cite{wang2020mimic} for details). 
	
	\item Semantic grouping: Next, semantically similar variables are grouped based on clinical concepts (such as \textit{Heart Rate} is recorded as ItemID 211 in CareVUE EHR systems and ItemID 220045 under MetaVision EHR systems). A clinical taxonomy are used to aggregate features that are semantically equivalent \cite{wang2020mimic}.
	
	\item Hourly aggregation: Timestamps with different granularity are provided for different in three databases. Time-varying physiological signals such as \textit{Heart Rate} are frequently measured (e.g., most parameters under bedside monitoring are recorded every 2 minutes in HiRID dataset \cite{yeche2021hirid}). While other features such as laboratory test results are measured infrequently. Therefore, we hourly aggregate the timeseries further into a uniform hourly bucket.

	\item Imputation and normalization: Finally, imputation method in Section \ref{sec:imputation} are used and normalization are applied to obtain the final result of the data matrix.
\end{itemize}

\subsection{Cohort selection criteria.}\label{sec:cohort}
In line with the previous literature \cite{wang2020mimic, wu2017understanding}, the cohort are selected based on the following criteria:
(1) Only the first known ICU admission of the patient is selected. This is because the patients who have multiple ICU admission records typically require specific treatments for life-support intervention; (2) Patient has to be an adult at the time of ICU admission (at least 15); (3) The duration of the patients' ICU stay is at least 12 hours and less than 10 days. This is because the treatment for patients who have longer hours in the ICU stay usually indicates their physiological changes can not be directly linked to the positive effect of the treatment (but compensating for the life support treatment being taken off) \cite{wu2017understanding}.

\subsection{Imputation method.}\label{sec:imputation}
\hl{For continuous-valued timeseries, missing data is imputed based on method of \mbox{\textit{Simple Imputation}} \mbox{\cite{che2018recurrent}}.} The missing timeseries data is imputed as the last observed value, or individual-specific mean if no previous observation is provided. Else, if there is no observation for the subject, the imputation value is set to the global mean of the entire cohort. 
\hl{Compared to imputation methods developed upon customized RNN models or explicitly designed for the applied domains, it does not rely on additional information such as the prediction labels therefore more generalizable. Though simple, such method has been widely applied in clinical timeseries analysis \mbox{\cite{meyer2018machine}} including MIMIC-III datasets \mbox{\cite{wang2020mimic, che2018recurrent, purushotham2018benchmarking}}.}
\hl{For discrete-valued timeseries, we followed the preprocessing rules in MIMIC-EXTRACT. For intermittent interventions such as oral antibiotics, its status is regarded as `not applied' when missing. For intervention with multi-hour continuous duration, such as mechanical ventilation, the missed status is considered to be consistent with the previous status until the new administration occurs. Therefore, the imputation method was not applied to the discrete-valued data.}

\subsection{Timeseries features extraction.}\label{sec:featurelist}
Features of continuous-valued and discrete-valued timeseries are extracted for three critical care databases based on the following lists (for MIMIC-III dataset, see Table S1, S2; for eICU dataset see Table S3, S4; for HiRID dataset, see Table S5, S6).

\begin{table}[H]
	\caption{List of vital sign and laboratory test features for MIMIC-III dataset. Features are further extracted based on the preprocessed results of MIMIC-Extract (see Appendix A. Feature set in  \cite{wang2020mimic}). \red{The dimension of continuous-valued features for MIMIC-III dataset during model's training is 78.}  }
	\vspace{2ex}
	\resizebox{.99\textwidth}{!}{
		\begin{tabular}{p{5cm} p{5cm} p{5cm} p{5cm}}
			\toprule
			\textbf{Measurement} &  & & \\
			\midrule
			heart rate & respiratory rate & systolic blood pressure & diastolic blood pressure \\
			mean blood pressure & oxygen saturation  & temperature  &  glucose\\
			central venous pressure & hematocrit & potassium & sodium\\
			chloride & pulmonary artery pressure systolic & hemoglobin & ph\\
			creatinine & blood urea nitrogen & bicarbonate & platelets\\
			anion gap & co2 (etco2, pco2, etc.) & partial pressure of carbon dioxide & magnesium\\
			white blood cell count & positive end-expiratory pressure set & calcium & fraction inspired oxygen set\\
			red blood cell count & 
			mean corpuscular hemoglobin concentration & 
			mean corpuscular hemoglobin & 
			mean corpuscular volume \\
			tidal volume observed & 
			partial thromboplastin time & 
			prothrombin time inr & 
			prothrombin time pt \\
			phosphate & 
			phosphorous & 
			peak inspiratory pressure & 
			calcium ionized \\
			respiratory rate set & 
			fraction inspired oxygen & 
			tidal volume set & 
			partial pressure of oxygen \\
			cardiac index & 
			co2 & 
			systemic vascular resistance & 
			potassium serum \\
			tidal volume spontaneous & 
			plateau pressure & 
			pulmonary artery pressure mean & 
			cardiac output thermodilution \\
			lactate & 
			lactic acid & 
			bilirubin & 
			asparate aminotransferase \\
			alanine aminotransferase & 
			alkaline phosphate & 
			positive end-expiratory pressure & 
			albumin \\
			troponin-t & 
			neutrophils & 
			lymphocytes & 
			monocytes \\
			ph urine & 
			fibrinogen & 
			lactate dehydrogenase & 
			basophils \\
			cardiac output fick & 
			creatinine urine & 
			pulmonary capillary wedge pressure & 
			red blood cell count urine \\
			white blood cell count urine & 
			cholesterol & 
			cholesterol hdl & 
			post void residual \\
			cholesterol ldl & 
			chloride urine & &\\
			\bottomrule
		\end{tabular}
	}
\end{table}

\newpage
\begin{table}[H]
	\caption[Caption for LOF1]{List of medical intervention features for MIMIC-III dataset, \red{where \textbf{Features} indicates the name of the intervention features during model's training, \textbf{Category of treatment} shows the category of treatment that the specific intervention feature belongs to, and \textbf{Source} is the corresponding chart(s) where the variable is extracted based on\footnotemark. The dimension of discrete-valued features for MIMIC-III dataset during model's training is 20.}  }
	\vspace{2ex}
	\resizebox{.99\textwidth}{!}{
		\begin{tabular}{l|l|l}
			\toprule
			\textbf{\red{Category of treatment}} & \textbf{\red{Features}} & \textbf{Source}\\
			\midrule
			Oxygen therapy &  
			supplemental oxygen & \texttt{chartevents, procedureevents\_mv} \\
			& mechanical ventilation & \\
			\midrule
			Vasopressor & adenosine & \texttt{inputevents\_cv, inputevents\_mv} \\
			& dobutamine & \\
			& dopamine & \\
			& epinephrine & \\
			& isuprel & \\
			& milrinone & \\
			& norepinephrine & \\
			& phenylephrine & \\
			& vasopressin & \\
			\midrule
			Antibiotics & antibiotics & \texttt{prescriptions} \\
			\midrule
			Renal therapy & continuous renal replacement therapy & \texttt{chartevents} \\
			\midrule
			Invasive lines 
			& arterial line & \texttt{procedureevents\_mv}, \texttt{chartevents} \\
			& central line &\\
			\midrule
			Colloid bolus & colloid bolus & \texttt{inputevents\_mv, inputevents\_cv, chartevents} \\
			\midrule
			Crystalloid bolus & crystalloid bolus & \texttt{inputevents\_mv, inputevents\_cv } \\
			
			\bottomrule
		\end{tabular}
	}
\end{table}
\footnotetext{https://github.com/MIT-LCP/mimic-code}

\newpage
\begin{table}[H]
	\caption{List of vital sign and laboratory test features for eICU dataset. Features are selected base on the recommendation from Rocheteau et al \cite{rocheteau2021temporal}. \red{The dimension of continuous-valued features for eICU dataset during model's training is 55.}}
	\vspace{2ex}
	\resizebox{.99\textwidth}{!}{
		\begin{tabular}{p{5cm} p{5cm} p{5cm} p{5cm}}
			\toprule
			\textbf{Measurement} &  & & \\
			\midrule
			 Hct & calcium & anion gap & MCH\\
			 troponin - I  & MCHC & PT & PT - INR\\
			 -eos & potassium & -basos & albumin\\
			 -polys & lactate & glucose & creatinine \\
			 AST (SGOT) & Hgb & MPV & WBC $\times$ 1000 \\
			 ALT (SGPT) & HCO3 & MCV & -lymphs \\
			 Exhaled MV & RDW & chloride & sodium \\
			 bicarbonate & pH & urinary specific gravity & SaO2\\ Tidal Volume (set) & -monos & Heart Rate & BUN\\ platelets $\times$ 1000 & total bilirubin & Exhaled TV (patient) & alkaline phos \\
			 Noninvasive BP Diastolic & Noninvasive BP Mean  & Noninvasive BP Systolic & Base Excess \\ 
			 paO2 & FiO2 & Temperature & RBC \\
			 PTT & magnesium & RR & SpO2\\
			 total protein & paCO2 & phosphate & \\
			\bottomrule
		\end{tabular}
	}
\end{table}

\newpage
\begin{table}[H]
	\caption[Caption for LOF2]{List of medical intervention features for eICU dataset, \red{where \textbf{Features} indicates the name of the intervention features during model's training, \textbf{Category of treatment} shows the category of treatment that the specific intervention feature belongs to, and \textbf{Source} is the corresponding chart(s) where the variable is extracted based on\footnotemark. The dimension of discrete-valued features for eICU dataset during model's training is 19.}}
	\vspace{2ex}
	\resizebox{.98\textwidth}{!}{
		\begin{tabular}{l|l|l}
			\toprule
			\textbf{\red{Category of treatment}} & \textbf{\red{Features}} & \textbf{Source}\\
			\midrule
			
			Oxygen therapy &  
			supplemental oxygen & \texttt{respiratorycharting, nursecharting, treatment} \\
			& mechanical ventilation & \\
			\midrule
			
		    Vasopressor &
		    dopamine & \texttt{infusionDrug} \\
		    & epinephrine &\\
		    & norepinephrine &\\
		    & phenylephrine &\\
		    & vasopressin &\\
		    & milrinone &\\
		    & dobutamine &\\
		    \midrule
		    
		    Anesthesia & 
		    fentanyl & \texttt{infusionDrug} \\
		    & propofol &\\
		    & midazolam &\\
		    & dexmedetomidine &\\
		    \midrule
		    
		    Anticoagulants & 
		    heparin & \texttt{infusionDrug} \\
		    \midrule
		    
		    Insulin & 
		    insulin & \texttt{infusionDrug} \\		    
			\midrule
			
			Antibiotics & 
			antibiotics & \texttt{medication} \\	
			
			\bottomrule
		\end{tabular}
	}
\end{table}
\footnotetext{https://github.com/MIT-LCP/eicu-code}

\newpage
\begin{table}[H]
	\caption[Caption for hi]{List of vital sign and laboratory test features for HiRID dataset. Features are extracted based on the official HiRID preprocessing codes (meta-variables from \texttt{Merging stage} \footnotemark) \cite{yeche2021hirid}. \red{The dimension of continuous-valued features for HiRID dataset during model's training is 50.}}
	\vspace{2ex}
	\resizebox{.99\textwidth}{!}{
		\begin{tabular}{p{4cm} p{4cm} p{4cm} p{4cm}}
			\toprule
			\textbf{Measurement} &  & & \\
			\midrule
			HR  &  T Central & ABPs &  ABPd \\
			ABPm & NIBPs & NIBPd   &  NIBPm \\
			PAPm   &  PAPs & PAPd &  CO \\
			SvO2(m)  & ZVD  & ST1  &  ST2  \\
			ST3  &  SpO2  & ETCO2 &  RR  \\
			OUTurine/h  & ICP  & Liquor/h &  a-BE \\
			a\_COHb  &  a\_Hb  & a\_HCO3-  &  a\_Lac \\
			a\_MetHb  &   a\_pH   & a\_pCO2  &   a\_PO2 \\
			a\_SO2  &   K+ & Na+  &   Cl-   \\
			Ca2+ ionizied  &   phosphate  & Mg\_lab   &   Urea \\
			creatinine  &   INR  & glucose  &   Hb \\
			MCHC  &  MCV  & platelet count  &   MCH \\
			C-reactive protein  &   total white blood cell count  & &\\ 
			\bottomrule
		\end{tabular}
	}
\end{table}
\footnotetext{https://github.com/ratschlab/HIRID-ICU-Benchmark}

\newpage
\begin{table}[H]
	\caption[Caption for 1]{List of medical intervention features for HiRID dataset, \red{where \textbf{Features} indicates the name of the intervention features during model's training, \textbf{Category of treatment} shows the category of treatment that the specific intervention feature belongs to, and \textbf{Source} is the corresponding feature names in the official HiRID preprocessing codes (meta-variables from \texttt{Merging stage} \footnotemark ) \cite{yeche2021hirid} that we extracted based on. The dimension of discrete-valued features for HiRID dataset during model's training is 39.}}
	\vspace{2ex}
	\begin{center}
	\resizebox{.6\textwidth}{!}{
		\begin{tabular}{l|l|l}
			\toprule
			\textbf{\red{Category of treatment}} & \textbf{\red{Features}} & \textbf{Source}\\
			\midrule
			Oxygen therapy &  supplemental oxygen & \texttt{vm23}\\
			&  mechanical ventilation & \texttt{vm60}\\
			\midrule
			Crystalloids & crystalloids & \texttt{vm33}\\
			\midrule
			Colloids & colloids & \texttt{vm34}\\
			\midrule
			Renal therapy & haemofiltration & \texttt{vm72}\\
			\midrule
			Blood transfusion & packed red blood cells & \texttt{pm35} \\
			& FFP & \texttt{pm36} \\
			& platelets & \texttt{pm37} \\
			\midrule
			Vaspressor/inotropes & norepinephrine & \texttt{pm39} \\
			& epinephrine & \texttt{pm40} \\
			& dobutamine & \texttt{pm41} \\
			& milrinone & \texttt{pm42} \\
			& levosimendan & \texttt{pm43} \\
			& theophyllin & \texttt{pm44} \\
			& vasopressin & \texttt{pm45} \\
			& desmopressin & \texttt{pm46} \\
			\midrule
			Vasodilators & vasodilators & \texttt{pm47} \\
			\midrule
			Antihypertensives & ACE inhibitors & \texttt{pm48} \\
			& Calcium channel blockers & \texttt{pm50} \\
			& Beta-blocker & \texttt{pm51} \\
			\midrule
			Antiarrhythmics & adenosine & \texttt{pm53} \\  
			& digoxin & \texttt{pm54} \\ 
			& amiodarone & \texttt{pm55} \\ 
			& atropine & \texttt{pm56} \\
			\midrule
			Antibiotics & antibiotics & \texttt{pm73} \\
			& antimycotic & \texttt{pm74} \\
			& antiviral & \texttt{pm75} \\
			\midrule
			%Sedation & benzodiazepine & \texttt{pm77} \\
			%& alpha-2 agonists & \texttt{pm78} \\
			%& barbiturate & \texttt{pm79} \\
			%& propofol & \texttt{pm80} \\
			%\midrule
			Insulin & insulin & \texttt{pm82, pm83} \\
			\midrule
			Pain killers & opioid & \texttt{pm86} \\
			& non-opioid & \texttt{pm87} \\
			\midrule
			Steroids & steroids & \texttt{pm91} \\
			\midrule
			Anticoagulants & heparin & \texttt{pm95} \\
			\bottomrule
		\end{tabular}
	}
	\end{center}
\end{table}
\footnotetext{https://github.com/ratschlab/HIRID-ICU-Benchmark}

\newpage
\section{Model training.}
\subsection{Implementation details}
\red{During the model training of EHR-M-GAN, the hyperparameters are optimized based on the comparison between the synthetic data and leave-out real data, estimated by mean maximum discrepancy (MMD) for continuous data and mean squared errors (MSEs) over the Bernoulli probability for discrete data, as the scoring functions. Visual inspection is also used during training to intuitively compare the resemblance between synthetic and real data. Table \ref{tb:hyper} shows the hyperparameter values of the network architecture for searching over. The optimal hyperparameters for GANs' training is listed in our GitHub codebase (see \textit{train\_config.py} file). The model which generates the best results is saved and used for the final results.
\begin{table}[H]
        \centering
	\caption[Caption for hi]{\red{List of hyperparameters of EHR-M-GAN.}}
	\vspace{2ex}
	\resizebox{.99\textwidth}{!}{
 {\color{Red}
	\begin{tabular}{p{6cm}| p{8cm}}
	\toprule
	\textbf{Hyperparameters} & \textbf{Searching space} \\
	\midrule
Batch size  &  \{128, 256, 512\} \\
Epochs for pretraining & \{200, 500, 800\} \\
Epochs for training GANs & \{500, 800\} \\
Rounds for jointly training \textit{G}/\textit{D}/\textit{V} & \{3/1/1, 5/1/1\} \\
Learning rate for pretraining & \{0.001, 0.0001, 0.0005\} \\
Learning rate for training GANs & \{0.001, 0.0001, 0.0005\} \\
Depths for encoders and decoders & \{3, 5\} \\
Depths for generators & \{3, 5\} \\
Depths for discriminators & \{1, 3, 5\} \\
Sizes for encoders and decoders & \{64, 128, 256\} \\
Sizes for generators & \{256, 512\} \\
Sizes for discriminators & \{256, 512\} \\
Weight scalar for pretraining  & \{0.01, 0.1, 0.25, 0.5, 1, 2, 5\} \\
Weight scalar for training GANs  & \{0.1, 0.5, 1, 5, 10, 20\} \\
Optimizer & Adam \\
	\bottomrule
	\end{tabular}}
	}
\end{table}\label{tb:hyper}

During the pretraining stage of dual-VAE module, we implemented the VAEs with recurrent neural network based on Google DeepMind's ``DRAW'' --- Deep Recurrent Attentive Writer \cite{gregor2015draw}.
Instead of automatically generating the entire images/timeseries at once, it utilizes a sequential variational auto-encoding framework that enables the iterative generation of multivariate timeseries. The reconstruction loss on the leave-out validation set (i.e., the ``one-to-one'' mapping) is used for optimizing the hyperparameters in dual-VAEs (see Table \ref{tb:hyper}).

Furthermore, to stablize GANs' training and overcome the problem of mode collapse, training strategies such as feature matching loss is utilized \cite{salimans2016improved}. Feature matching is a regularizing objective that prevents the generator in GANs from overtraining on the current discriminator. It has been shown effective to stablize the GANs' training as it calculates the \textit{statistics} of the real data per minibatch, instead of directly maximizing the output of the discriminator. The formal definition of feature matching loss is described as follows:
$$L=\left\|\mathbb{E}_{x \sim p_{\text {data }}} \mathbf{f}(\mathbf{x})-\mathbb{E}_{z \sim p_z(\mathbf{z})} \mathbf{f}(G(\mathbf{z}))\right\|_2^2$$
where $f(x)$ is the feature representation of the intermediate layer of the discriminator (layer before the final classification).}

\subsection{Ablation study for training dual-VAE}
\hl{Multiple losses are placed when optimizing the shared latent space in the dual-VAE module. Except for the standard evidence lower bound (ELBO) loss in VAE, external losses, namely (1) Matching loss; (2) Contrastive loss, and (3) Semantic loss (for the conditional variation of our proposed model) are used. Also, during the implementation, the weight-sharing constraint is adopted for specific layers in dual-VAE's encoder and decoder pairs to extract the high-level representations from mixed-type inputs (see Section S.1.C \mbox{\textit{Shared latent space learning using dual-VAE}} for details). In order to analyze the contribution of each aforementioned component when training dual-VAE, we perform an ablation study by varying the corresponding training configurations (see Table \mbox{\ref{supp_ablation_vae}}) using MIMIC-III dataset as an example. The performance for synthesizing continuous-valued timeseries is evaluated by maximum mean discrepancy (MMD) and discriminative score. For discrete-valued timeseries, the performance of GANs is evaluated by dimensional-wise probability (DWP) quantified by the averaged root mean squared errors (RMSEs) across all feature dimensions (see \mbox{\textit{Dimension-wise probability}} section in the main text for details) and discriminative score. The results of the ablation study are shown in Table \mbox{\ref{supp_ablation_vae}}.
}
%% Also, to incentivize dual-VAE to better bridge the gap between domains of continuous-valued and discrete-valued timeseries, we enforce a  [65, 66] within specific layers of both the encoders pairs (EncC, EncD), and the decoders pairs (DecC, DecD). The weight-sharing constraint can extract and broadcast high-level concepts of the input features across domains of mixed-type data.
\begin{table}[H]
	\caption{The ablation study for components in Dual-VAE on MIMIC-III dataset. `Baseline' represents the proposed GAN models (EHR-M-GAN or EHR-M-GAN$\texttt{cond}$) with all components included. The quality of synthetic continuous-valued timeseries is evaluated by MMD and discriminative score (both the lower the better). The quality of synthetic discrete-valued timeseries is evaluated by averaged RMSEs in DWP and discriminative  score (both the lower the better).}
	\label{supp_ablation_vae}
	%\vspace{2ex}
	\begin{center}
	\resizebox{1.\textwidth}{!}{
        \begin{tabular}{l l cccc}
        \toprule
         &  & \multicolumn{2}{c}{Continuous-valued data} & \multicolumn{2}{c}{Discrete-valued data} \\ 
         \cmidrule(lr){3-4} \cmidrule(lr){5-6} Model & Training configuration & \multicolumn{1}{c}{MMD} & \multicolumn{1}{c}{Discriminative score} & \multicolumn{1}{c}{DWP (RMSEs)} & \multicolumn{1}{c}{Discriminative score} \\ \midrule
            EHR-M-GAN & Baseline & 
            \textBF{0.692 $\pm$ 0.034} & 
            \textBF{0.746 $\pm$ 0.018} & 
            \textBF{0.0104 $\pm$ 0.0006} & %%%
            \textBF{0.813 $\pm$ 0.026} \\
             & w/o Matching loss & 
             0.722 $\pm$ 0.023 & 
             0.758 $\pm$ 0.015 & 
             0.0112 $\pm$ 0.0010 &   %%%
             0.827 $\pm$ 0.019 \\
             & w/o Contrastive loss & 
             0.719 $\pm$ 0.017 &
             0.762 $\pm$ 0.012 &
             0.0109 $\pm$ 0.0009 &   %%%
             0.830 $\pm$ 0.023 \\
             & w/o Shared weights & 
             0.704 $\pm$ 0.031 & 
             0.749 $\pm$ 0.019 & 
             0.0107 $\pm$ 0.0008 & %%%
             0.816 $\pm$ 0.035 \\ 
         \midrule
            EHR-M-GAN$\texttt{cond}$ & Baseline & 
            \textBF{0.604 $\pm$ 0.027} & 
            \textBF{0.729 $\pm$ 0.025} & 
            \textBF{0.0093 $\pm$ 0.0005} & 
            \textBF{0.784 $\pm$ 0.024} \\
             & w/o Matching loss & 
             0.634 $\pm$ 0.026 & 
             0.736 $\pm$ 0.017 &
             0.0106 $\pm$ 0.0013 &
             0.795 $\pm$ 0.022 \\
             & w/o Contrastive loss &
             0.629 $\pm$ 0.022 & 
             0.739 $\pm$ 0.020 & 
             0.0108 $\pm$ 0.0007 & 
             0.796 $\pm$ 0.028 \\
             & w/o Semantic loss & 
             0.647 $\pm$ 0.034 & 
             0.743 $\pm$ 0.011 & 
             0.0114 $\pm$ 0.0004 & 
             0.798 $\pm$ 0.030 \\
             & w/o Shared weights & 
             0.609 $\pm$ 0.035 & 
             0.732 $\pm$ 0.014 & 
             0.0097 $\pm$ 0.0012 &
             0.786 $\pm$ 0.027  \\ 
         \bottomrule
        \end{tabular}
	}
	\end{center}

\end{table}

%% result
As shown in Table \mbox{\ref{supp_ablation_vae}}, both matching loss and contrastive loss contribute to the improvement of EHR-M-GAN's performance when generating mixed-type timeseries data. For example, the absence of the contrastive loss leads to a noticeable degradation in the quality of the synthetic discrete-valued timeseries (evaluated by discriminative score). Also, removing the matching loss causes the increase of the MMD between real and synthetic continuous-valued timeseries. The weight-sharing scheme between the encoder and decoder architectures in the dual-VAE also boosts GANs' performance but within a limited range. For EHR-M-GAN$\texttt{cond}$ model, the effectiveness of the components that appear in EHR-M-GAN can still be observed. On the other hand, semantic loss, which injects conditional information into the networks, plays a major role in synthesizing more realistic patient trajectories. The results in Table \mbox{\ref{supp_ablation_vae}} show that the impact of the semantic loss exceeds the other two losses in learning the valid shared latent representations in dual-VAE.

\newpage
\section{Results.}
\subsection{Dimension-wise probability.}

\begin{figure*}[h]
    \captionsetup{skip=-5pt}
			\centering
			\begin{subfigure}[b]{0.175\textwidth}
				\centering
				\includegraphics[width=1.1\linewidth]{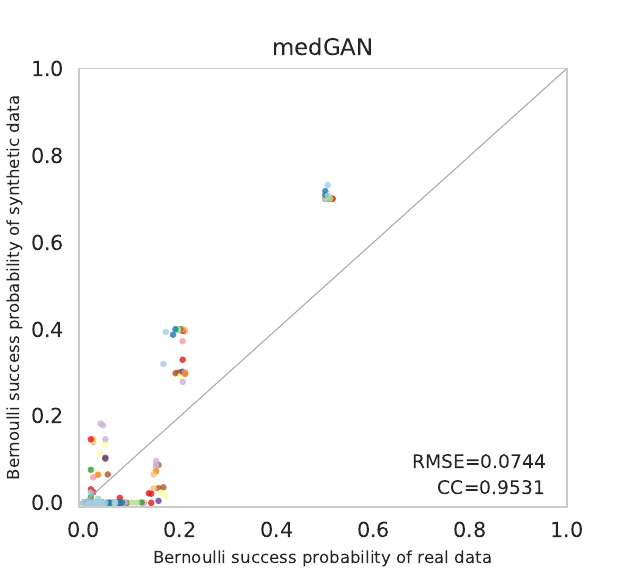} 
				\captionsetup{justification=centering}
				{{\small }}    
			\end{subfigure}
			\quad
			\begin{subfigure}[b]{0.175\textwidth}  
				\centering 
				\includegraphics[width=1.1\linewidth]{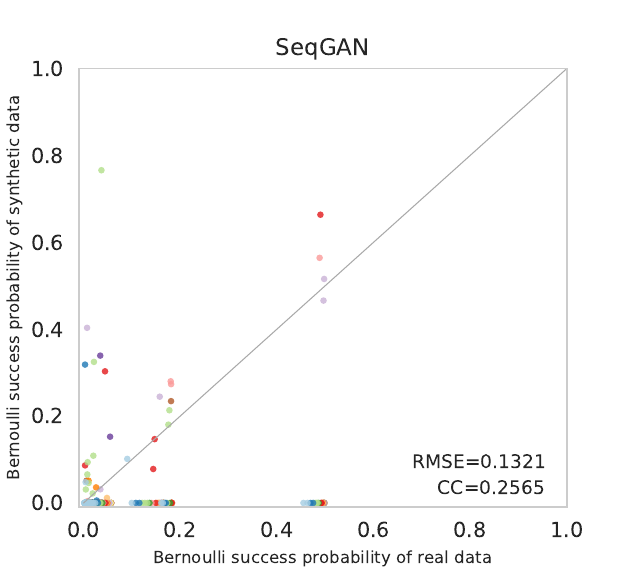} 
				\captionsetup{justification=centering}
				{{\small }} 
			\end{subfigure}
			\quad
			\begin{subfigure}[b]{0.175\textwidth}  
				\centering 
				\includegraphics[width=1.1\linewidth]{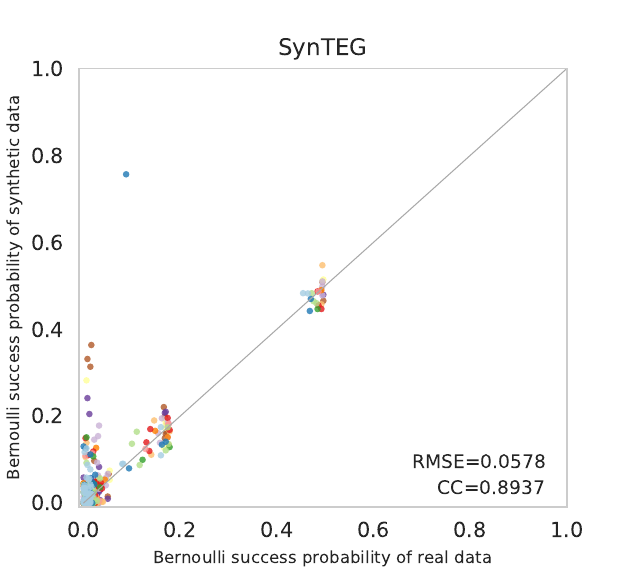} 
				\captionsetup{justification=centering}
				{{\small }} 
			\end{subfigure}
			\quad
			\begin{subfigure}[b]{0.175\textwidth}  
				\centering 
				\includegraphics[width=1.1\linewidth]{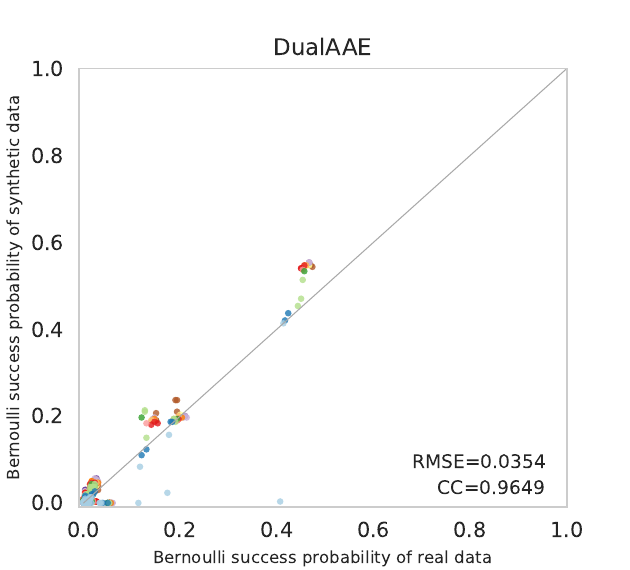}  
				\captionsetup{justification=centering}
				{{\small }} 
			\end{subfigure}
			\quad
			\begin{subfigure}[b]{0.175\textwidth}  
				\centering 
				\includegraphics[width=1.1\linewidth]{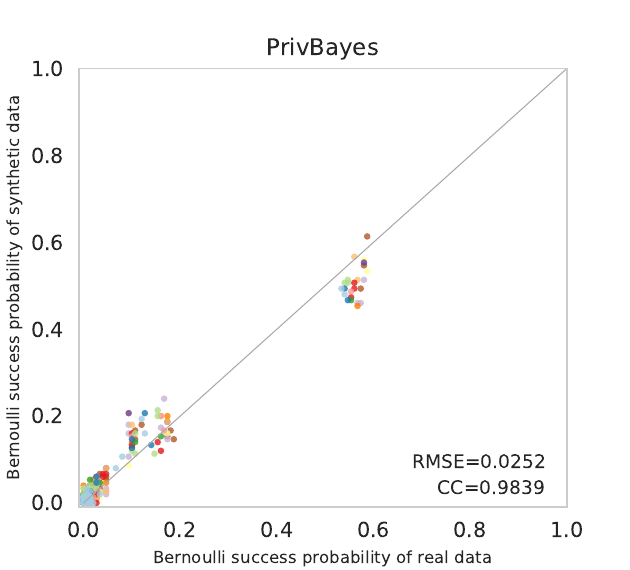}
				\captionsetup{justification=centering}
				{{\small }} 
			\end{subfigure}
			%%%%%%%%%%%%%%%%%
			\medskip
			\begin{subfigure}[b]{0.175\textwidth}
				\centering
				\includegraphics[width=1.1\linewidth]{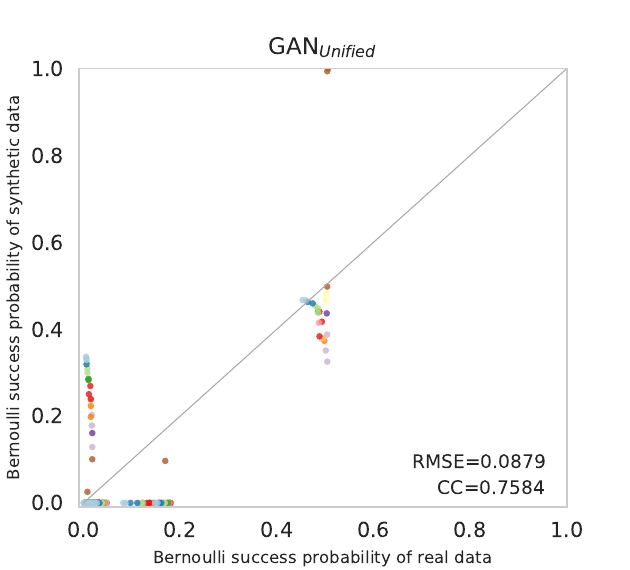}
				\captionsetup{justification=centering}
				{{\small }} 
			\end{subfigure}
			\quad
			\begin{subfigure}[b]{0.175\textwidth}
				\centering
				\includegraphics[width=1.1\linewidth]{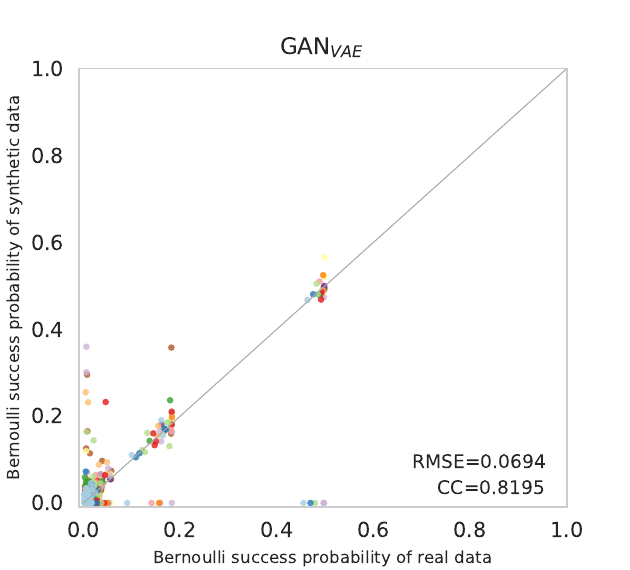}
				\captionsetup{justification=centering}
				{{\small }} 
			\end{subfigure}
			\quad
			\begin{subfigure}[b]{0.175\textwidth}  
				\centering 
				\includegraphics[width=1.1\linewidth]{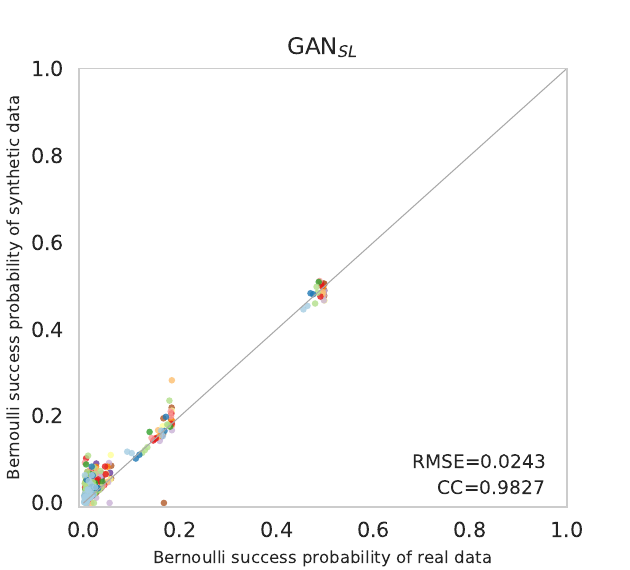}  
				\captionsetup{justification=centering}
				{{\small }} 
			\end{subfigure}
			\quad
			\begin{subfigure}[b]{0.175\textwidth}  
				\centering 
				\includegraphics[width=1.1\linewidth]{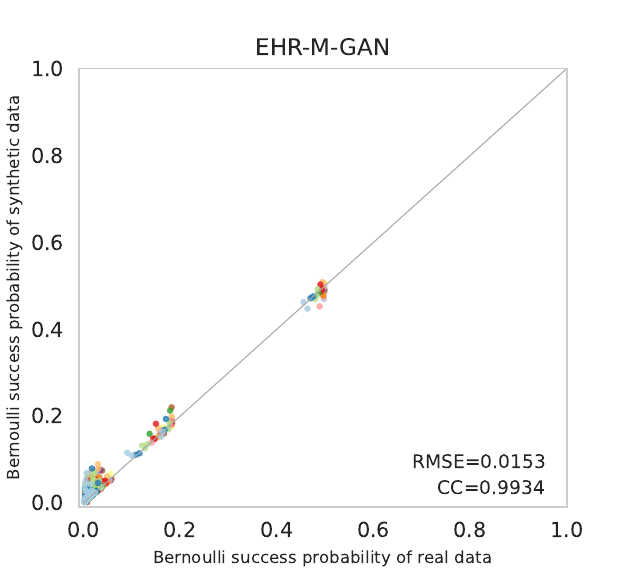}  
				\captionsetup{justification=centering}
				{{\small }} 
			\end{subfigure}
			\quad
			\begin{subfigure}[b]{0.175\textwidth}  
				\centering 
				\includegraphics[width=1.1\linewidth]{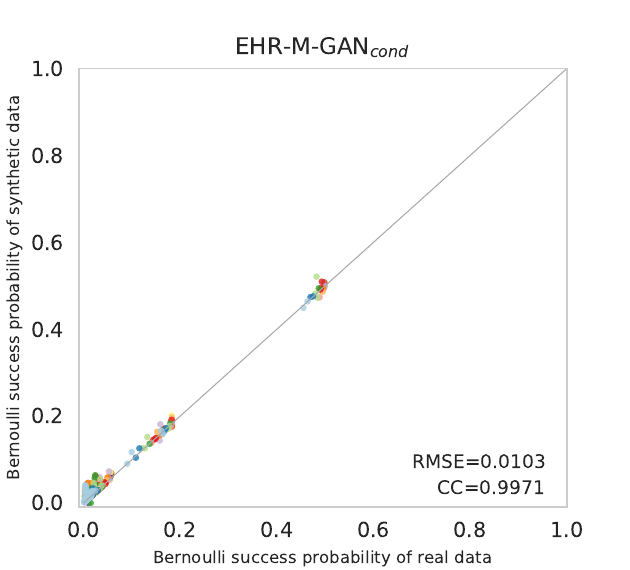}   
				\captionsetup{justification=centering}
				{{\small }} 
			\end{subfigure}	
			
			\caption{\textbf{Scatterplot of the dimension-wise probability test on eICU dataset.} The x-axis and y-axis represents the probability distribution for the real data and synthetic data with same sample size, respectively. The optimal performance appears along the diagonal line.
            Each dot represents a treatment status at a particular time in the patient EHR data. The optimal performance appears along the diagonal line.
			The corresponding CCs ($[0, 1]$, the higher the better) and RMSEs ($\left[ 0, +\infty \right) $, the lower the better) are also calculated to quantify the probability distribution similarities between the real and synthetic EHRs timeseries.}
\end{figure*}

\begin{figure*}[h]
    \captionsetup{skip=-5pt}
			\centering
			\begin{subfigure}[b]{0.175\textwidth}
				\centering
				\includegraphics[width=1.1\linewidth]{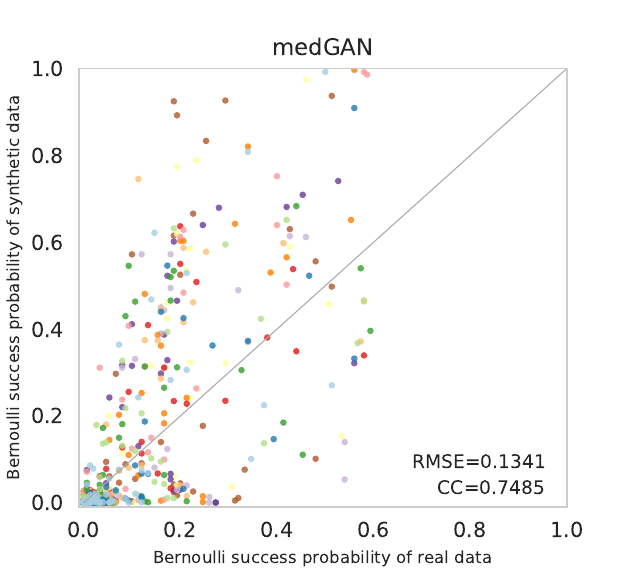} 
				\captionsetup{justification=centering}
				{{\small }} 
			\end{subfigure}
			\quad
			\begin{subfigure}[b]{0.175\textwidth}  
				\centering 
				\includegraphics[width=1.1\linewidth]{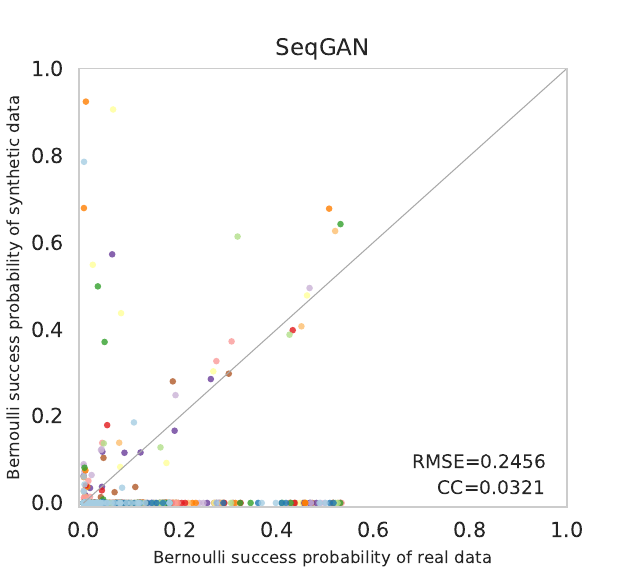} 
				\captionsetup{justification=centering}
				{{\small }} 
			\end{subfigure}
			\quad
			\begin{subfigure}[b]{0.175\textwidth}  
				\centering 
				\includegraphics[width=1.1\linewidth]{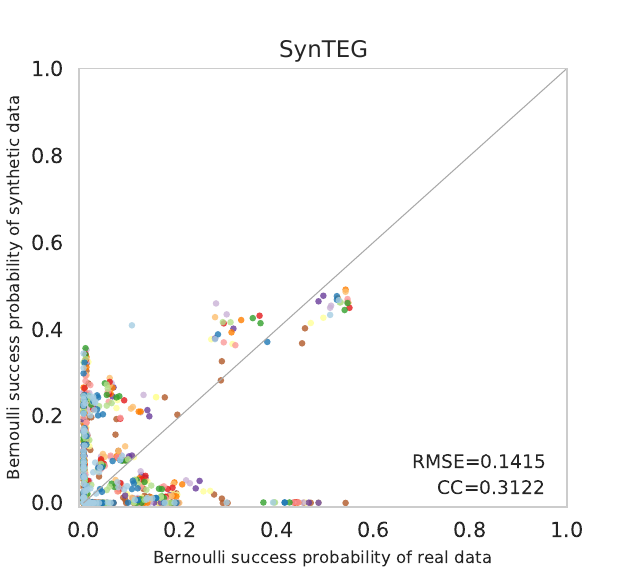} 
				\captionsetup{justification=centering}
				{{\small }} 
			\end{subfigure}
			\quad
			\begin{subfigure}[b]{0.175\textwidth}  
				\centering 
				\includegraphics[width=1.1\linewidth]{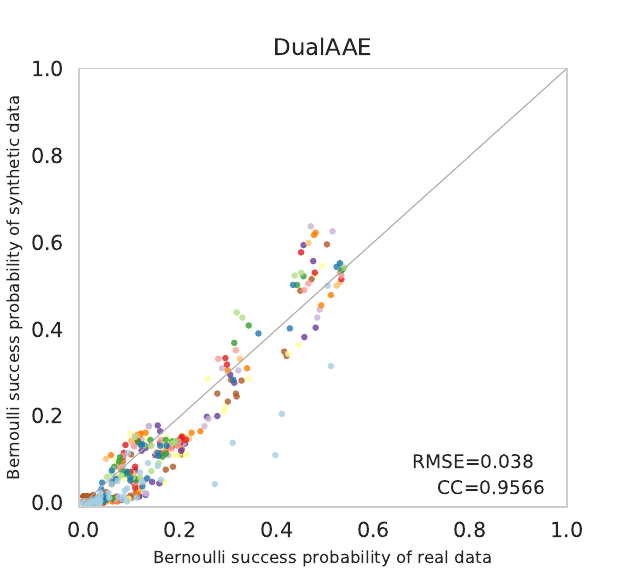}  
				\captionsetup{justification=centering}
				{{\small }} 
			\end{subfigure}
			\quad
			\begin{subfigure}[b]{0.175\textwidth}  
				\centering 
				\includegraphics[width=1.1\linewidth]{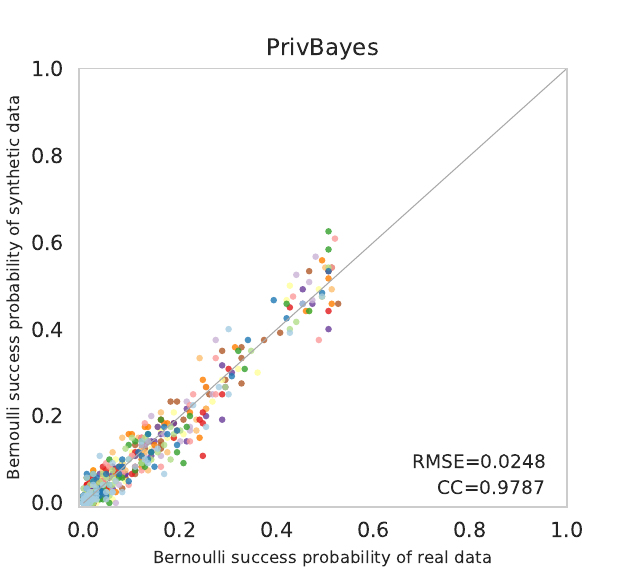}
				\captionsetup{justification=centering}
				{{\small }} 
			\end{subfigure}
			%%%%%%%%%%%%%%%%%
			\medskip
			\begin{subfigure}[b]{0.175\textwidth}
				\centering
				\includegraphics[width=1.1\linewidth]{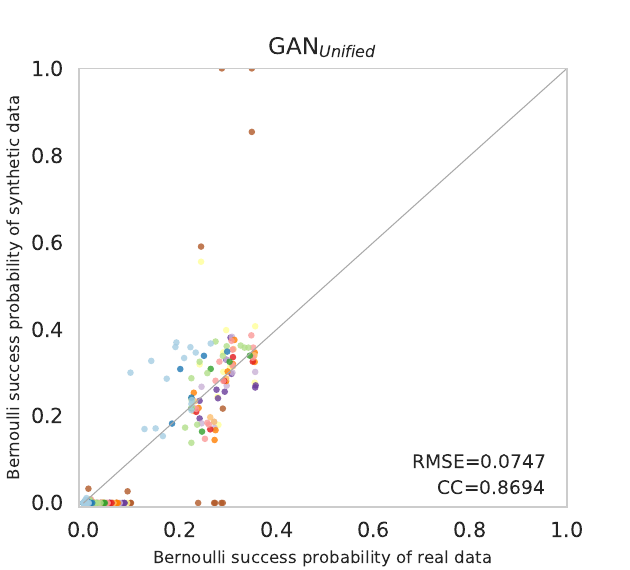}
				\captionsetup{justification=centering}
				{{\small }} 
			\end{subfigure}
			\quad
			\begin{subfigure}[b]{0.175\textwidth}
				\centering
				\includegraphics[width=1.1\linewidth]{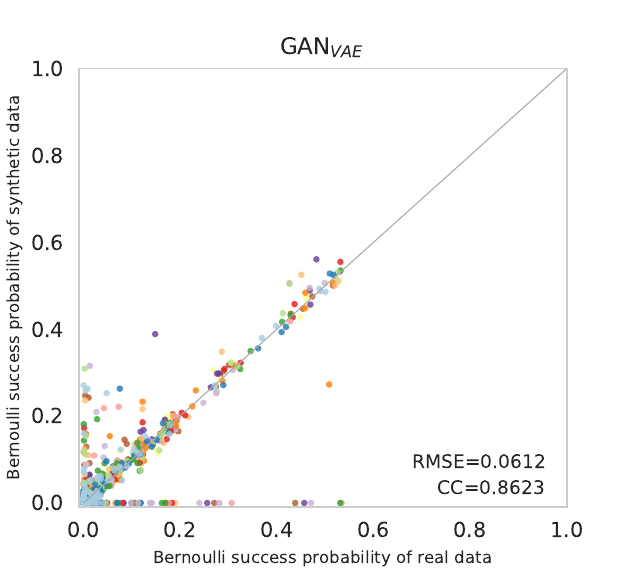}
				\captionsetup{justification=centering}
				{{\small }} 
			\end{subfigure}
			\quad
			\begin{subfigure}[b]{0.175\textwidth}  
				\centering 
				\includegraphics[width=1.1\linewidth]{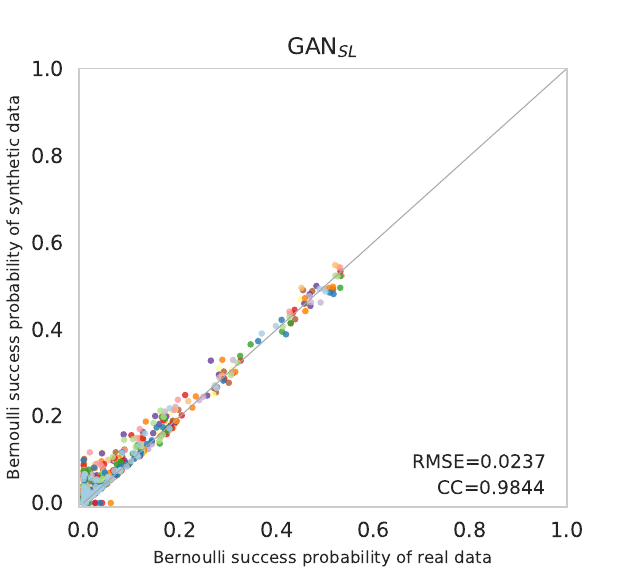}  
				\captionsetup{justification=centering}
				{{\small }} 
			\end{subfigure}
			\quad
			\begin{subfigure}[b]{0.175\textwidth}  
				\centering 
				\includegraphics[width=1.1\linewidth]{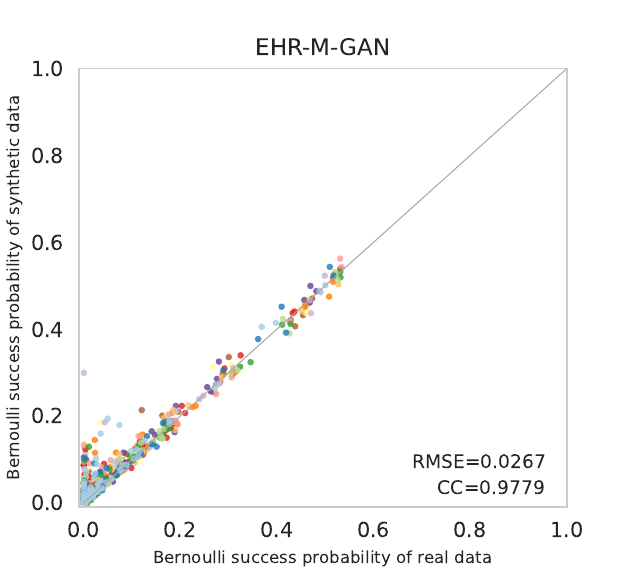}  
				\captionsetup{justification=centering}
				{{\small }} 
			\end{subfigure}
			\quad
			\begin{subfigure}[b]{0.175\textwidth}  
				\centering 
				\includegraphics[width=1.1\linewidth]{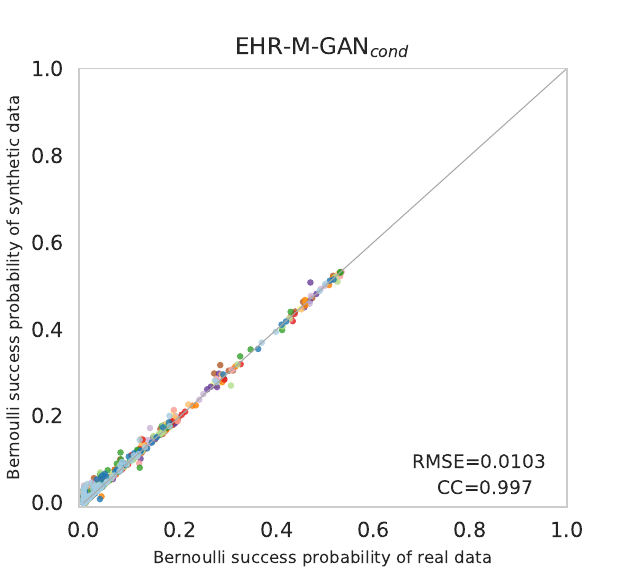}  
				\captionsetup{justification=centering}
				{{\small }}  
			\end{subfigure}	
			
			\caption{\textbf{Scatterplot of the dimension-wise probability test on eICU dataset.} The x-axis and y-axis represents the probability distribution for the real data and synthetic data with same sample size, respectively. The optimal performance appears along the diagonal line.
            Each dot represents a treatment status at a particular time in the patient EHR data. The optimal performance appears along the diagonal line.
			The corresponding CCs ($[0, 1]$, the higher the better) and RMSEs ($\left[ 0, +\infty \right) $, the lower the better) are also calculated to quantify the probability distribution similarities between the real and synthetic EHRs timeseries.}
\end{figure*}

\newpage
\subsection{Temporal characteristics.}
\begin{figure*}[h!]
    \captionsetup{skip=-5pt}
	\centering
	\begin{subfigure}[b]{0.22\textwidth}
		\centering
		\includegraphics[width=1.1\linewidth]{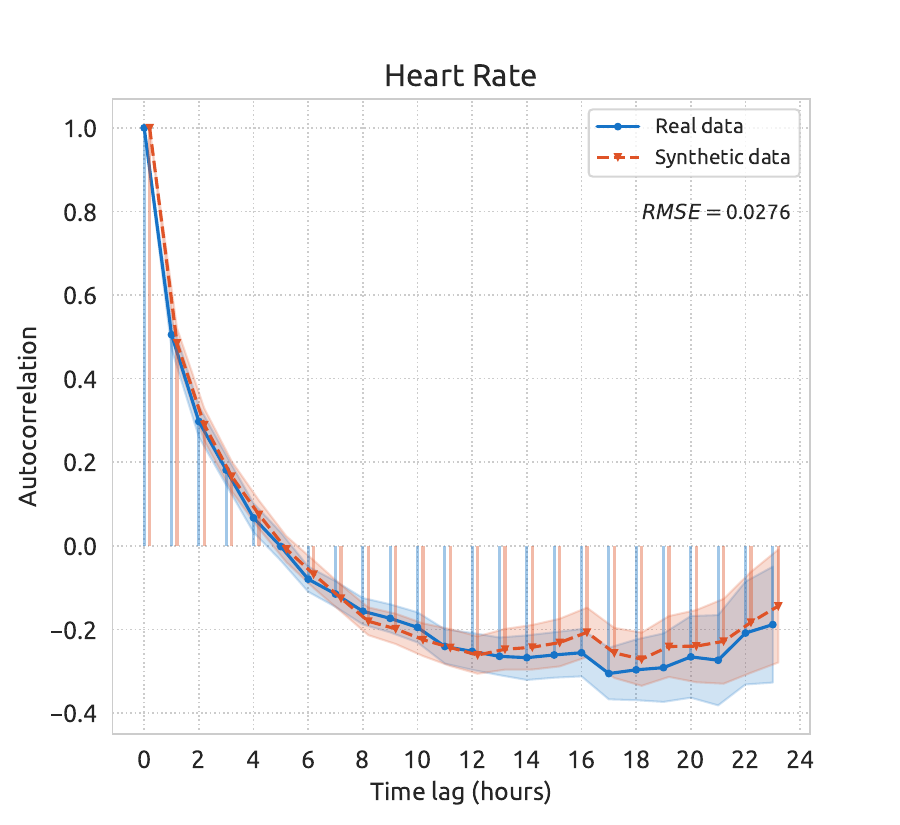}
		\captionsetup{justification=centering}
% 		\caption[ ]%
		{{\small }}    
	\end{subfigure}
\quad
	\begin{subfigure}[b]{0.22\textwidth}  
		\centering 
		\includegraphics[width=1.1\linewidth]{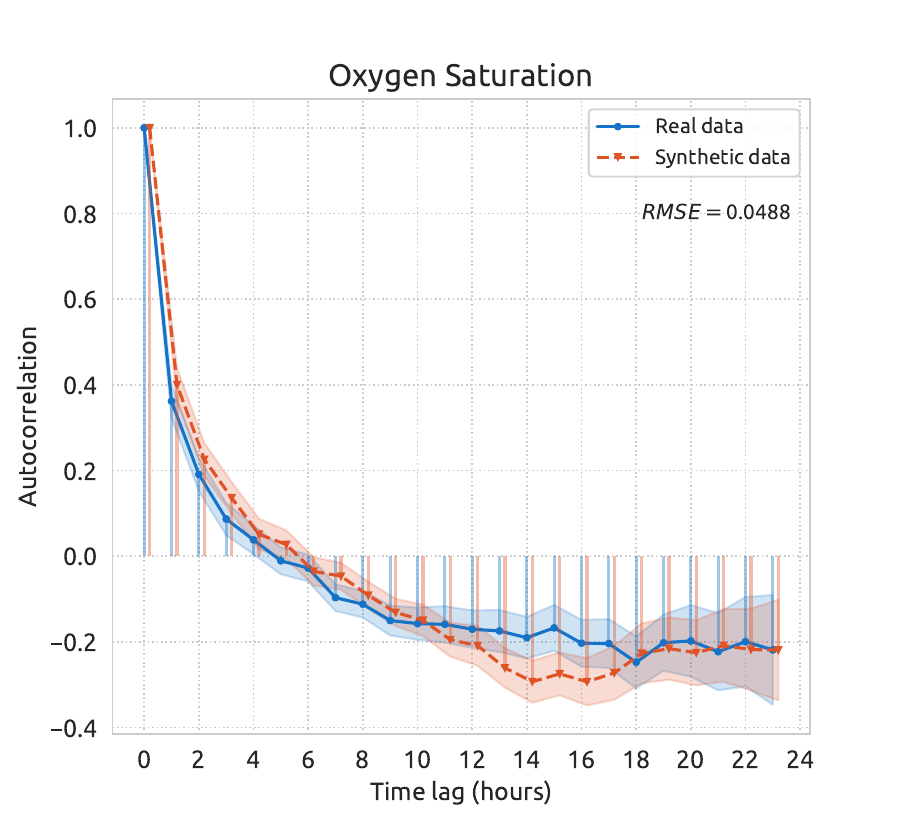}
		\captionsetup{justification=centering}
% 		\caption[ ]%
		{{\small  }}    
	\end{subfigure}
\quad
	\begin{subfigure}[b]{0.22\textwidth}  
		\centering 
		\includegraphics[width=1.1\linewidth]{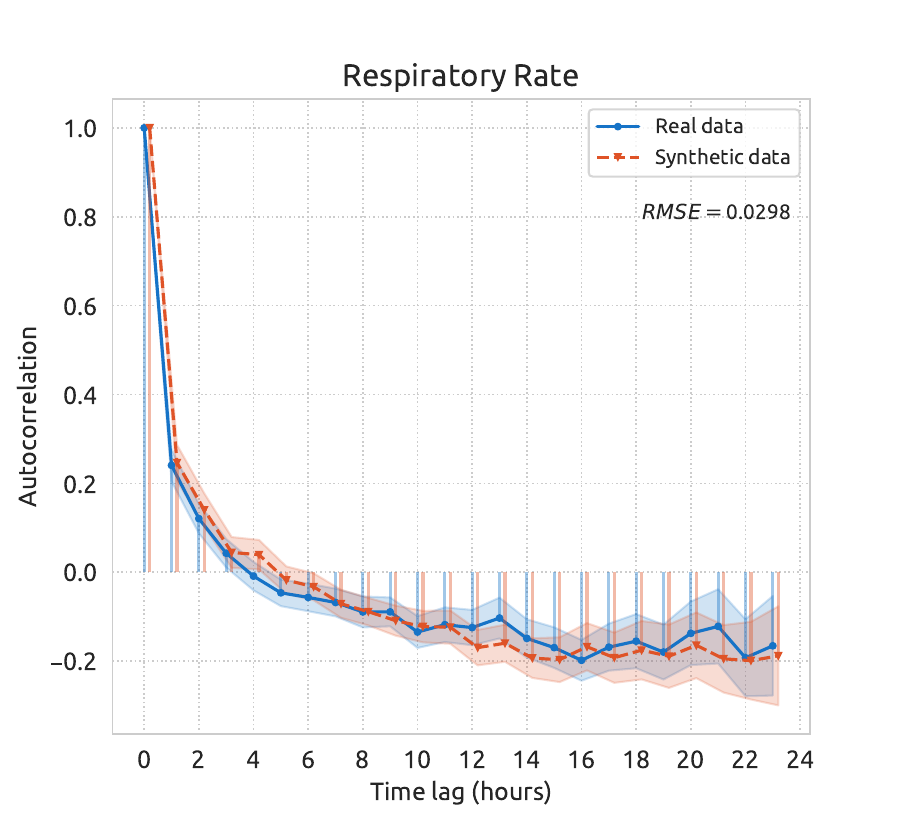}
		\captionsetup{justification=centering}
% 		\caption[ ]%
		{{\small }}    
	\end{subfigure}
\quad
	\begin{subfigure}[b]{0.22\textwidth}  
		\centering 
		\includegraphics[width=1.1\linewidth]{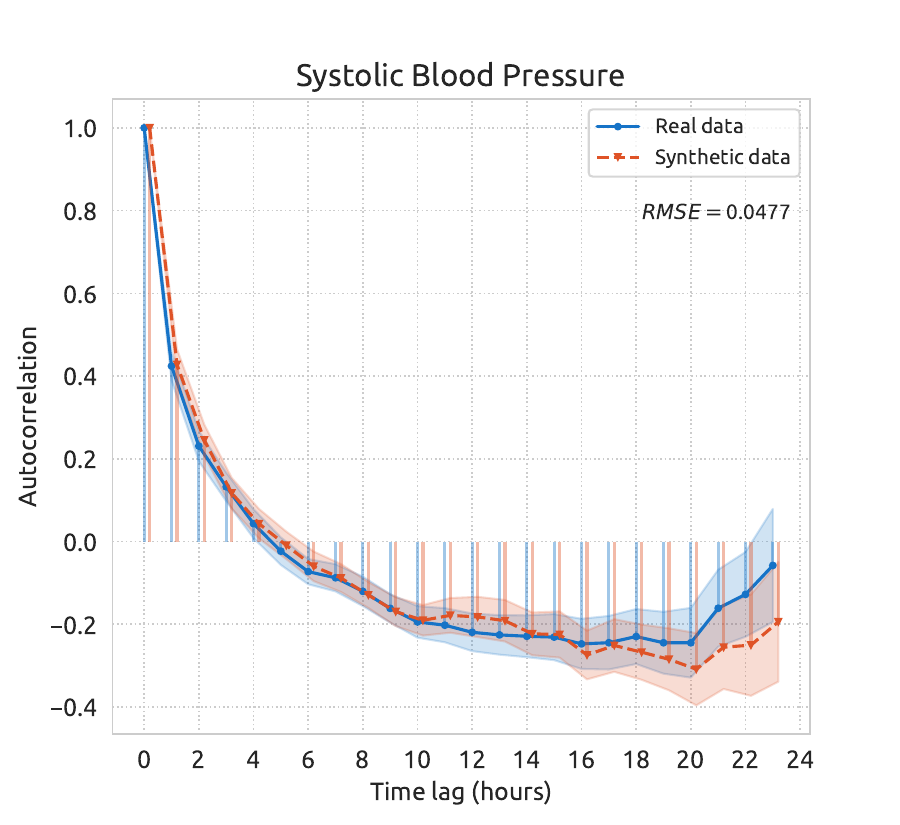}
		\captionsetup{justification=centering}
% 		\caption[ ]%
		{{\small }}    
	\end{subfigure}
	\medskip
	\begin{subfigure}[b]{0.22\textwidth}
		\centering
		\includegraphics[width=1.1\linewidth]{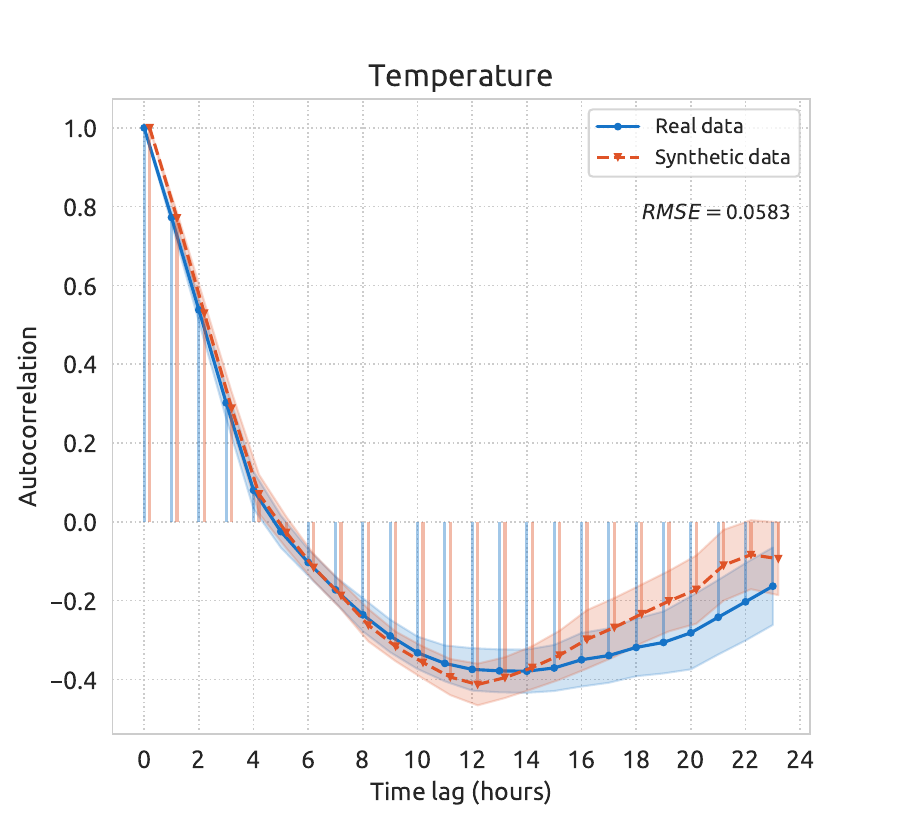}
		\captionsetup{justification=centering}
% 		\caption[ ]%
		{{\small }}    
	\end{subfigure}
\quad
	\begin{subfigure}[b]{0.22\textwidth}  
		\centering 
		\includegraphics[width=1.1\linewidth]{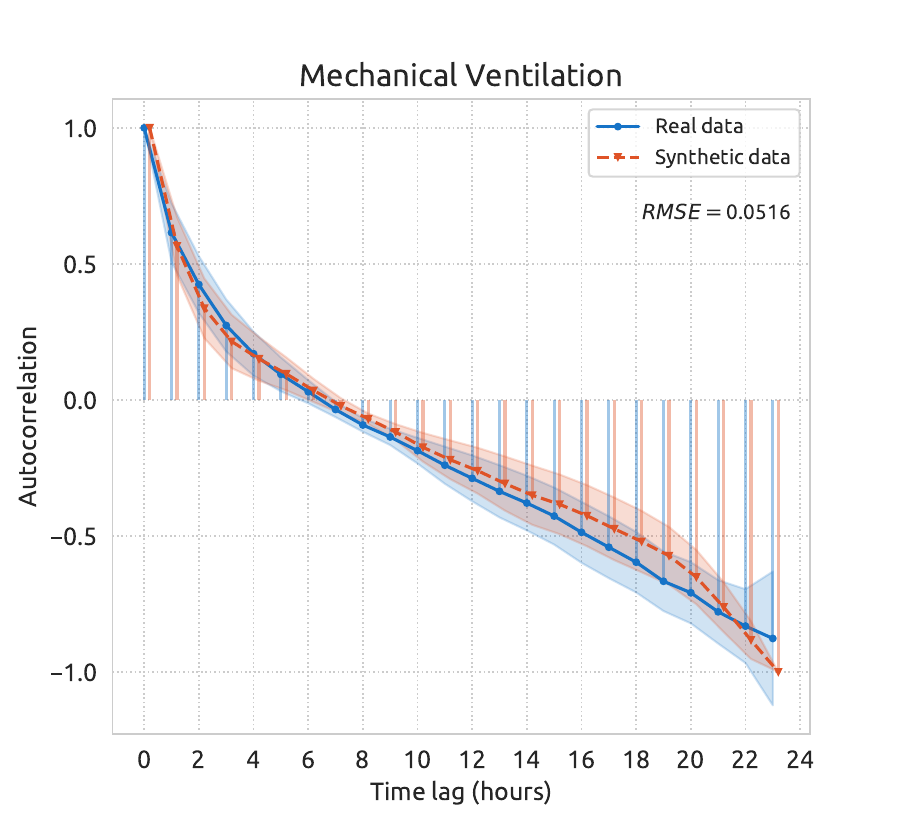}
		\captionsetup{justification=centering}
% 		\caption[ ]%
		{{\small  }}    
	\end{subfigure}
\quad
	\begin{subfigure}[b]{0.22\textwidth}  
		\centering 
		\includegraphics[width=1.1\linewidth]{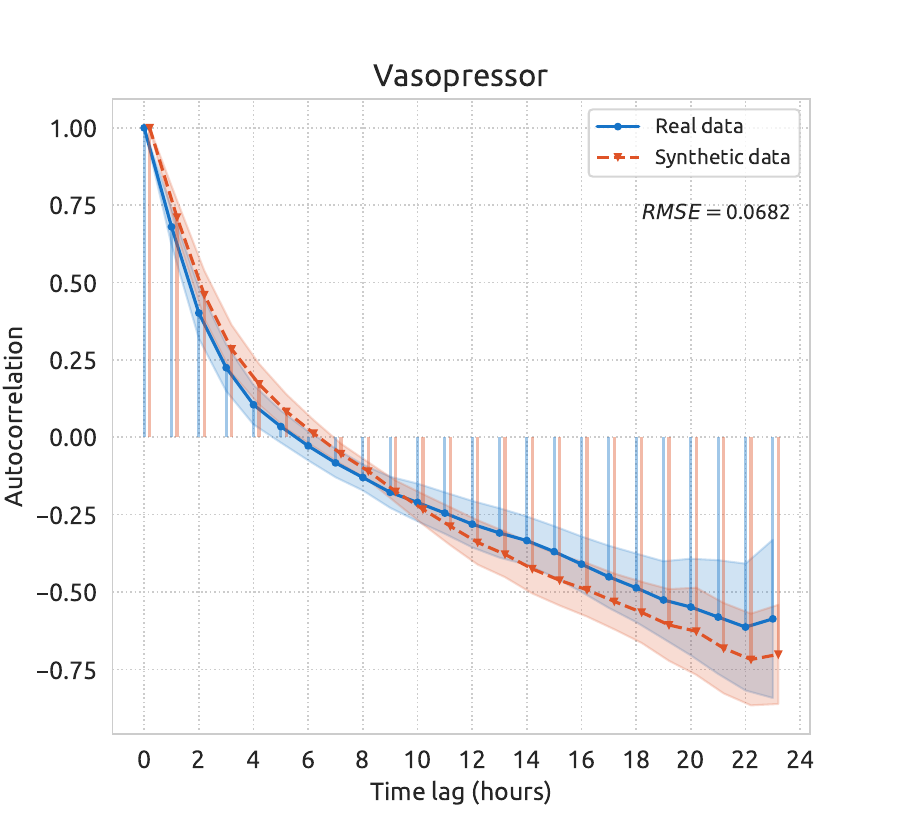}
		\captionsetup{justification=centering}
% 		\caption[ ]%
		{{\small }}    
	\end{subfigure}	
	\caption[ ]{\textbf{Autocorrelation function (ACF) of real data and synthetic data generated by EHR-M-GAN on MIMIC-III dataset.} } 
	\label{fig_acf_mimic}
\end{figure*}

\hl{Fig. \mbox{\ref{fig_acf_mimic}}-\mbox{\ref{fig_acf_hirid}} show the autocorrelation function (ACF) of real timeseries and synthetic timeseries generated by EHR-M-GAN on continuous-valued features (including \mbox{\textit{Heart Rate}}, \mbox{\textit{Oxygen Saturation}}, \mbox{\textit{Respiratory Rate}}, \mbox{\textit{Systolic Blood Pressure}}, and \mbox{\textit{Temperature}}) and discrete-valued features (including \textit{Vasopressor} and \mbox{\textit{Mechanical Ventilation}}). 
The averaged ACF is calculated over the population sampled randomly from both real and synthetic patient data. The averaged autocorrelation for real patient trajectories (solid blue line) and synthetic patient trajectories (red dashed line) are calculated, with the light colored regions indicating the corresponding 95\% confidence interval. The root-mean-square errors (RMSEs) are also calculated for the two curves on each variable to quantitatively evaluate the temporal characteristics captured by the synthetic data.}

\newpage

\begin{figure*}[h!]
    \captionsetup{skip=-5pt}
	\centering
	\begin{subfigure}[b]{0.22\textwidth}
		\centering
		\includegraphics[width=1.1\linewidth]{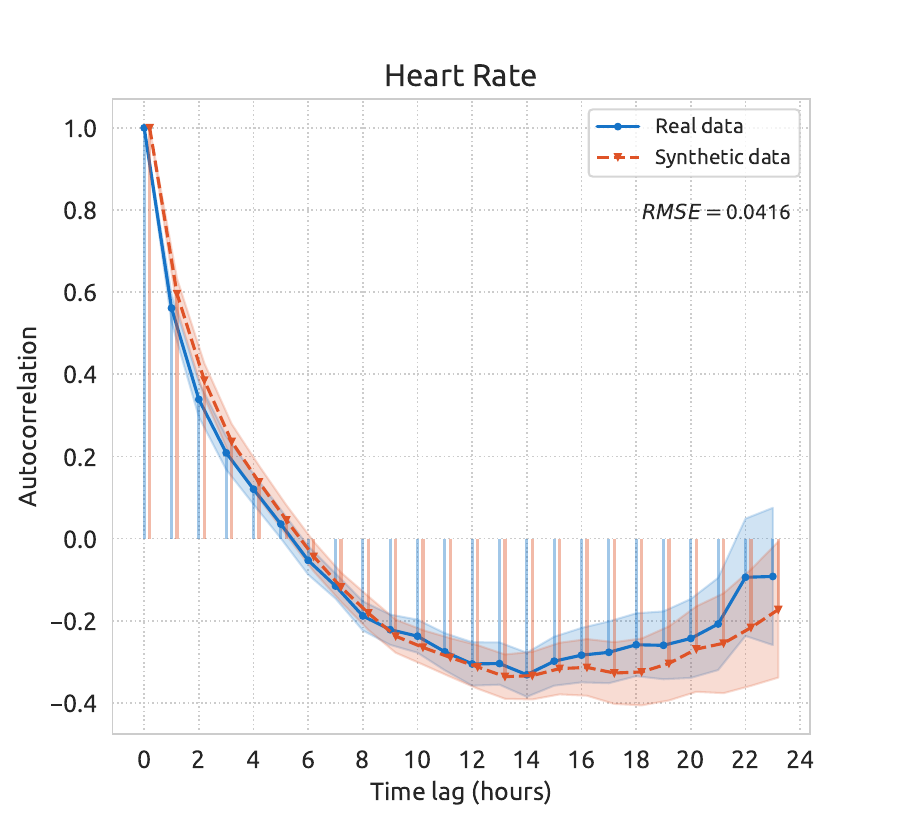}
		\captionsetup{justification=centering}
% 		\caption[ ]%
		{{\small }}    
	\end{subfigure}
\quad
	\begin{subfigure}[b]{0.22\textwidth}  
		\centering 
		\includegraphics[width=1.1\linewidth]{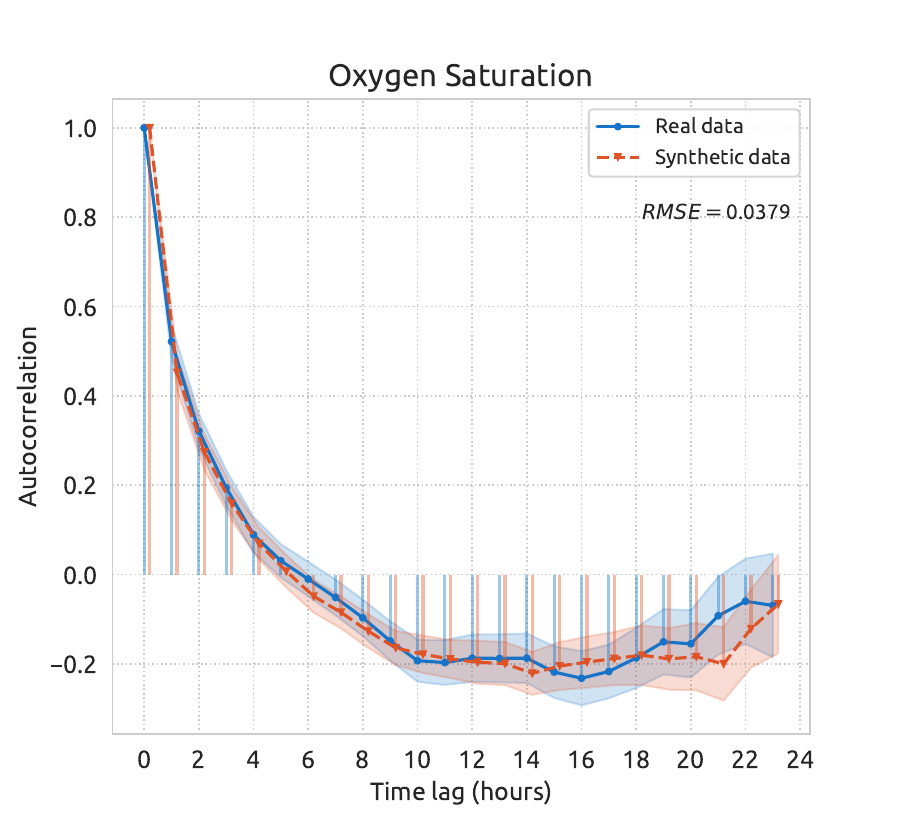}
		\captionsetup{justification=centering}
% 		\caption[ ]%
		{{\small  }}    
	\end{subfigure}
\quad
	\begin{subfigure}[b]{0.22\textwidth}  
		\centering 
		\includegraphics[width=1.1\linewidth]{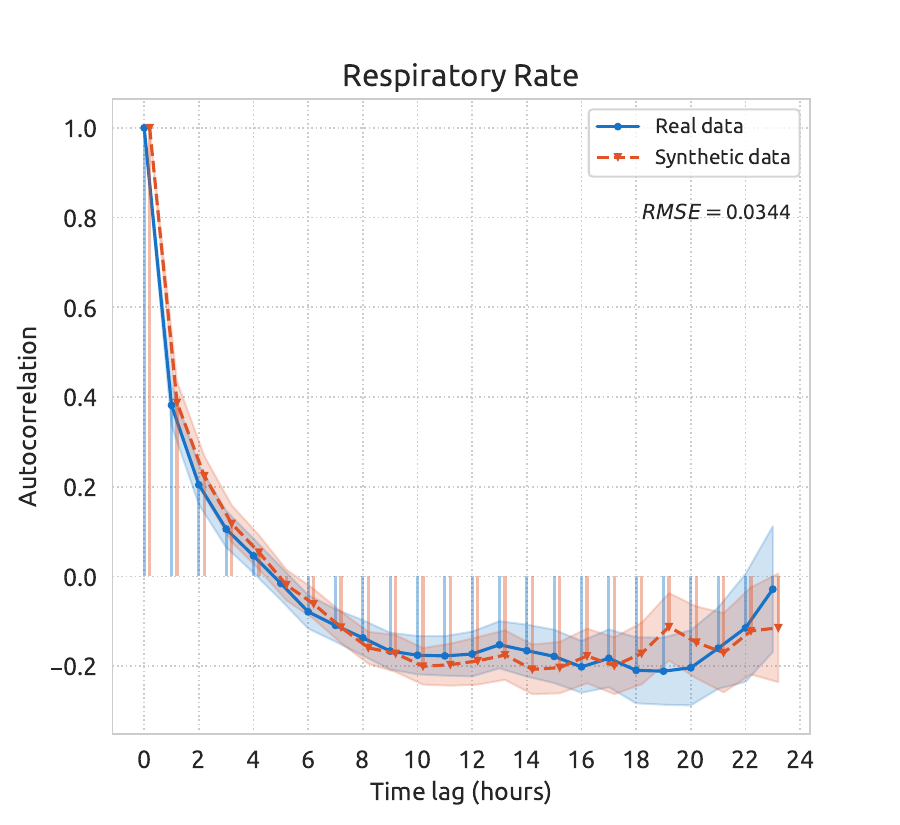}
		\captionsetup{justification=centering}
% 		\caption[ ]%
		{{\small }}    
	\end{subfigure}
\quad
	\begin{subfigure}[b]{0.22\textwidth}  
		\centering 
		\includegraphics[width=1.1\linewidth]{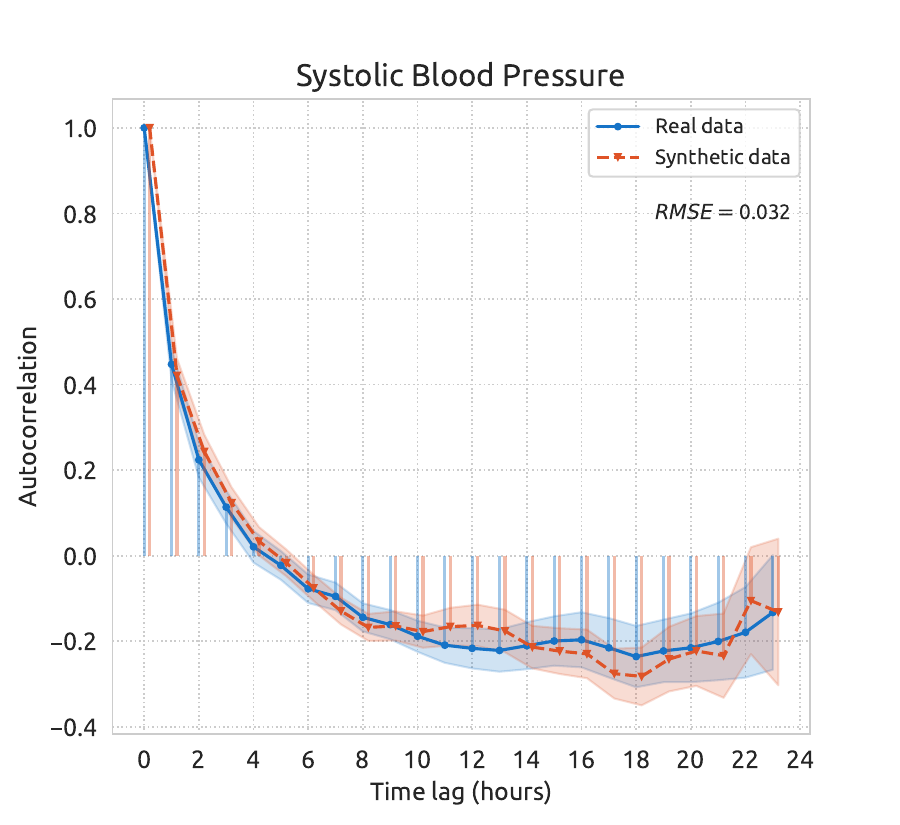}
		\captionsetup{justification=centering}
% 		\caption[ ]%
		{{\small }}    
	\end{subfigure}
	\medskip
	\begin{subfigure}[b]{0.22\textwidth}
		\centering
		\includegraphics[width=1.1\linewidth]{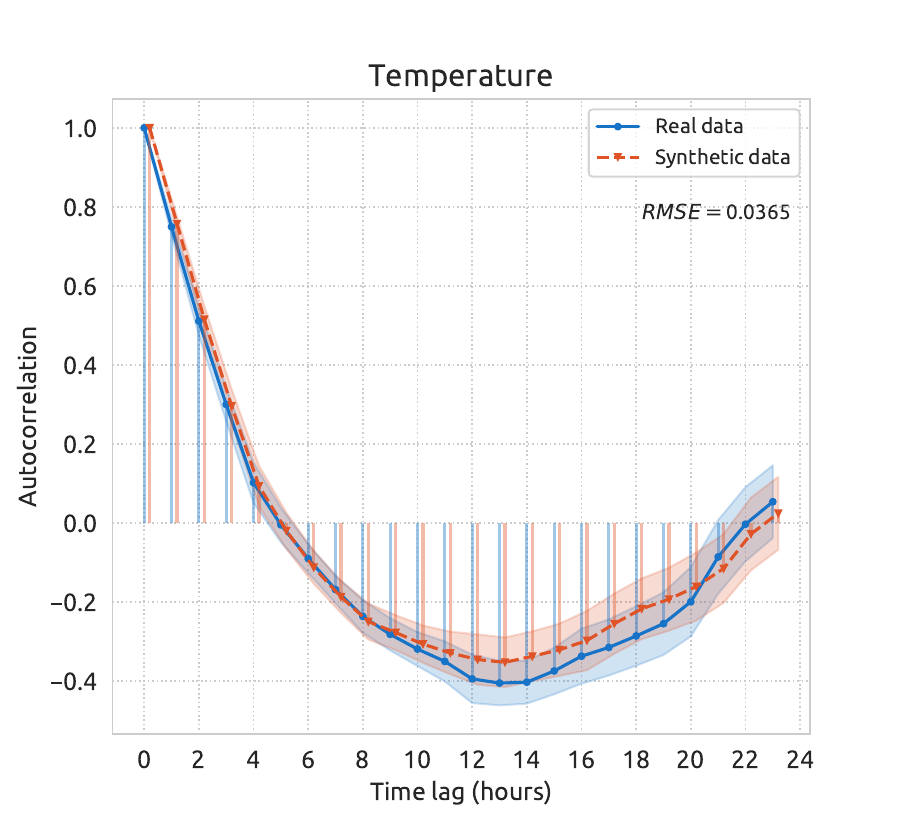}
		\captionsetup{justification=centering}
% 		\caption[ ]%
		{{\small }}    
	\end{subfigure}
\quad
	\begin{subfigure}[b]{0.22\textwidth}  
		\centering 
		\includegraphics[width=1.1\linewidth]{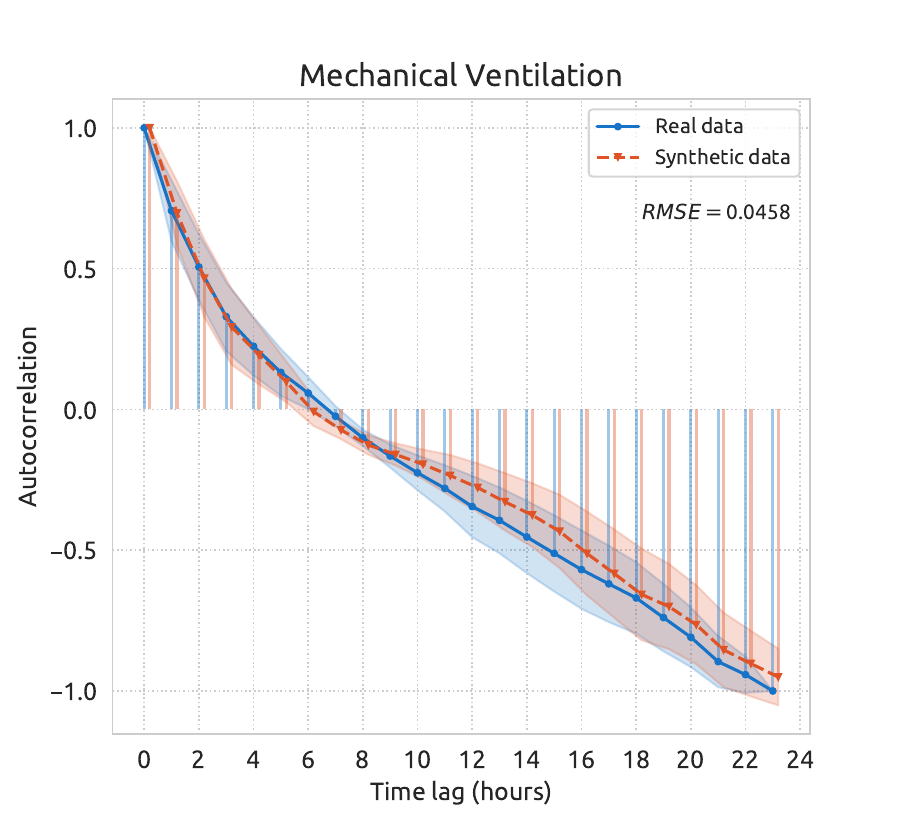}
		\captionsetup{justification=centering}
% 		\caption[ ]%
		{{\small  }}    
	\end{subfigure}
\quad
	\begin{subfigure}[b]{0.22\textwidth}  
		\centering 
		\includegraphics[width=1.1\linewidth]{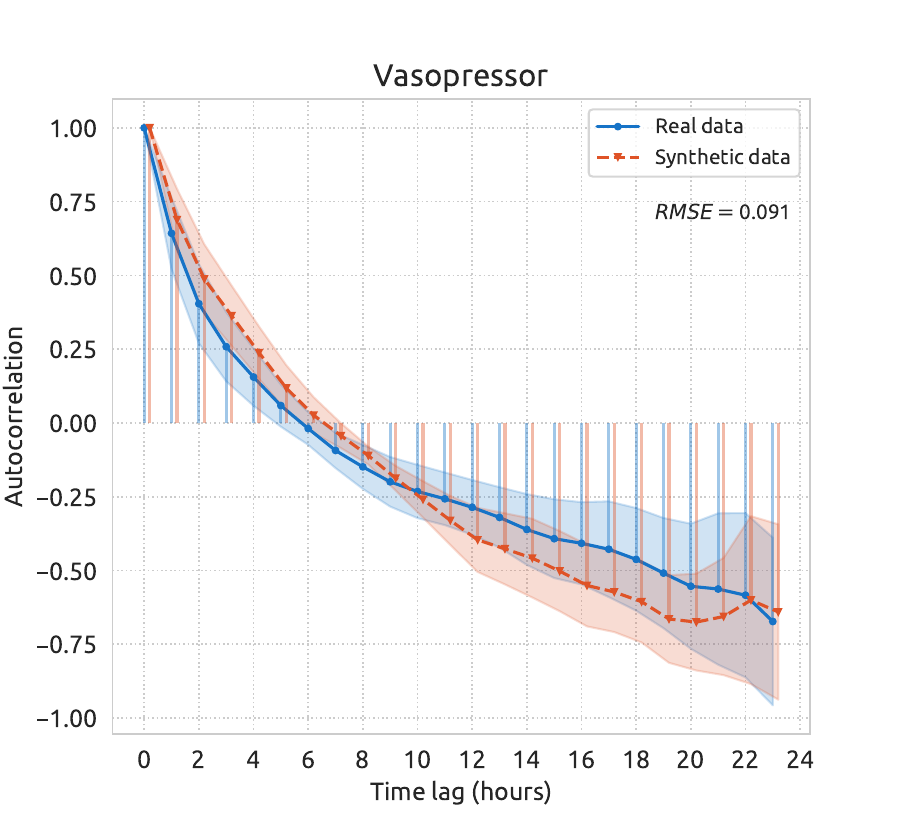}
		\captionsetup{justification=centering}
% 		\caption[ ]%
		{{\small }}    
	\end{subfigure}	
	\caption[ ]{\textbf{Autocorrelation function (ACF) of real data and synthetic data generated by EHR-M-GAN on eICU dataset.}  } 
	\label{fig_acf_eicu}
\end{figure*}

\newpage
\begin{figure*}[h!]
    \captionsetup{skip=-5pt}
	\centering
	\begin{subfigure}[b]{0.22\textwidth}
		\centering
		\includegraphics[width=1.1\linewidth]{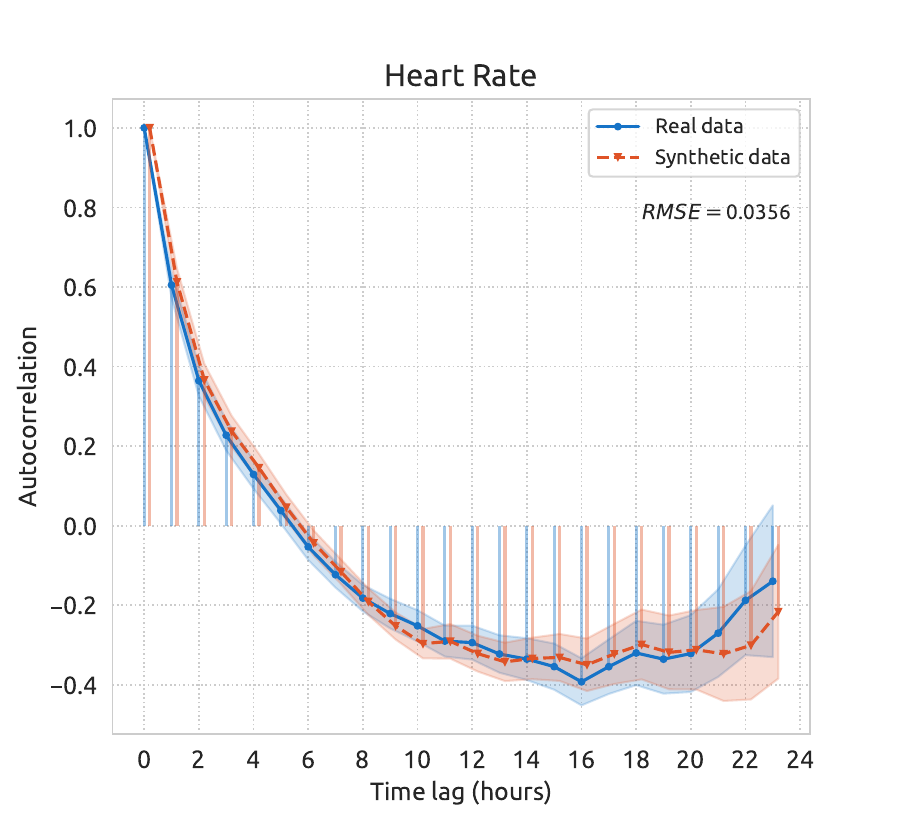}
		\captionsetup{justification=centering}
% 		\caption[ ]%
		{{\small }}    
	\end{subfigure}
\quad
	\begin{subfigure}[b]{0.22\textwidth}  
		\centering 
		\includegraphics[width=1.1\linewidth]{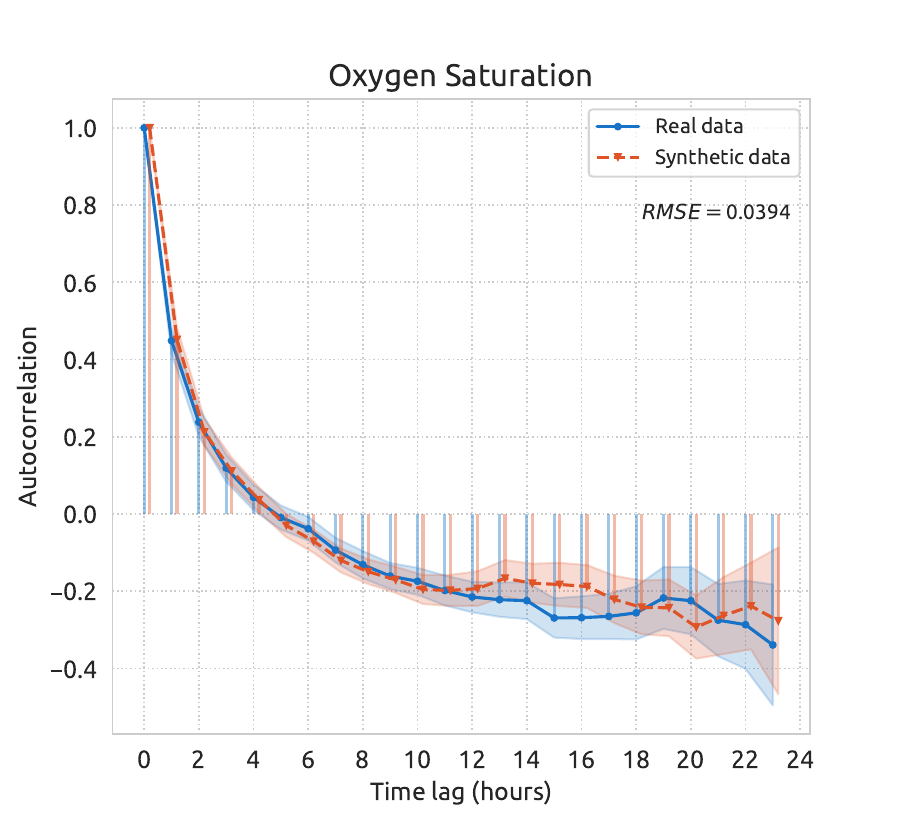}
		\captionsetup{justification=centering}
% 		\caption[ ]%
		{{\small  }}    
	\end{subfigure}
\quad
	\begin{subfigure}[b]{0.22\textwidth}  
		\centering 
		\includegraphics[width=1.1\linewidth]{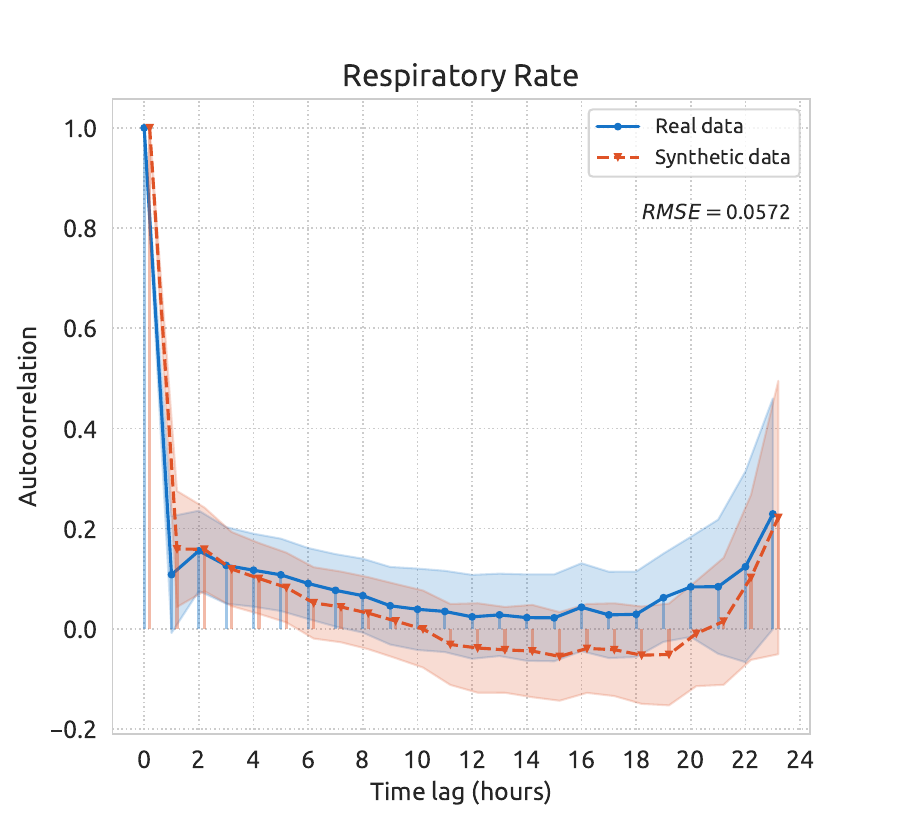}
		\captionsetup{justification=centering}
% 		\caption[ ]%
		{{\small }}    
	\end{subfigure}
\quad
	\begin{subfigure}[b]{0.22\textwidth}  
		\centering 
		\includegraphics[width=1.1\linewidth]{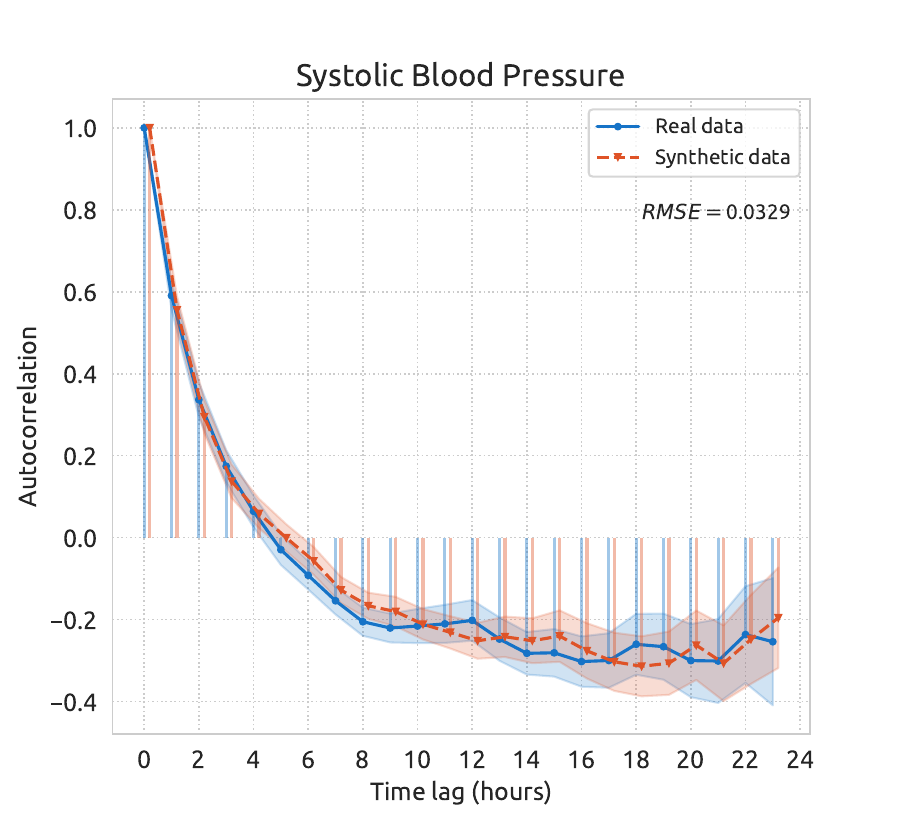}
		\captionsetup{justification=centering}
% 		\caption[ ]%
		{{\small }}    
	\end{subfigure}
	\medskip
	\begin{subfigure}[b]{0.22\textwidth}
		\centering
		\includegraphics[width=1.1\linewidth]{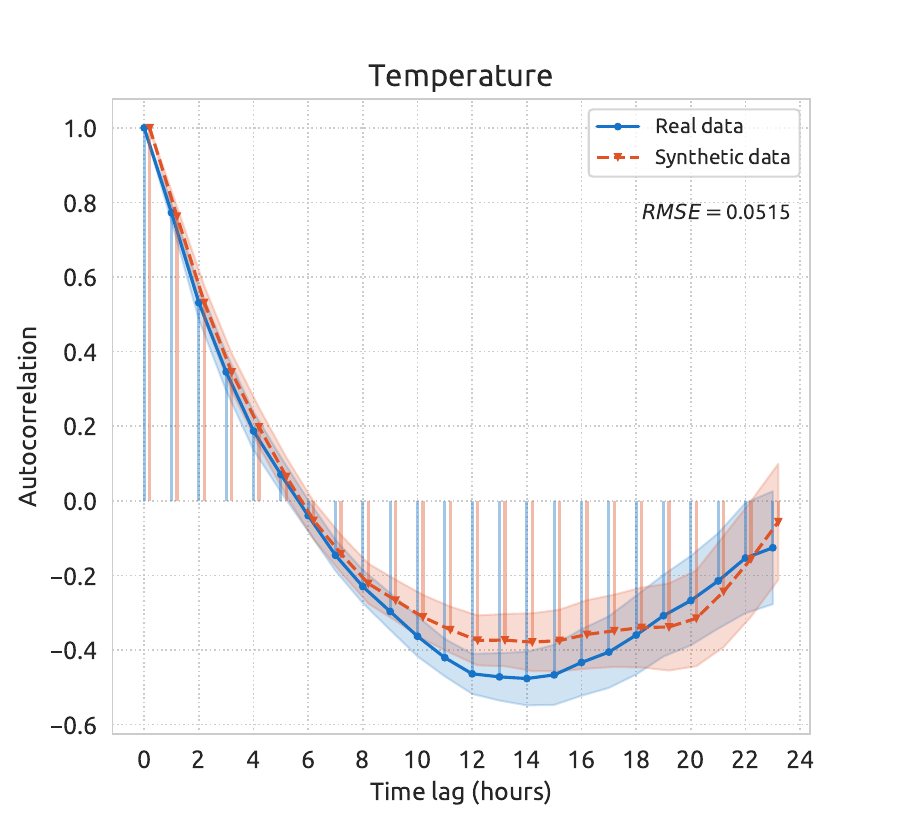}
		\captionsetup{justification=centering}
% 		\caption[ ]%
		{{\small }}    
	\end{subfigure}
\quad
	\begin{subfigure}[b]{0.22\textwidth}  
		\centering 
		\includegraphics[width=1.1\linewidth]{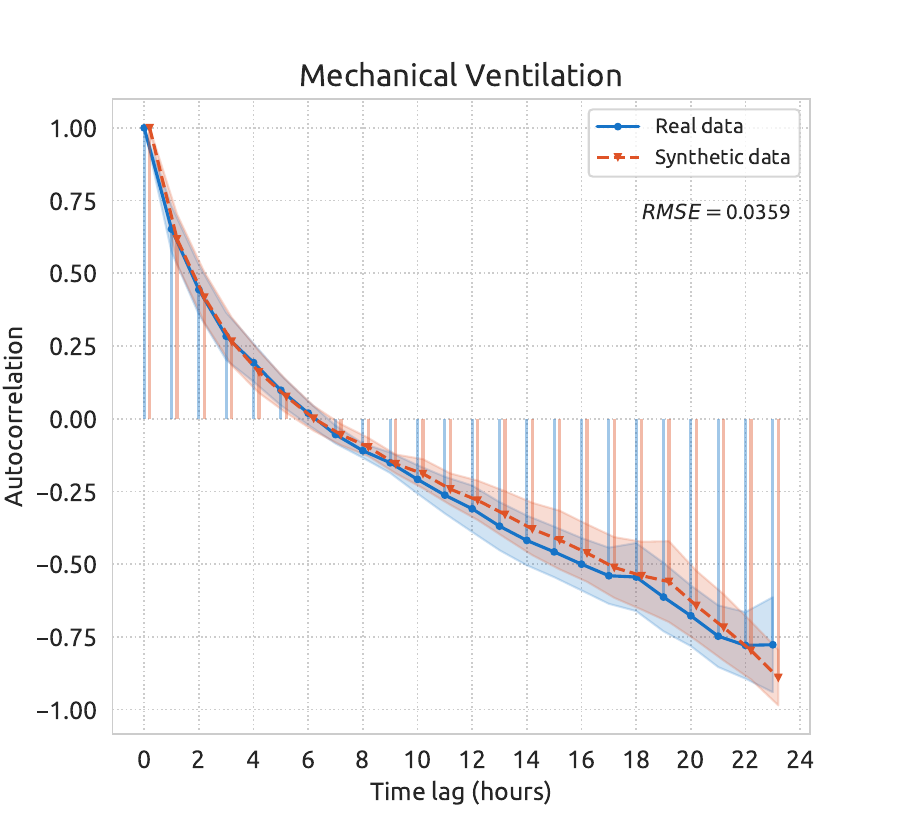}
		\captionsetup{justification=centering}
% 		\caption[ ]%
		{{\small  }}    
	\end{subfigure}
\quad
	\begin{subfigure}[b]{0.22\textwidth}  
		\centering 
		\includegraphics[width=1.1\linewidth]{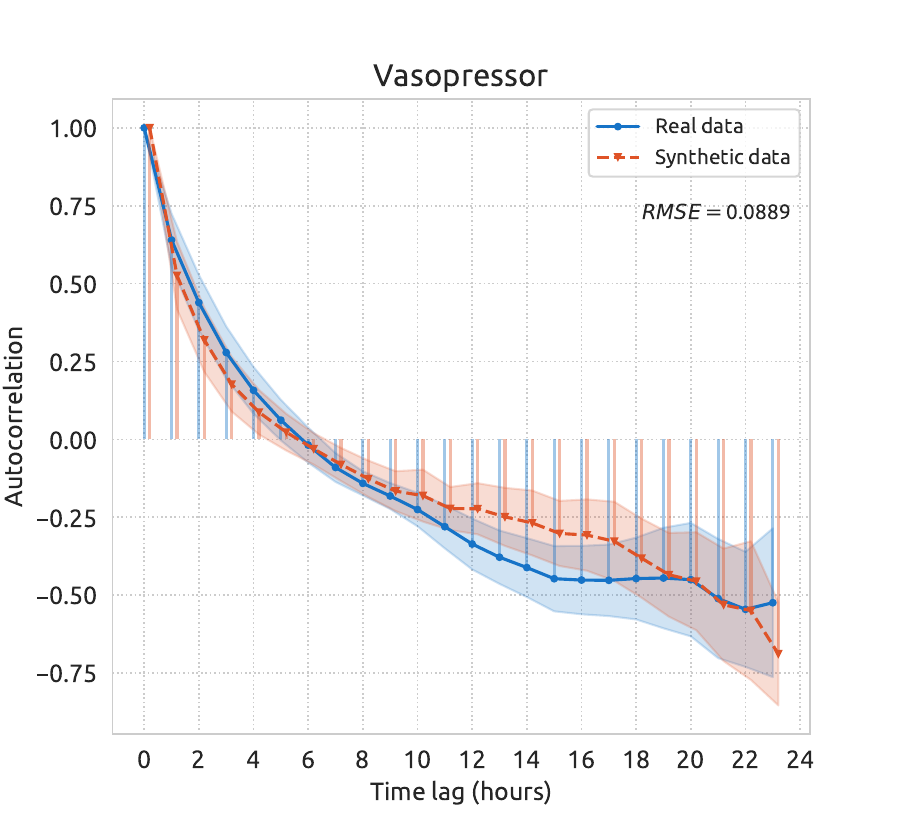}
		\captionsetup{justification=centering}
% 		\caption[ ]%
		{{\small }}    
	\end{subfigure}	
	\caption[ ]{\textbf{Autocorrelation function (ACF) of real data and synthetic data generated by EHR-M-GAN on HiRID dataset.} } 
	\label{fig_acf_hirid}
\end{figure*}

\newpage
\subsection{Embedding visualisation.}
\setlength{\parindent}{10pt}
\red{We apply t-SNE to qualitatively visualise the latent representations generated by EHR-M-GAN and EHR-M-GAN$_{\mathtt{cond}}$ on three critical care databases. The latent embedding vectors are induced by the encoders in the \textit{dual-VAE} during learning the shared latent space representations (See Methods section, p12, for details).
The t-SNE embedding results on raw timeseries are also included for comparison.}

It can be seen that better separability of the representation clusters in the shared latent space is shown in the embeddings obtained from EHR-M-GAN$_{\mathtt{cond}}$ compared with EHR-M-GAN and raw data. This illustrates the superiority of the EHR-M-GAN$_{\mathtt{cond}}$ in terms of learning the contextual information from the patient trajectories. 
\red{It therefore can be inferred that the conditional extension of the proposed model can further yield benefits by synthesizing condition-specific EHR timeseries with respect to distinctive patient health status.}

\begin{figure*}[h!]   
	\centering
	\begin{subfigure}[b]{0.32\textwidth}
		\centering
		\includegraphics[width=\textwidth]{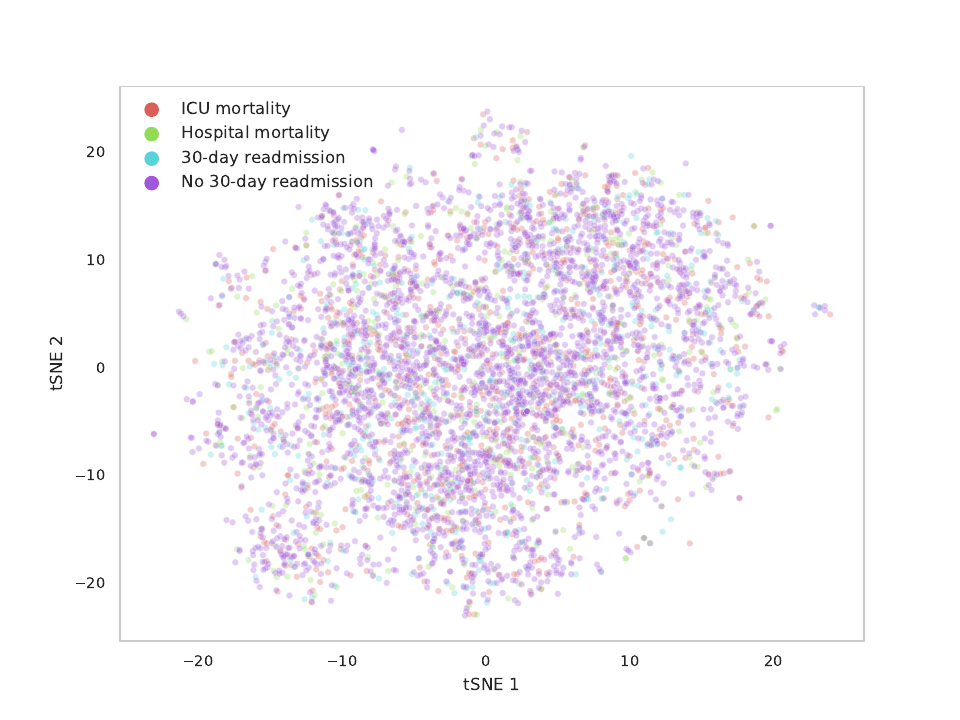}
		\captionsetup{justification=centering}
		\caption[ ]%
		{{\small  }}    
		\label{fig_tsne_raw}
	\end{subfigure}
	\hfill
	\begin{subfigure}[b]{0.32\textwidth}  
		\centering 
		\includegraphics[width=\textwidth]{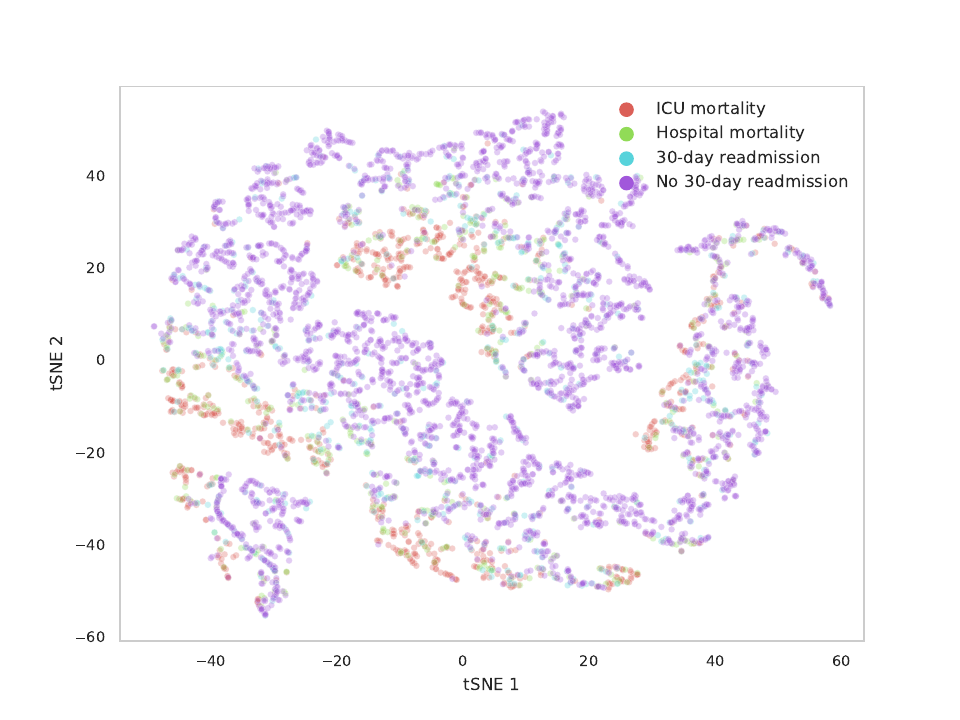}
		\captionsetup{justification=centering}
		\caption[ ]%
		{{\small    }}    
		\label{fig_tsne_emb}
	\end{subfigure}
	\hfill
	\begin{subfigure}[b]{0.32\textwidth}  
		\centering 
		\includegraphics[width=\textwidth]{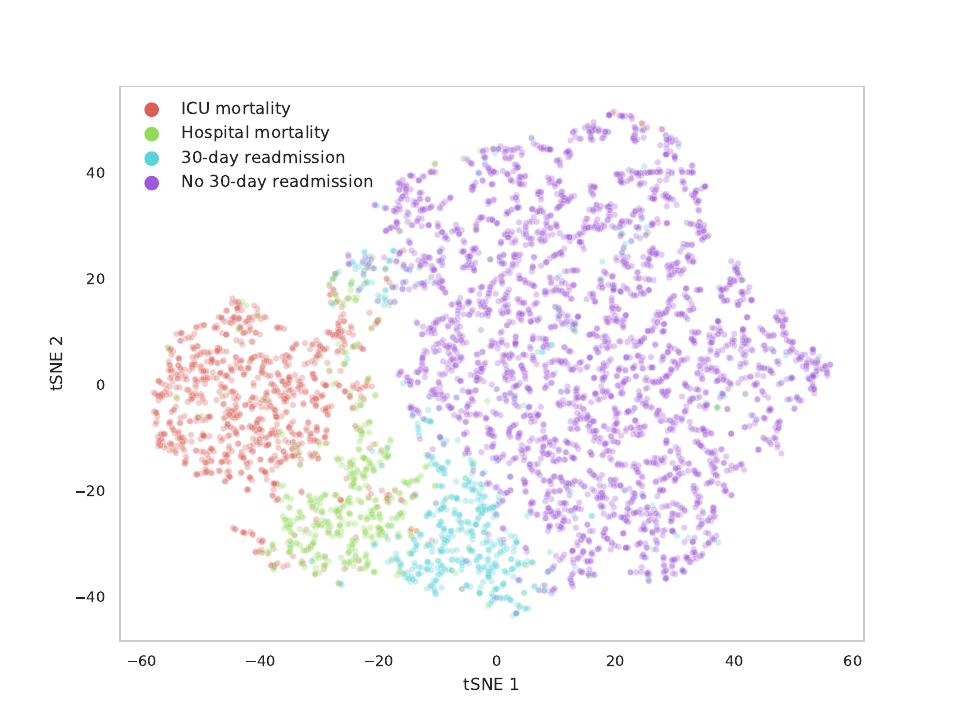}
		\captionsetup{justification=centering}
		\caption[ ]%
		{{\small   }}    
		\label{fig_tsne_emb}
	\end{subfigure}
		
	\caption[ ]
	{\textbf{t-SNE embedding visualization from MIMIC-III dataset} on (a) raw patient trajectories, (b) latent embeddings generated with EHR-M-GAN, and (c) latent embeddings generated with EHR-M-GAN$_{\mathtt{cond}}$.} 
	\label{fig_tsne1}
\end{figure*}

\begin{figure*}[h!]    
	\centering
	\begin{subfigure}[b]{0.32\textwidth}
		\centering
		\includegraphics[width=\textwidth]{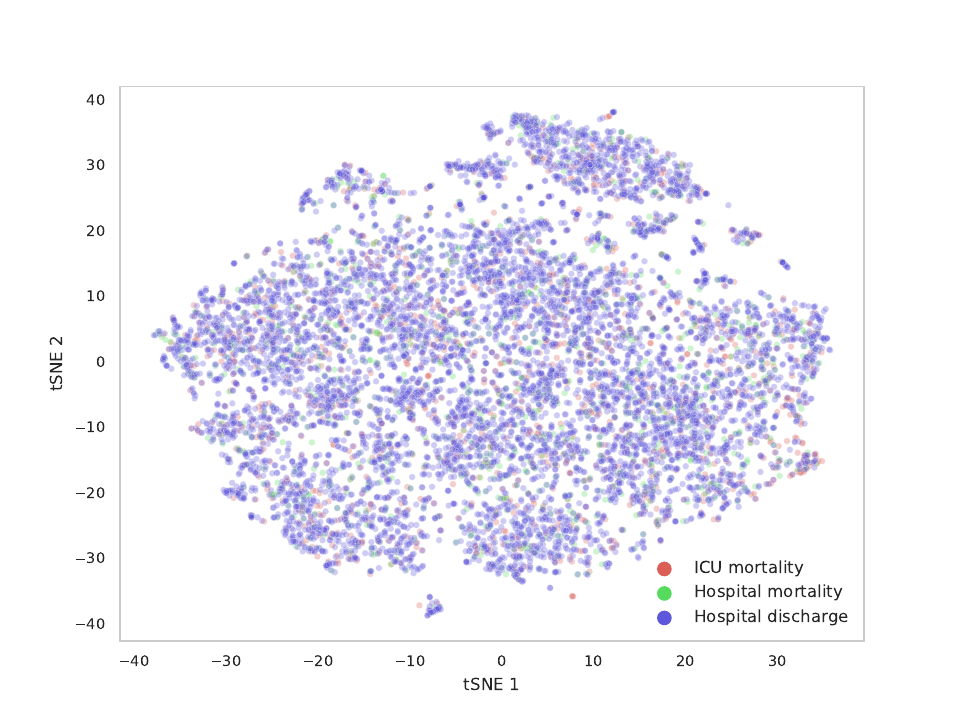}
		\captionsetup{justification=centering}
		\caption[ ]%
		{{\small   }}    
		\label{fig_tsne_raw}
	\end{subfigure}
	\hfill
	\begin{subfigure}[b]{0.32\textwidth}  
		\centering 
		\includegraphics[width=\textwidth]{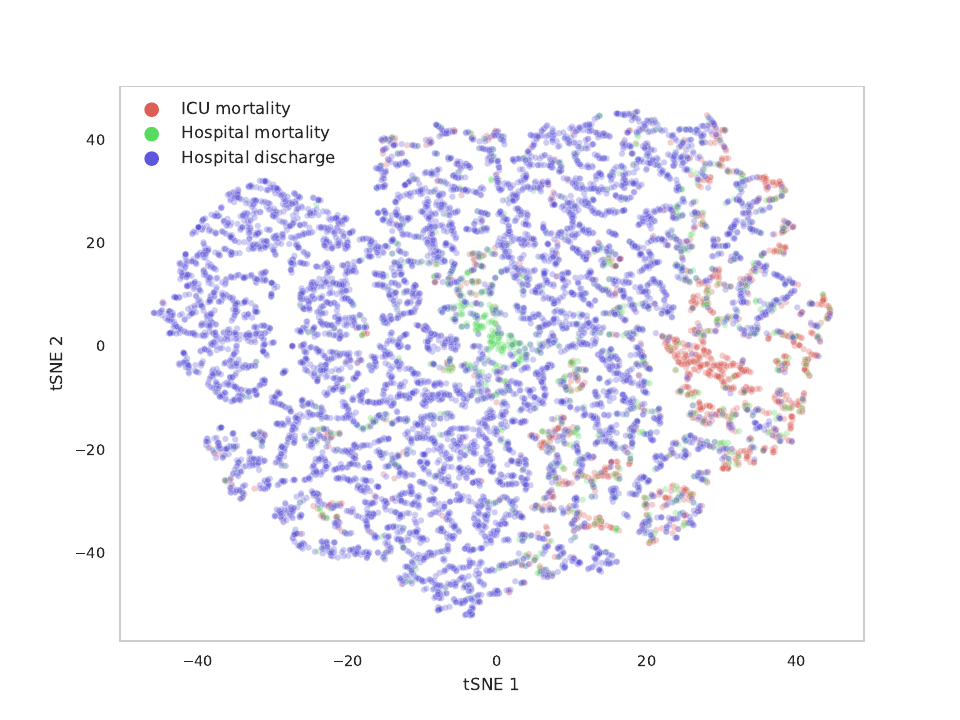}
		\captionsetup{justification=centering}
		\caption[ ]%
		{{\small  }}    
		\label{fig_tsne_emb}
	\end{subfigure}
	\hfill
	\begin{subfigure}[b]{0.32\textwidth}  
		\centering 
		\includegraphics[width=\textwidth]{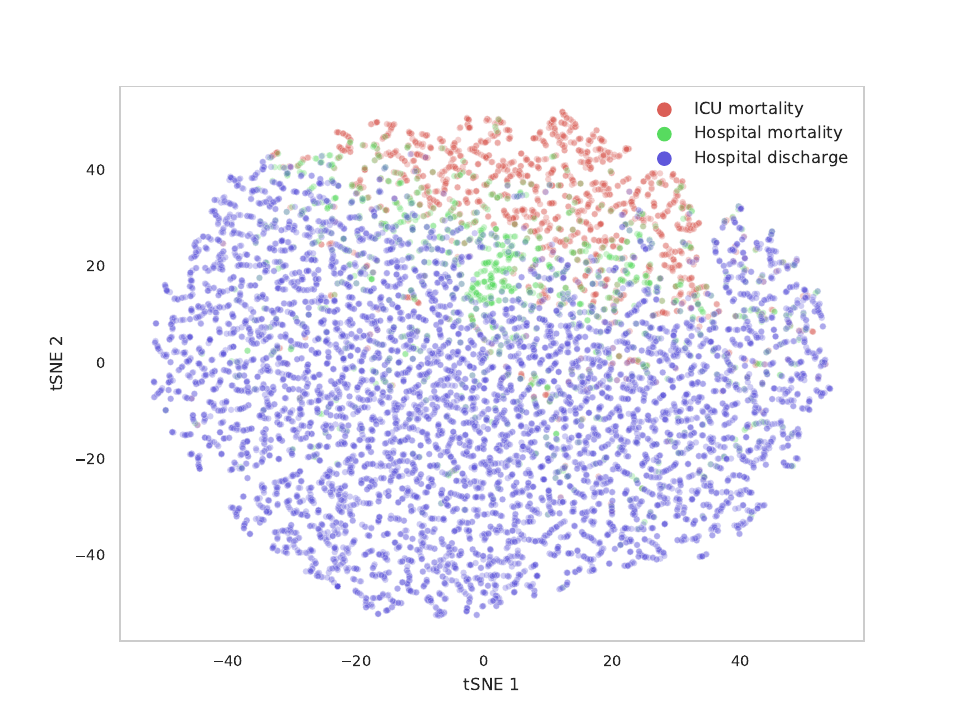}
		\captionsetup{justification=centering}
		\caption[ ]%
		{{\small   }}    
		\label{fig_tsne_emb}
	\end{subfigure}
		
	\caption[ ]
	{\textbf{t-SNE embedding visualization from eICU dataset} on (a) raw patient trajectories, (b) latent embeddings generated with EHR-M-GAN, and (c) latent embeddings generated with EHR-M-GAN$_{\mathtt{cond}}$.} 
	\label{fig_tsne2}
\end{figure*}

\begin{figure*}[h!]
	\centering
	\begin{subfigure}[b]{0.32\textwidth}
		\centering
		\includegraphics[width=\textwidth]{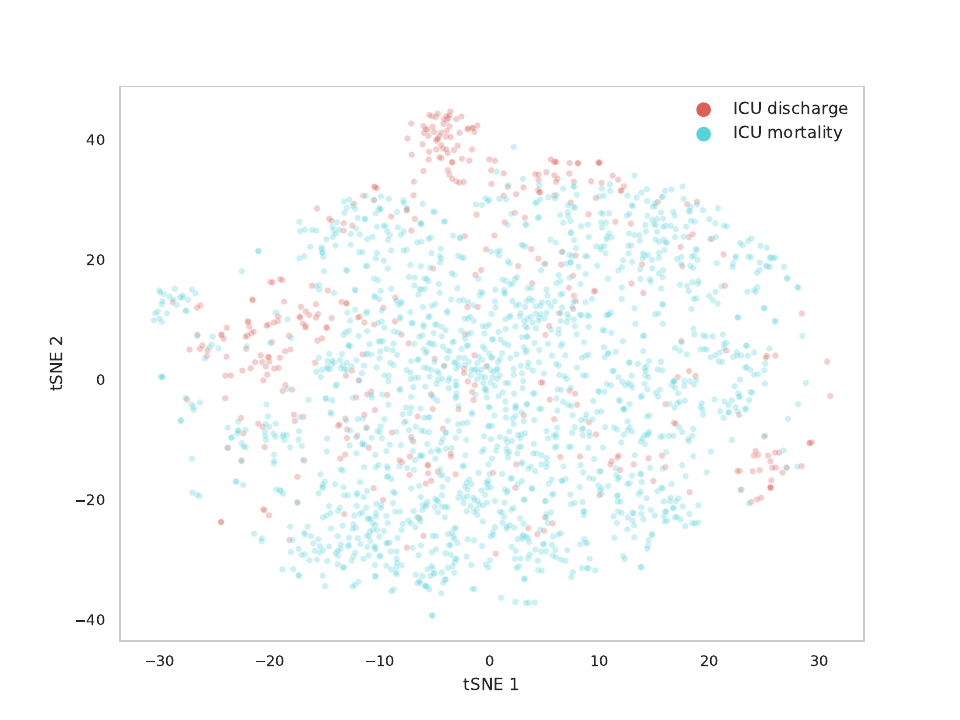}
		\captionsetup{justification=centering}
		\caption[ ]%
		{{\small }}    
		\label{fig_tsne_raw}
	\end{subfigure}
	\hfill
	\begin{subfigure}[b]{0.32\textwidth}  
		\centering 
		\includegraphics[width=\textwidth]{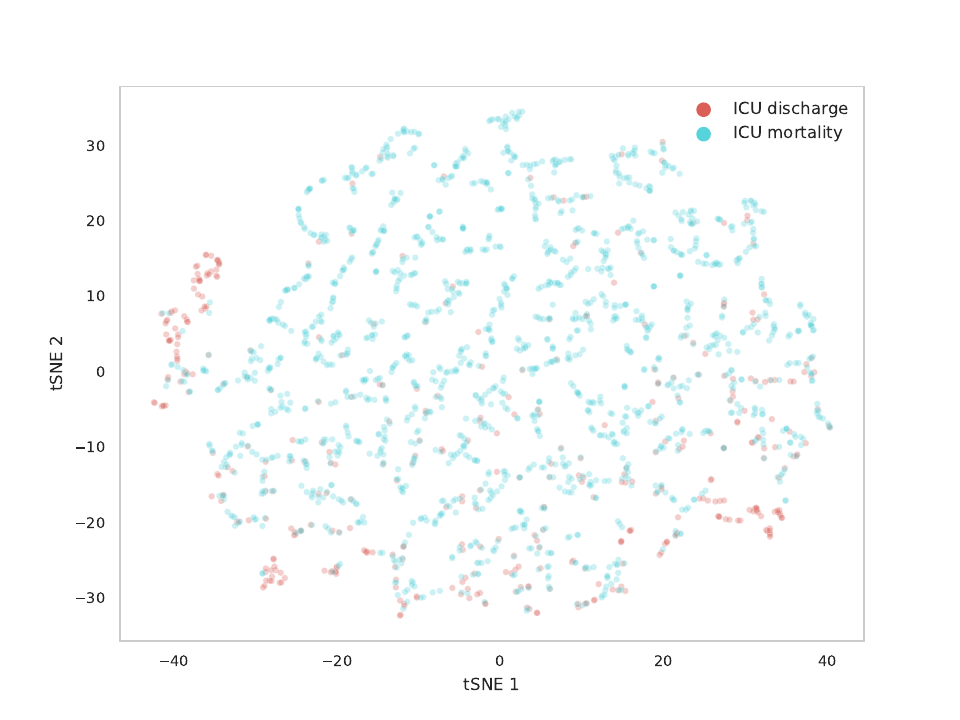}
		\captionsetup{justification=centering}
		\caption[ ]%
		{{\small  }}    
		\label{fig_tsne_emb}
	\end{subfigure}
	\hfill
	\begin{subfigure}[b]{0.32\textwidth}  
		\centering 
		\includegraphics[width=\textwidth]{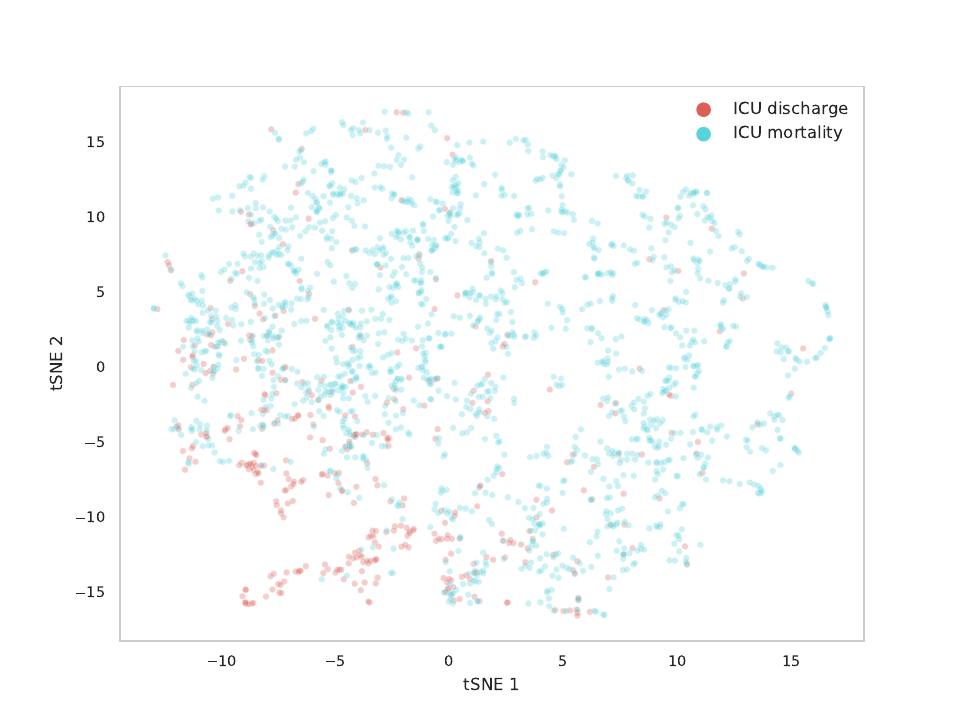}
		\captionsetup{justification=centering}
		\caption[ ]%
		{{\small }}    
		\label{fig_tsne_emb}
	\end{subfigure}
		
	\caption[ ]
	{\textbf{t-SNE embedding visualization from HiRID dataset} on (a) raw patient trajectories, (b) latent embeddings generated with EHR-M-GAN, and (c) latent embeddings generated with EHR-M-GAN$_{\mathtt{cond}}$.} 
	\label{fig_tsne3}
\end{figure*}

\newpage
\subsection{Patient trajectories visualisation.} 
\begin{figure*}[h!]
			\centering
			\begin{subfigure}[b]{0.135\textwidth}
				\centering
				\includegraphics[width=1\linewidth]{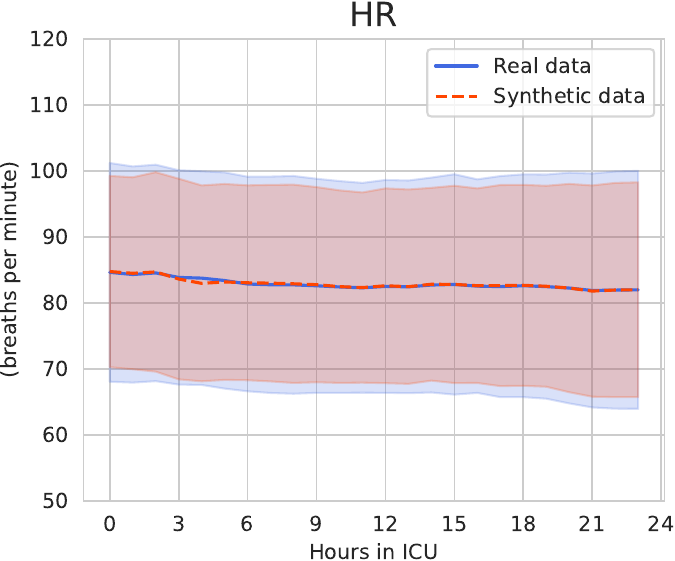} 
			\end{subfigure}
			 %\hspace{0.25em}
			\begin{subfigure}[b]{0.135\textwidth}  
				\centering 
				\includegraphics[width=1\linewidth]{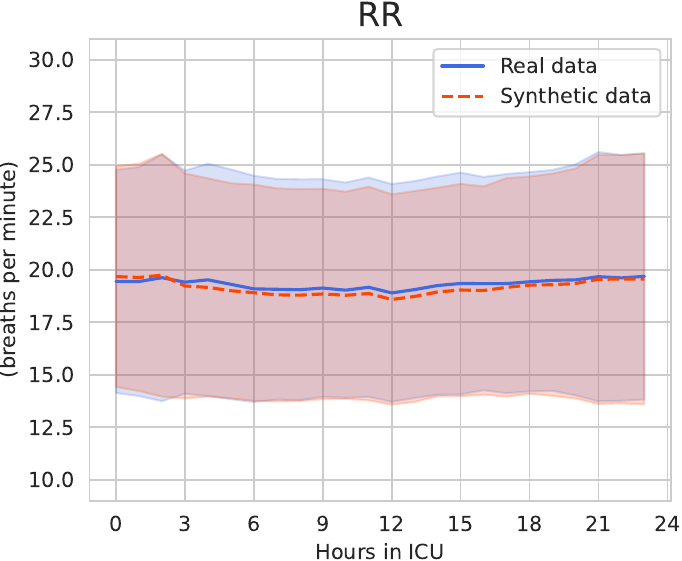} 
				\end{subfigure}
			 %\hspace{0.25em}
			\begin{subfigure}[b]{0.135\textwidth}  
				\centering 
				\includegraphics[width=1\linewidth]{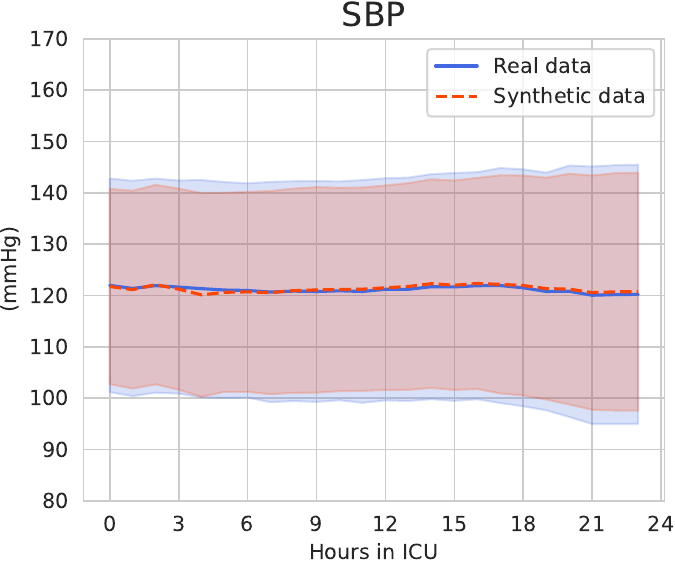} 
				\end{subfigure}
			 %\hspace{0.25em}
			\begin{subfigure}[b]{0.135\textwidth}  
				\centering 
				\includegraphics[width=1\linewidth]{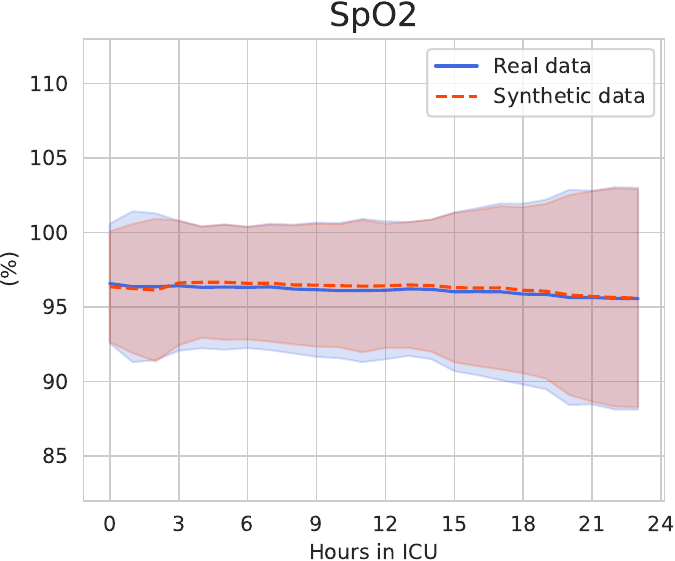} 
				\end{subfigure}
			 %\hspace{0.25em}
			\begin{subfigure}[b]{0.135\textwidth}  
				\centering 
				\includegraphics[width=1\linewidth]{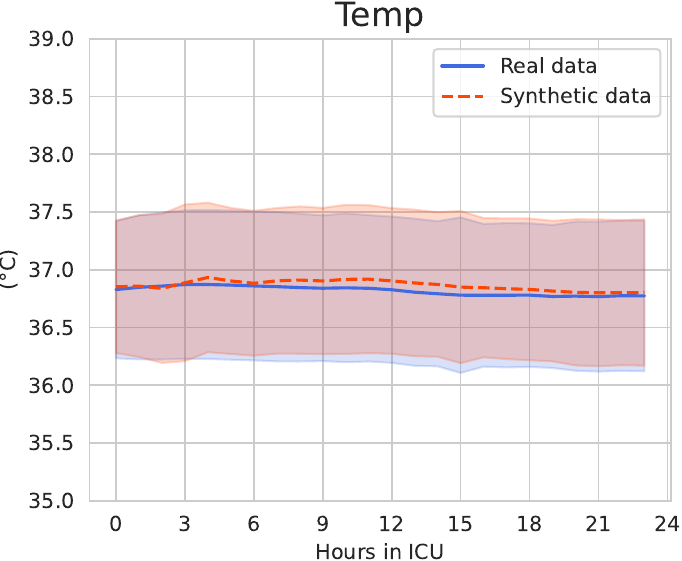} 
				\end{subfigure}
			 %\hspace{0.25em}
			\begin{subfigure}[b]{0.135\textwidth}  
				\centering 
				\includegraphics[width=1\linewidth]{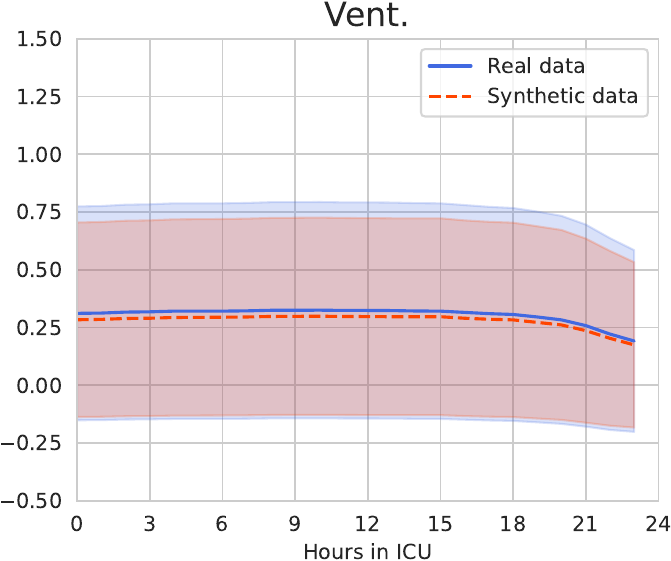} 
				\end{subfigure}
			 %\hspace{0.25em}
			\begin{subfigure}[b]{0.135\textwidth}  
				\centering 
				\includegraphics[width=1\linewidth]{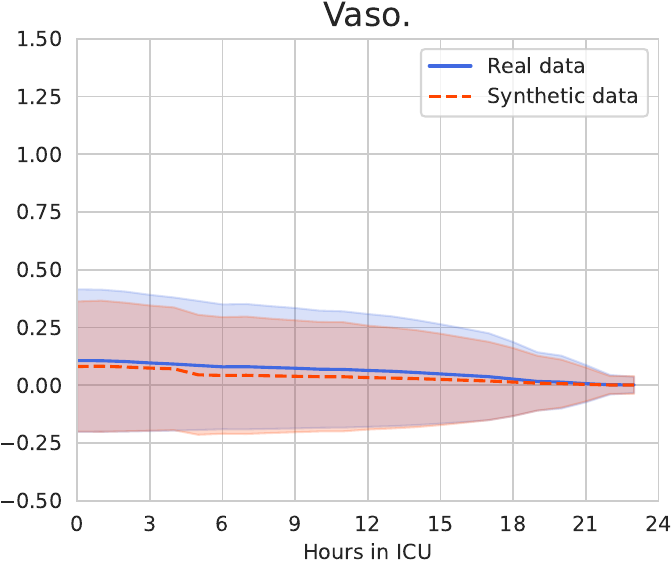} 
				\end{subfigure}
\caption{\red{\textbf{Comparison of patient trajectories.} The distribution of values at each timepoint (mean and standard deviation) are compared between real and synthetic patient trajectory produced by EHR-M-GAN$_{\mathtt{cond}}$, under the condition of \textbf{ICU mortality}.}}
	\label{fig_patient_traj}
\end{figure*}

\begin{figure*}[h!]
			\centering
			\begin{subfigure}[b]{0.135\textwidth}
				\centering
				\includegraphics[width=1\linewidth]{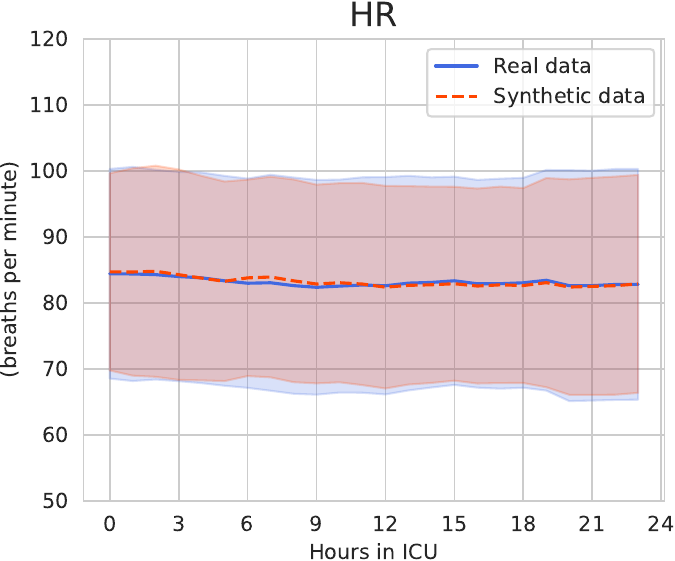} 
			\end{subfigure}
			 %\hspace{0.25em}
			\begin{subfigure}[b]{0.135\textwidth}  
				\centering 
				\includegraphics[width=1\linewidth]{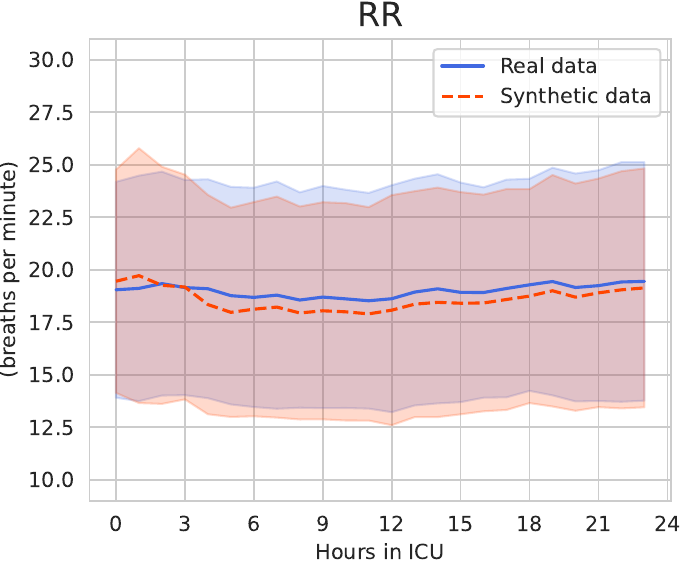} 
				\end{subfigure}
			 %\hspace{0.25em}
			\begin{subfigure}[b]{0.135\textwidth}  
				\centering 
				\includegraphics[width=1\linewidth]{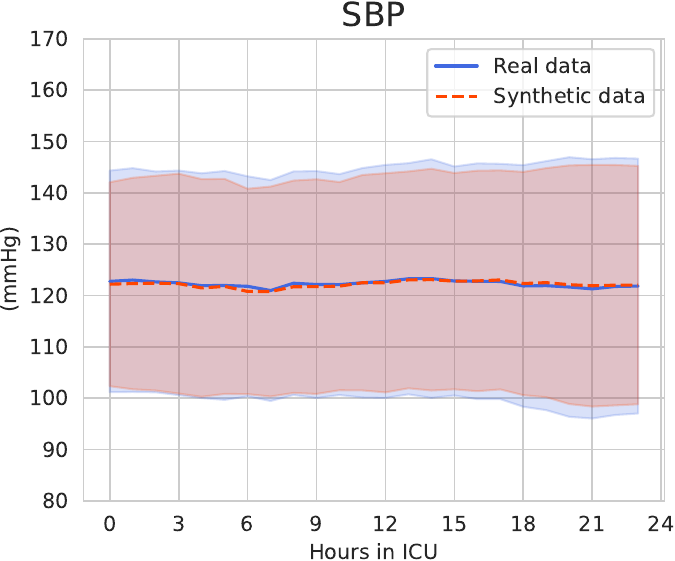} 
				\end{subfigure}
			 %\hspace{0.25em}
			\begin{subfigure}[b]{0.135\textwidth}  
				\centering 
				\includegraphics[width=1\linewidth]{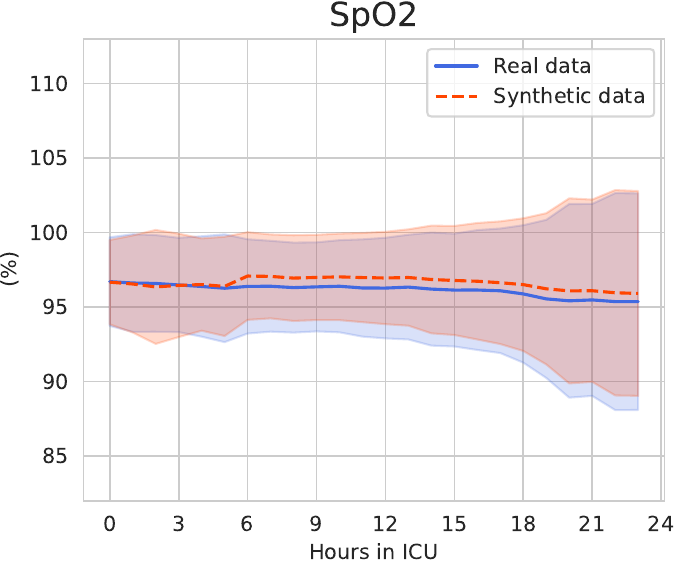} 
				\end{subfigure}
			 %\hspace{0.25em}
			\begin{subfigure}[b]{0.135\textwidth}  
				\centering 
				\includegraphics[width=1\linewidth]{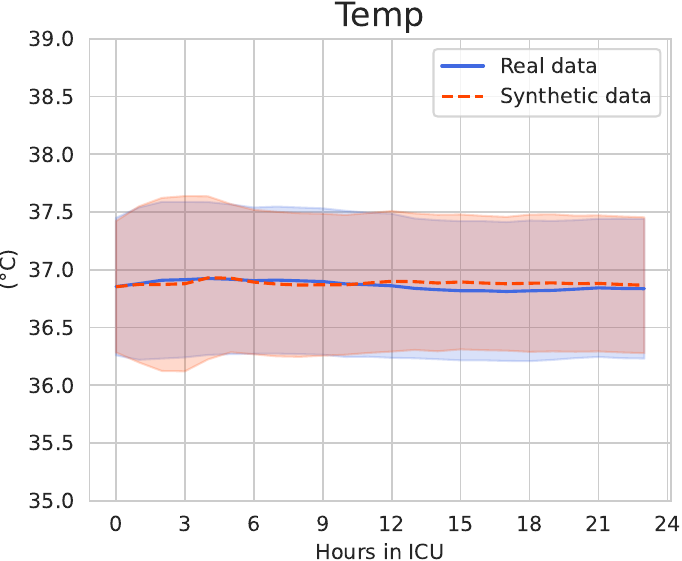} 
				\end{subfigure}
			 %\hspace{0.25em}
			\begin{subfigure}[b]{0.135\textwidth}  
				\centering 
				\includegraphics[width=1\linewidth]{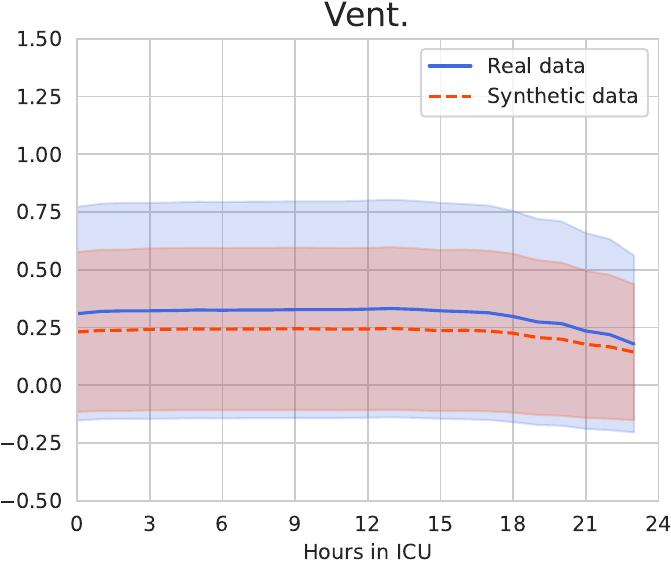} 
				\end{subfigure}
			 %\hspace{0.25em}
			\begin{subfigure}[b]{0.135\textwidth}  
				\centering 
				\includegraphics[width=1\linewidth]{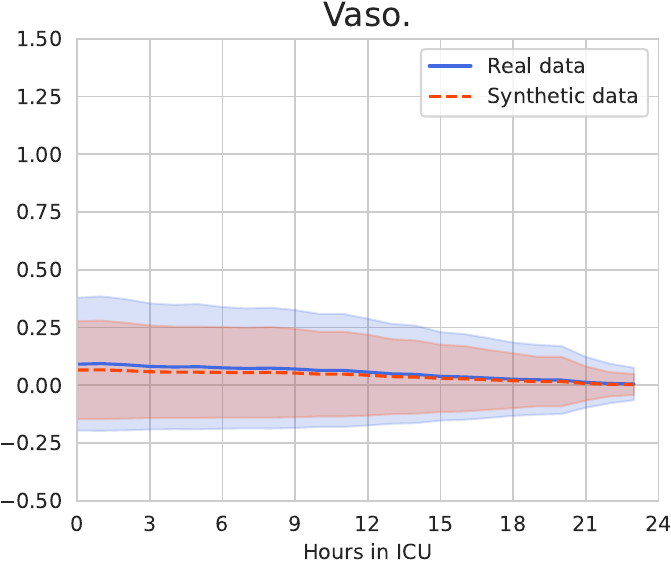} 
				\end{subfigure}
\caption{\red{\textbf{Comparison of patient trajectories.} The distribution of values at each timepoint (mean and standard deviation) are compared between real and synthetic patient trajectory produced by EHR-M-GAN$_{\mathtt{cond}}$, under the condition of \textbf{Hospital mortality}.}}
	\label{fig_patient_traj}
\end{figure*}

\begin{figure*}[h!]
			\centering
			\begin{subfigure}[b]{0.135\textwidth}
				\centering
				\includegraphics[width=1\linewidth]{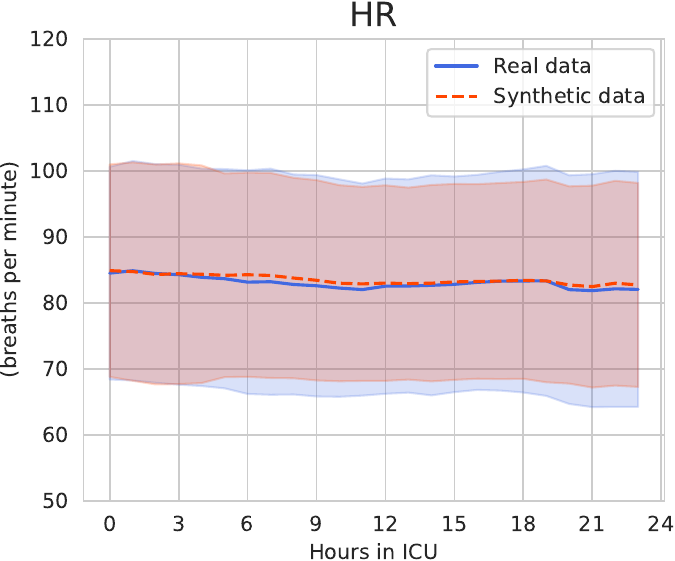} 
			\end{subfigure}
			 %\hspace{0.25em}
			\begin{subfigure}[b]{0.135\textwidth}  
				\centering 
				\includegraphics[width=1\linewidth]{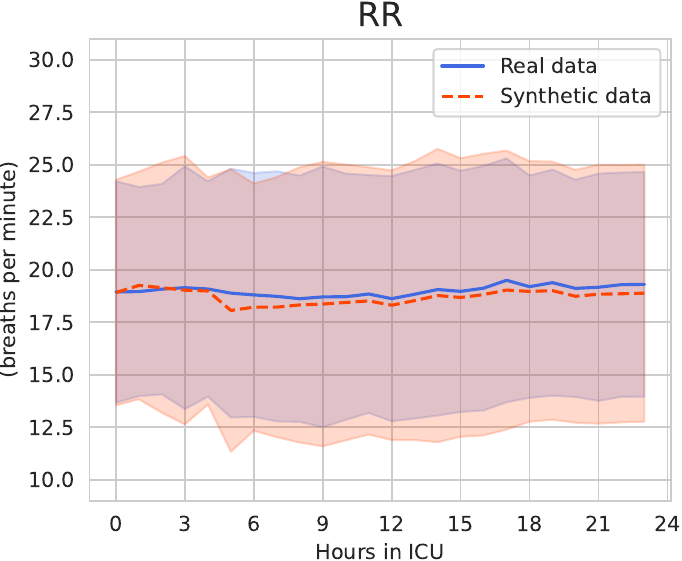} 
				\end{subfigure}
			 %\hspace{0.25em}
			\begin{subfigure}[b]{0.135\textwidth}  
				\centering 
				\includegraphics[width=1\linewidth]{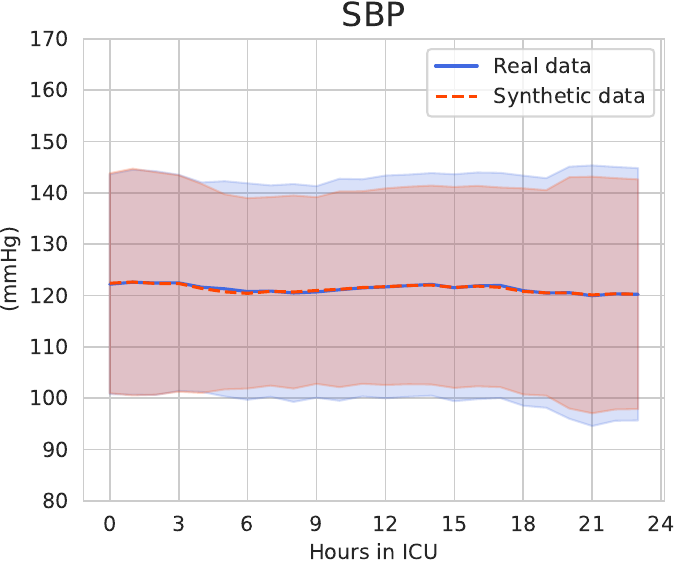} 
				\end{subfigure}
			 %\hspace{0.25em}
			\begin{subfigure}[b]{0.135\textwidth}  
				\centering 
				\includegraphics[width=1\linewidth]{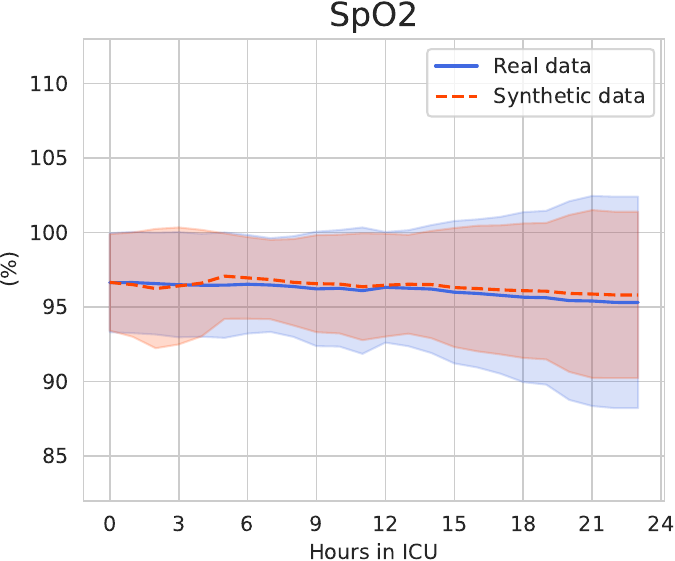} 
				\end{subfigure}
			 %\hspace{0.25em}
			\begin{subfigure}[b]{0.135\textwidth}  
				\centering 
				\includegraphics[width=1\linewidth]{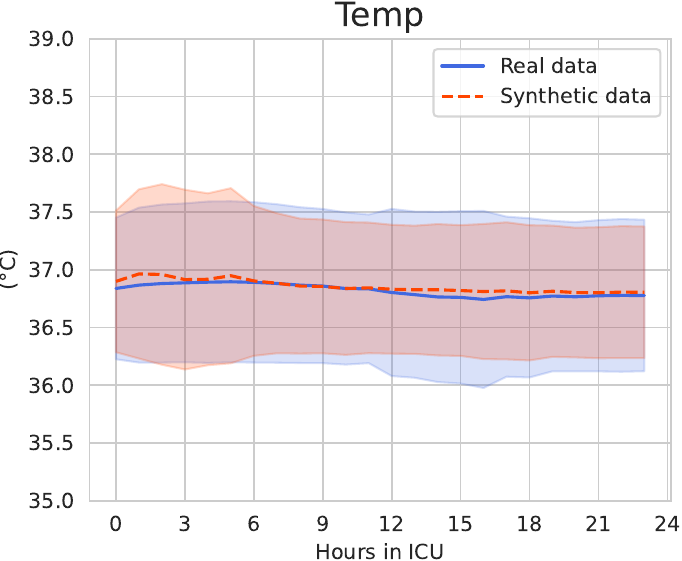} 
				\end{subfigure}
			 %\hspace{0.25em}
			\begin{subfigure}[b]{0.135\textwidth}  
				\centering 
				\includegraphics[width=1\linewidth]{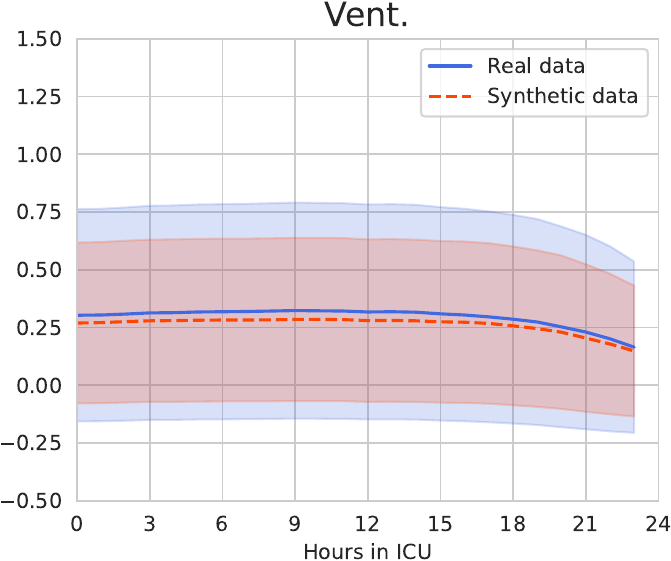} 
				\end{subfigure}
			 %\hspace{0.25em}
			\begin{subfigure}[b]{0.135\textwidth}  
				\centering 
				\includegraphics[width=1\linewidth]{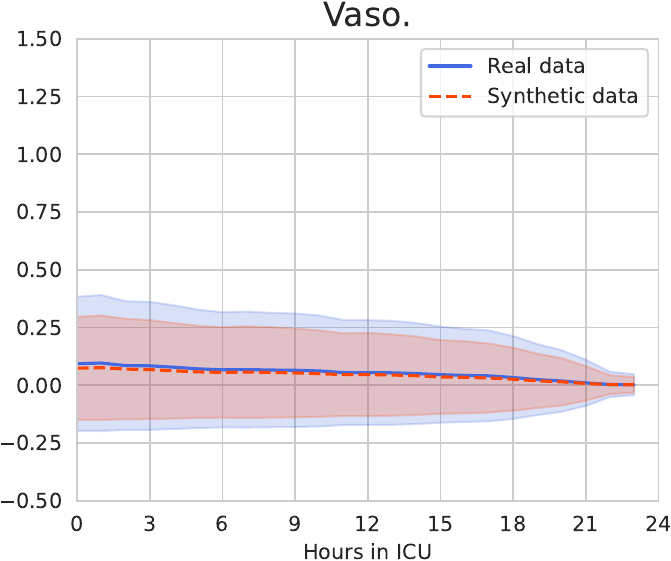} 
				\end{subfigure}
\caption{\red{\textbf{Comparison of patient trajectories.} The distribution of values at each timepoint (mean and standard deviation) are compared between real and synthetic patient trajectory produced by EHR-M-GAN$_{\mathtt{cond}}$, under the condition of \textbf{30-day readmission}.}}
	\label{fig_patient_traj}
\end{figure*}

\begin{figure*}[h!]
			\centering
			\begin{subfigure}[b]{0.135\textwidth}
				\centering
				\includegraphics[width=1\linewidth]{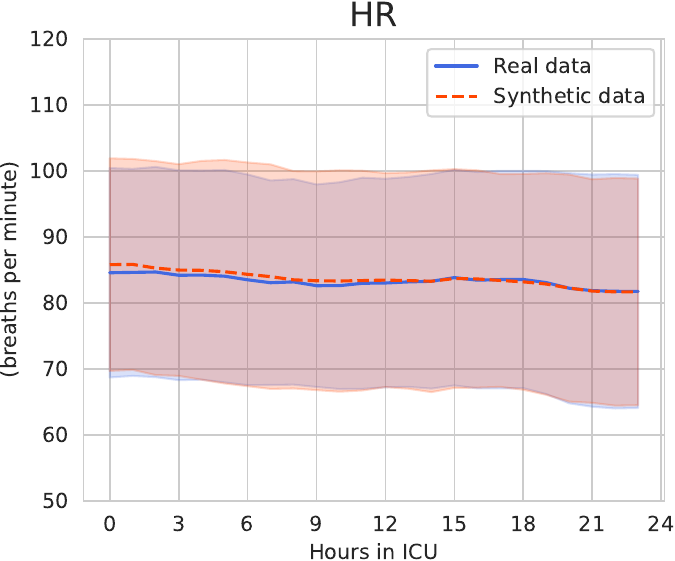} 
			\end{subfigure}
			 %\hspace{0.25em}
			\begin{subfigure}[b]{0.135\textwidth}  
				\centering 
				\includegraphics[width=1\linewidth]{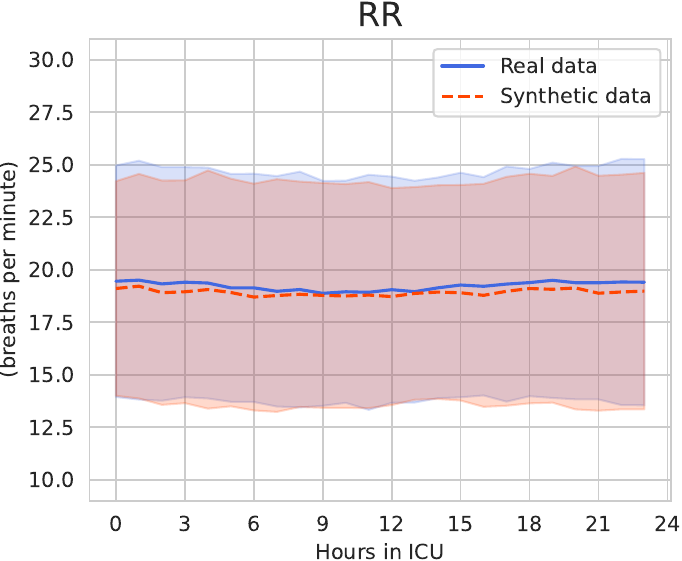} 
				\end{subfigure}
			 %\hspace{0.25em}
			\begin{subfigure}[b]{0.135\textwidth}  
				\centering 
				\includegraphics[width=1\linewidth]{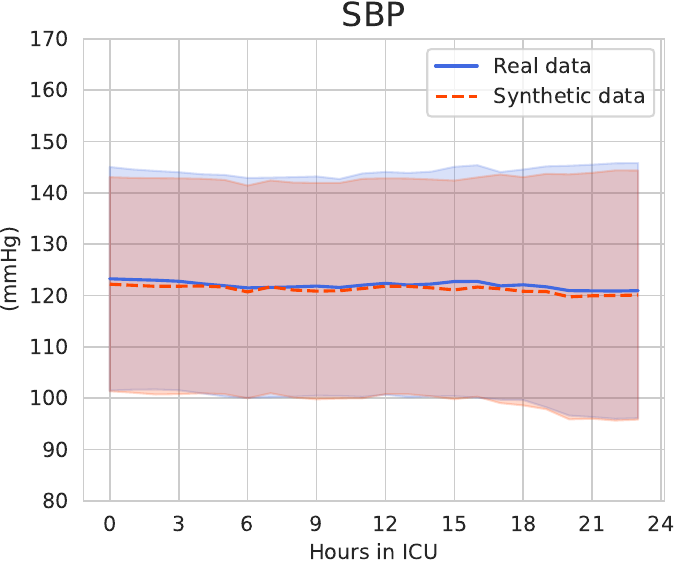} 
				\end{subfigure}
			 %\hspace{0.25em}
			\begin{subfigure}[b]{0.135\textwidth}  
				\centering 
				\includegraphics[width=1\linewidth]{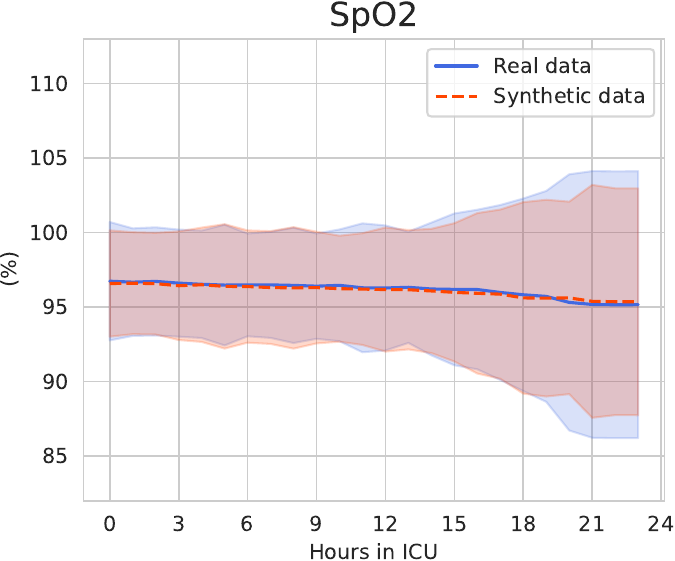} 
				\end{subfigure}
			 %\hspace{0.25em}
			\begin{subfigure}[b]{0.135\textwidth}  
				\centering 
				\includegraphics[width=1\linewidth]{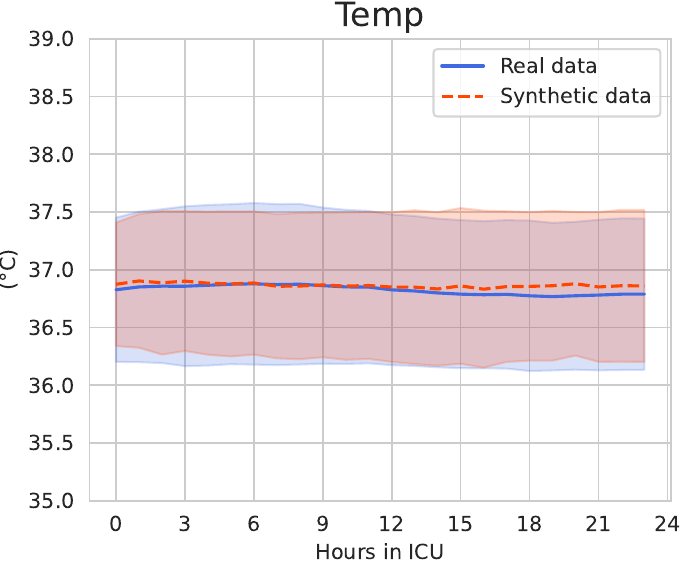} 
				\end{subfigure}
			 %\hspace{0.25em}
			\begin{subfigure}[b]{0.135\textwidth}  
				\centering 
				\includegraphics[width=1\linewidth]{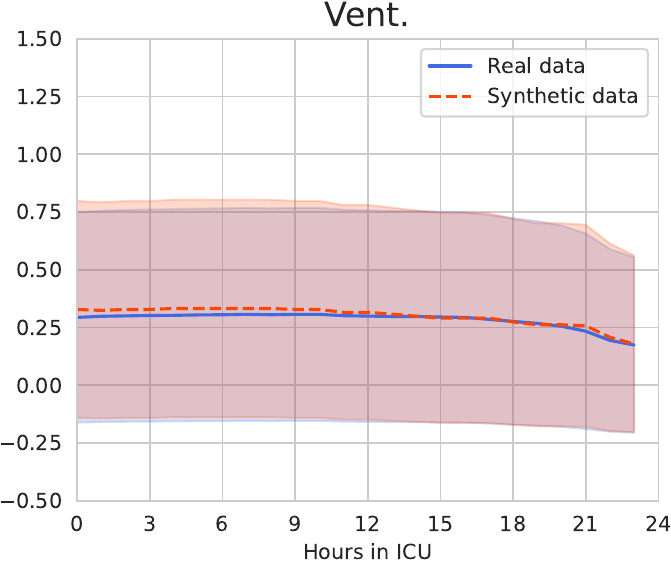} 
				\end{subfigure}
			 %\hspace{0.25em}
			\begin{subfigure}[b]{0.135\textwidth}  
				\centering 
				\includegraphics[width=1\linewidth]{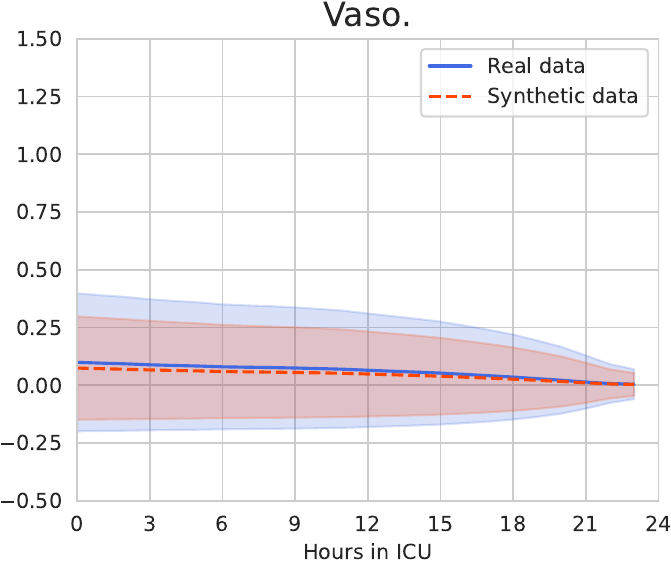} 
				\end{subfigure}
\caption{\red{\textbf{Comparison of patient trajectories.} The distribution of values at each timepoint (mean and standard deviation) are compared between real and synthetic patient trajectory produced by EHR-M-GAN$_{\mathtt{cond}}$, under the condition of \textbf{No 30-day readmission}.}}
	\label{fig_patient_traj}
\end{figure*}

% \begin{figure}[h!]

% 	\begin{subfigure}{\columnwidth}
% 		\centering
% 		\includegraphics[width=1.\textwidth]{fig/traj_plot/traj_mimic_cond1.pdf}
% 		\caption{Sampled patient trajectories on MIMIC-III dataset using EHR-M-GAN$_{\mathtt{cond}}$ under the conditional information of patient status \textbf{Hospital mortality}.}
% 	\end{subfigure}
% 	\vskip 5pt
% 	\begin{subfigure}{\columnwidth}
% 		\centering
% 		\includegraphics[width=1.\textwidth]{fig/traj_plot/traj_mimic_cond2.pdf}
% 		\caption{Sampled patient trajectories on MIMIC-III dataset using EHR-M-GAN$_{\mathtt{cond}}$ under the conditional information of patient status \textbf{30-day readmission}.}
% 	\end{subfigure}
% 	\vskip 5pt
% 	\begin{subfigure}{\columnwidth}
% 		\centering
% 		\includegraphics[width=1.\textwidth]{fig/traj_plot/traj_mimic_cond3.pdf}
% 		\caption{Sampled patient trajectories on MIMIC-III dataset using EHR-M-GAN$_{\mathtt{cond}}$ under the conditional information of patient status \textbf{No 30-day readmission}.}
% 	\end{subfigure}
% 	\caption{\textbf{Synthetic patient trajectories visulisation for MIMIC-III dataset.} Timeseries are generated by EHR-M-GAN$_{\mathtt{cond}}$, within 24 hours before patients being discharged from ICU on specified features, including \textit{Oxygen Saturation}, \textit{Systolic Blood Pressure}, \textit{Respiratory Rate}, \textit{Heart Rate}, \textit{Temperature}, \textit{Vasopressor} and \textit{Mechanical Ventilation}. The binary value of intervention signs indicates whether such medical intervention are put into use.}\label{fig_traj}
% \end{figure}

\newpage

\newpage
% Bibliography
\bibliography{ref_appendix}